\documentclass{article}

\PassOptionsToPackage{numbers, sort&compress}{natbib}
\usepackage[main, final]{neurips_2026}

\usepackage[utf8]{inputenc} 
\usepackage[T1]{fontenc}    
\usepackage[colorlinks,linkcolor={blue!50!black},citecolor={blue!50!black},urlcolor={blue!50!black},backref=page,bookmarks=false]{hyperref}
\usepackage{url}            
\usepackage{amsfonts}       
\usepackage{amsmath}        
\usepackage{nicefrac}       
\usepackage{microtype}      
\usepackage{xcolor}         
\usepackage{graphicx}       
\usepackage{enumitem}       
\usepackage{algorithm}
\usepackage{algpseudocode}
\usepackage{subcaption}
\usepackage{enumitem}
\usepackage{makecell}
\usepackage{booktabs}

\usepackage{listings}
\usepackage{float}

\definecolor{PinductorBlue}{RGB}{30,90,180}
\definecolor{PinductorBlueLight}{RGB}{220,232,250}
\definecolor{CurtisOrange}{RGB}{200,100,30}
\definecolor{CurtisOrangeLight}{RGB}{255,235,210}

\newif\ifshowdiff
\showdifffalse
\ifshowdiff
  \newcommand{\oldtext}[1]{{\color{red}#1}}
  \newcommand{\newtext}[1]{{\color{blue}#1}}
\else
  \newcommand{\oldtext}[1]{}
  \newcommand{\newtext}[1]{#1}
\fi

\lstdefinestyle{oursprompt}{%
  basicstyle=\ttfamily\scriptsize,
  breaklines=true, breakatwhitespace=true,
  columns=fullflexible, keepspaces=true, showstringspaces=false,
  frame=single, rulecolor=\color{PinductorBlue},
  backgroundcolor=\color{PinductorBlueLight!40},
  framesep=4pt, xleftmargin=4pt, xrightmargin=4pt,
  aboveskip=2pt, belowskip=8pt,
  extendedchars=true, inputencoding=utf8, literate={—}{{--}}1{─}{{-}}1
}

\lstdefinestyle{curtisprompt}{%
  basicstyle=\ttfamily\scriptsize,
  breaklines=true, breakatwhitespace=true,
  columns=fullflexible, keepspaces=true, showstringspaces=false,
  frame=single, rulecolor=\color{CurtisOrange},
  backgroundcolor=\color{CurtisOrangeLight!40},
  framesep=4pt, xleftmargin=4pt, xrightmargin=4pt,
  aboveskip=2pt, belowskip=8pt,
  extendedchars=true, inputencoding=utf8, literate={—}{{--}}1{─}{{-}}1
}

\lstdefinestyle{auditcode}{%
  basicstyle=\ttfamily\scriptsize, breaklines=true, columns=fullflexible,
  keepspaces=true, showstringspaces=false, frame=single,
  rulecolor=\color{black!40}, framesep=4pt, xleftmargin=4pt, xrightmargin=4pt,
  aboveskip=2pt, belowskip=8pt, language=Python
}

\newcommand{\promptheader}[2]{%
  \par\smallskip\noindent
  \colorbox{#1}{\strut~\color{white}\bfseries #2~}\par\nopagebreak\vspace{1pt}%
}

\usepackage{xcolor}

\AtBeginDocument{%
  \setlength{\abovedisplayskip}{8pt plus 2pt minus 2pt}%
  \setlength{\belowdisplayskip}{8pt plus 2pt minus 2pt}%
  \setlength{\abovedisplayshortskip}{6pt plus 2pt minus 2pt}%
  \setlength{\belowdisplayshortskip}{7pt plus 2pt minus 2pt}%
}

\title{Learning POMDP World Models from Observations with Language-Model Priors}

\author{%
\makebox[\textwidth][c]{%
\begin{tabular}{c}
Valentin Six\textsuperscript{1,*},
Frederik Panse\textsuperscript{1,4,*},
Mathis Fajeau\textsuperscript{*},
Lancelot Da Costa\textsuperscript{1,*},
\\
Mridul Sharma\textsuperscript{2,\textdagger},
Alfonso Amayuelas\textsuperscript{3,\textdagger},
Tim Z. Xiao\textsuperscript{1,4},
David Hyland\textsuperscript{5},
\\
Philipp Hennig\textsuperscript{4},
Bernhard Sch\"olkopf\textsuperscript{1,6}
\\[1.2em]
{\normalfont\small
\textsuperscript{1}Max Planck Institute for Intelligent Systems \quad
\textsuperscript{2}IRIIS \quad
\textsuperscript{3}University of California, Santa Barbara
}
\\
{\normalfont\small
\textsuperscript{4}University of T\"ubingen \quad
\textsuperscript{5}University of Oxford \quad
\textsuperscript{6}ELLIS Institute T\"ubingen
}
\\[0.8em]
{\normalfont\small
\textsuperscript{*}Equal first-author contribution.
\quad
\textsuperscript{\textdagger}Equal second-author contribution.
}
\end{tabular}%
}
}

\begin{document}

\maketitle

\begin{abstract}

Whether navigating a building, operating a robot, or playing a game, an agent that acts effectively in an environment must first learn an internal model of how that environment works. Partially-observable Markov decision processes (POMDPs) provide a flexible modeling class for such internal world models, but learning them from observation-action trajectories alone is challenging and typically requires extensive environment interaction. We ask whether language-model priors can reduce costly interaction by leveraging prior knowledge, and introduce \mbox{\emph{Pinductor}} (POMDP-inductor): an LLM proposes candidate POMDP models from a few observation-action trajectories and iteratively refines them to optimize a belief-based likelihood score. Despite relying on strictly less information, \mbox{\emph{Pinductor}} achieves comparable performance and sample efficiency to LLM-based POMDP learning methods with privileged access to hidden states. Moreover, it is up to 330 times more sample-efficient than standard POMDP baselines such as DreamerV3, at comparable computational cost. Further results show that performance scales with LLM capability and degrades gracefully as semantic information about the environment is withheld. Together, these results position language-model priors as a practical tool for sample-efficient world-model learning under partial observability, and a step toward generalist agents in real-world environments. Code: \url{https://github.com/atomresearch/pinductor}.
\end{abstract}


\begin{figure}[h]
  \centering
    \vspace{-0.4cm}
  \includegraphics[width=0.8\linewidth]{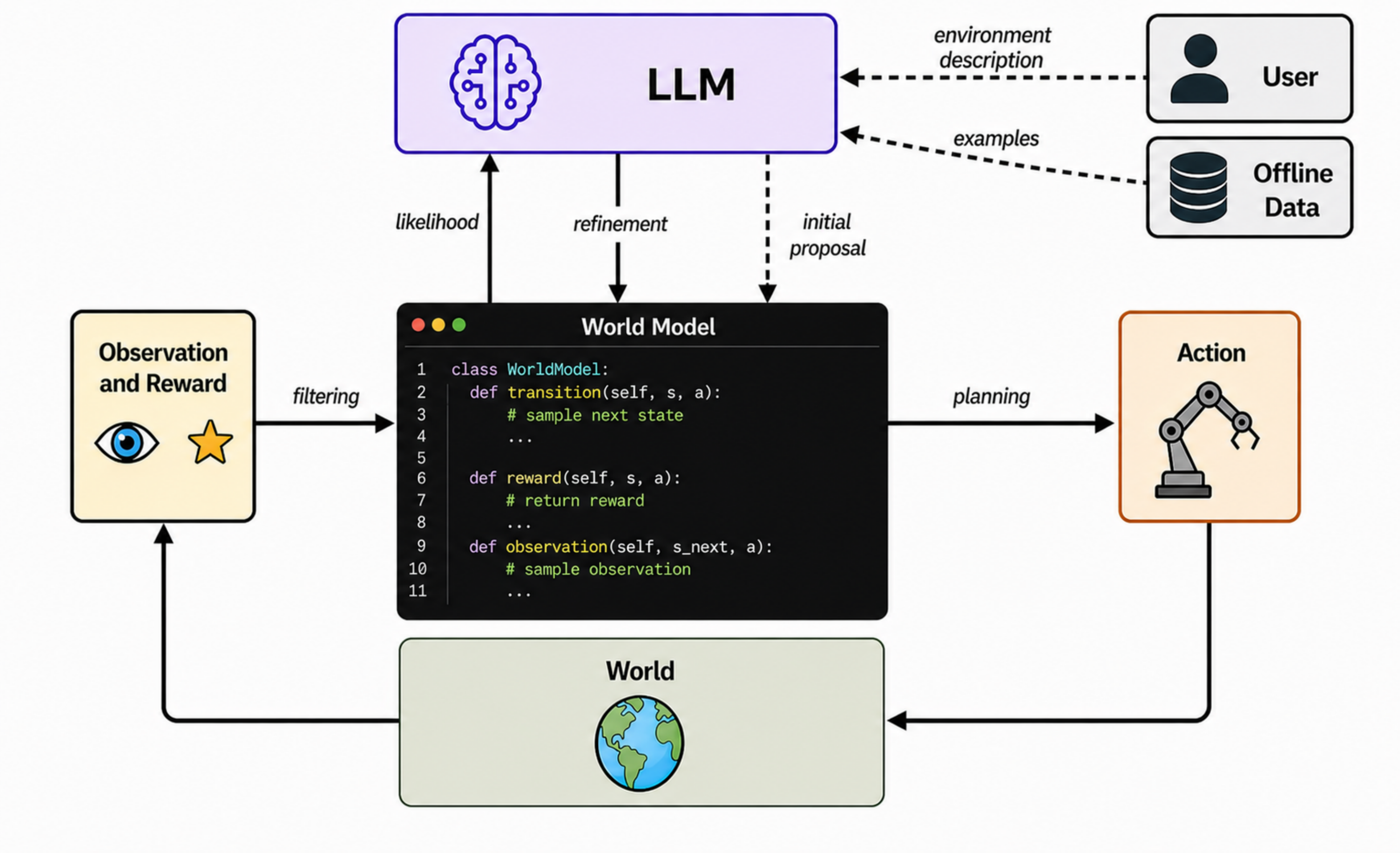}
  \caption{\textbf{\emph{Pinductor} architecture overview.}
Given a small set of offline observation-action trajectories and an environment description, an LLM proposes a POMDP world model in code (dashed arrows). The resulting model is used for filtering and planning during environment interaction, and is periodically refined by the LLM to optimize a belief-based likelihood objective (solid arrows).}
  \label{fig:grabs}
  \vspace{-0.6cm}
\end{figure}

\section{Introduction}

Consider an agent dropped into an unfamiliar building, deployed on a new robot, or placed in an unseen game. Before it can act competently, it must either explicitly or implicitly construct an internal model of how the world responds to its actions: what the hidden state is, how the environment evolves, what it can expect to observe, and what rewards its actions yield. Building such a model from first-person experience, rather than from a handwritten specification, is a long-standing problem in reinforcement learning and embodied AI \citep{sutton2018reinforcement, ha2018world, hafner2025training}.

When the world is not fully observable, a natural modeling formalism is the partially observable Markov decision process (POMDP), which explicitly represents hidden states, transitions, observations, and rewards \citep{astromOptimalControl1965,kaelbling1998planning}. POMDPs provide a flexible modeling class for internal world models under partial observability, but learning them is practically demanding: classical approaches, including tabular estimators, predictive-state representations, and deep recurrent latent models, typically require large numbers of environment interactions, strong structural assumptions, or both \citep{mossel2005learning, singh2004predictive, hafner2025training}. Directly specifying POMDPs by hand, meanwhile, requires careful engineering and precise knowledge of available solvers \citep{shani2013survey}.

A recent line of work asks whether large language models (LLMs) can substitute for some of this interaction by providing strong priors over world dynamics. Rather than using the LLM itself as a simulator, which can be slow, expensive, and prone to inconsistency \citep{hao2023reasoning, dainese2024generating}, these methods use the LLM to write an executable world model in code and then refine that code against observed trajectories \citep{tang2024worldcoder, dainese2024generating, piriyakulkij2025poe, liang2025visualpredicator}. Code-structured world models inherit the LLM's prior knowledge about environments while remaining precise, auditable, and cheap to query at planning time.

Almost all of this work, however, makes a critical simplifying assumption: that the latent state is available for learning. WorldCoder \citep{tang2024worldcoder}, GIF-MCTS \citep{dainese2024generating}, and most program-synthesis methods assume fully observable environments. The closest precursor to our work, POMDP Coder \citep{curtis2025llmguided}, extends LLM-guided program induction to POMDPs, but still relies on post-hoc full observability: after each episode, the agent is given access to the intermediate ground-truth states that it could not observe at decision time. In many settings, such as robots operating in cluttered or human-occupied spaces, or agents playing imperfect-information games, neither online nor post-hoc state access is available. Methods that require this privileged signal, therefore, cannot be applied. Whether LLM priors can still operate without ground-truth state supervision is the open question we address in this work.

To address this question, we introduce \emph{Pinductor} (POMDP-inductor), a method that induces executable POMDP world models from observation--action--reward trajectories alone. \emph{Pinductor} uses an LLM to propose candidate programs for the transition, observation, reward, and initial-state distributions, and then iteratively refines them using a belief-based likelihood score. Observation predictions are converted into soft likelihoods through a distance kernel, and candidate models are scored by their likelihood (expected under the beliefs induced by their own filtering dynamics). Because this objective is computed from observations and self-induced beliefs rather than from privileged states, \emph{Pinductor} applies in the strict POMDP setting where post-hoc state supervision is unavailable, given a symbolic state schema (i.e., a structured symbolic domain) and a task description. A visual overview of the method is provided in Fig.~\ref{fig:grabs}.

We evaluate \emph{Pinductor} on MiniGrid environments \citep{chevalier2018minimalistic}, on partially observable MineSweeper \citep{morad2023popgym} and Blocks World \citep{slaney2001blocks} games of varying complexity. Despite using strictly less information than recent LLM-based methods, \emph{Pinductor} matches their sample efficiency and performance, while largely outperforming standard tabular POMDP baselines, which struggle to learn from few trajectories. It is also competitive with deep RL baselines such as DreamerV3 and PPO-GRU that interact for 330x more online episodes. Additional experiments show that performance scales with LLM capability and degrades when semantic information about the environment is withheld, indicating that the method relies heavily on the LLM and availability of semantic information. Together, these results suggest that language-model priors can enable sample-efficient world-model learning without privileged state access, broadening the reach of existing methods to the partially observable settings that characterize the real world.

We summarize the paper's contributions as follows:
\begin{enumerate}[leftmargin=*, align=left]
    \item \textbf{Observation-only POMDP induction.}
    We show that LLM priors are sufficient to induce executable POMDP world models from observation--action--reward trajectories alone, without access to ground-truth latent states at training or inference time.

    \item \textbf{Belief-based model scoring.}
    We introduce a kernel-based likelihood objective that scores candidate POMDP programs under their filtered belief distributions, providing a per-step repair signal computable from observation--action--reward trajectories alone and optimized by the LLM.

    \item \textbf{End-to-end validation.}
    Across five partially observable MiniGrid tasks, a Blocks World domain, and three MineSweeper difficulties, we show that \emph{Pinductor}
    matches the reward and sample efficiency of a privileged-state LLM baseline,
    outperforms non-LLM baselines at equal interaction budget, is competitive with deep RL baselines trained for 330x more online episodes, and induces belief states that become increasingly
    concentrated on the true latent state during planning.
    
\end{enumerate}

\section{Related work}

\paragraph{LLM-guided POMDP and model induction}
The closest work to ours is POMDP Coder~\citep{curtis2025llmguided}, which uses an LLM to propose and repair executable POMDP components using a coverage objective. However, it assumes privileged access to hidden states during training, both in demonstrations and through post-hoc full observability during online interaction. Closest in setting, the closed-deck variant of Code World Models~\citep{lehrach2026code} also learns from observations and actions alone, but with deterministic dynamics whose sampled hidden histories must reproduce the observations exactly, rather than stochastic models scored by a soft belief-based likelihood. Other LLM-based approaches instead rely on extensive natural-language task descriptions to construct POMDP models~\citep{light2024pianist}, or use LLMs directly as planners rather than as model learners~\citep{tang2025tru}. In contrast, \emph{Pinductor} induces executable POMDP models from observation--action--reward trajectories and a minimal environment API, without access to hidden states.

\paragraph{LLMs for code-based world models}
Our approach fits within the broader paradigm of verbalized machine learning~\citep{xiao2025verbalized}, where models are represented in the LLM's token space, for example as executable code, and refined through language-based feedback. Prior work has used LLMs to generate code for reinforcement learning~\citep{liang2023code}, fully observable world models~\citep{tang2024worldcoder,ahmed2025synthesizing,lymperopoulos2026cassandraprogrammaticprobabilisticlearning}, and algorithms more generally~\citep{novikov2025alphaevolve}. \emph{Pinductor} differs by learning latent state generative models from trajectories in which the underlying state is never observed.

\paragraph{Learning in POMDPs}
A large body of work studies learning and planning under partial observability, including Bayes-adaptive methods~\citep{ross2007bayes,katt2019bayesian}, spectral and variational approaches~\citep{azizzadenesheli2016reinforcement,tschiatschek2018variational}, active-learning methods~\citep{bacci2021active}, and neural latent world models~\citep{Hafner2023ARXIV_Mastering_Diverse_Domains}. These methods can be effective, but typically require substantial data, specified model classes, tractable inference, or strong structural assumptions. \emph{Pinductor} is complementary: given a compact latent state space and code interface, it uses LLM priors to search for explicit, auditable POMDP programs from few trajectories.

\paragraph{Theory-based reinforcement learning}
Theory-based RL uses programmatic, intuitive theories expressed as code to support sample-efficient planning~\citep{tsividis2021human,pouncy2022inductive}. These methods demonstrate the value of structured executable world models, but to date search within hand-designed hypothesis spaces whose primitives are tailored to the benchmark environments. \emph{Pinductor} instead searches over executable POMDP components using an LLM prior, allowing it to induce models from sparse partially observed experience without a domain-specific theory language.

\section{Background on POMDPs}

A partially observable Markov decision process (POMDP) models sequential
decision-making when the agent cannot directly observe the underlying state.
In our setting, a POMDP is a tuple
\begin{equation}
  \label{eq: pomdp}
  (\mathcal{S},\, \mathcal{A},\, \mathcal{O},\, T,\, O,\, R,\, \rho_0,\, \gamma)
\end{equation}
where $\mathcal{S}$, $\mathcal{A}$, and $\mathcal{O}$ are finite sets of states,
actions, and observations, respectively, cf.~\citep{sutton2018reinforcement}. The transition model
$T(s,a,s') = P(s' \mid s,a)$ gives the probability of moving to state $s'$
after taking action $a$ in state $s$, while the observation model
$O(s',a,o) = P(o \mid s',a)$ gives the probability of observing $o$ after taking action $a$ and arriving in $s'$. The reward function $R(s,a,s')$ specifies the immediate
reward received for transitioning from $s$ to $s'$ via action $a$. The initial state distribution $\rho_0(s) = P(s_0 = s)$ defines the probability of starting in state $s \in \mathcal{S}$, and the discount factor $\gamma \in [0, 1)$ determines the present value of future rewards.

Because observations are generally non-Markovian, an agent cannot condition
only on its current observation to predict the future or choose an optimal
action. Instead, it maintains a \emph{belief state} $Q$, an approximate posterior probability
distribution over latent states. Given an action $a$ and a new observation $o$, the belief is updated by Bayes' rule:
\begin{equation}
  \label{eq:bayes}
  Q'(s') \propto O(s', a, o) \sum_{s \in \mathcal S} T(s, a, s') Q(s).
\end{equation}
This filtering update assumes access to an initial state distribution with which to initialize the belief, as well as a transition model and an observation model. In our setting, these models are not given and must be learned from experience.

For decision-making, the agent must also estimate a reward model. Given learned POMDP components, planning can then be performed in belief space using standard offline methods such as value iteration, online Monte Carlo tree search methods such as POMCP, or deterministic approximations such as DA*. In this paper, we focus on learning the model components: inducing the transition, observation, reward, and initial-state distributions required for belief filtering and planning from partially observed trajectories.

\section{Methodology}
\label{sec:methods}

\begin{figure}[h]
  \centering
  \includegraphics[width=\linewidth]{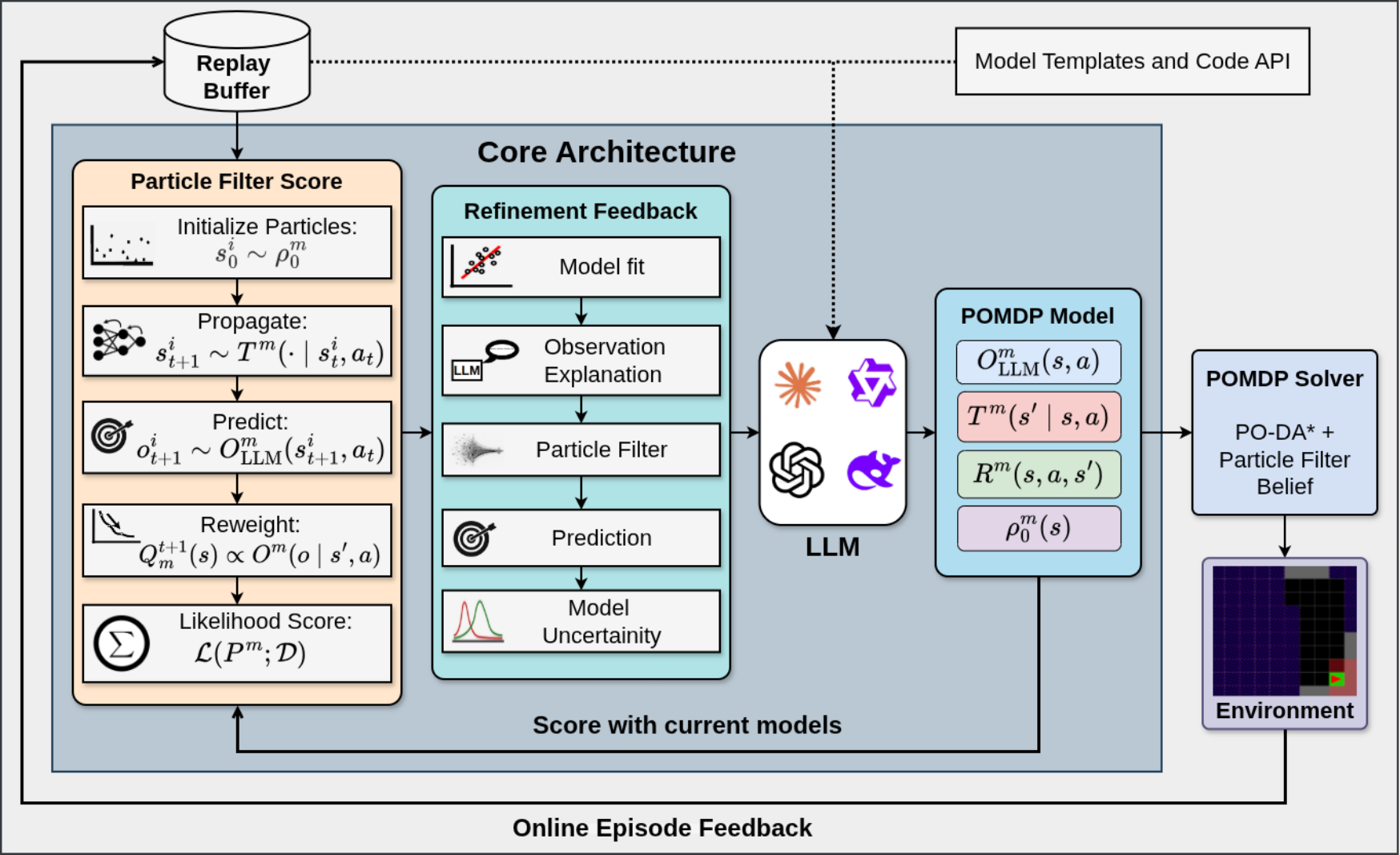}
  \caption{\textbf{\emph{Pinductor} pipeline.} State-free trajectories, model templates, and a code API prompt the LLM to propose candidate components \(\rho_0^m(s)\), \(T^m(s'\mid s,a)\), \(O^m_{\text{LLM}}(s,a)\), and \(R^m(s,a,s')\). For filtering, \(O^m_{\text{LLM}}\) is softened into the kernel likelihood \(O^m(o\mid s,a)\). A particle filter propagates belief states, reweights them by observation compatibility, and contributes to likelihood scores and diagnostics for LLM refinement. The selected model is used for planning under partial observability, and the loop repeats.}
  \label{fig:architecture}
\end{figure}

\emph{Pinductor} learns executable POMDP models from trajectories containing actions, observations, rewards, and termination signals, but no hidden-state labels. It follows a generate--evaluate--refine--plan structure: an LLM proposes executable model components, particle filtering evaluates whether the induced latent beliefs explain the observed trajectories, diagnostic feedback guides model refinement, and the selected model is used for belief-space planning. Unlike state-supervised model induction, \emph{Pinductor} never compares predicted states to ground-truth hidden states. Instead, it scores whether latent rollouts predict observations compatible with the data. Fig.~\ref{fig:grabs} gives an overview; Fig.~\ref{fig:architecture} illustrates the full pipeline; Alg.~\ref{alg:belief-llm-refinement} gives the pseudo-code. Additional methodological details and a discussion of component roles are provided in Appendix~\ref{app:method-details}.

\subsection{Problem formulation}

Consider POMDP models defined in \eqref{eq: pomdp}, where the state \(s_t\in\mathcal{S}\) is latent, the agent takes actions \(a_t\in\mathcal{A}\), and receives observations \(o_t\in\mathcal{O}\), rewards \(r_t\), and termination signals \(d_t\). The learner is given a dataset $\mathcal{D}=\{\tau_n\}_{n=1}^N$ of \(N\) state-free trajectories, collected from offline and/or online interaction $\tau=(o_0^{\tau},a_0^{\tau},r_1^{\tau},d_1^{\tau},o_1^{\tau},\ldots,
a_{H_\tau-1}^{\tau},r_{H_\tau}^{\tau},d_{H_\tau}^{\tau},o_{H_\tau}^{\tau})$
where \(H_\tau\) denotes the horizon of trajectory \(\tau\); thus, \(H_\tau\) may vary across trajectories. The hidden states \(s_{0:H_\tau}^{\tau}\) are never observed.

\emph{The problem:} Given $(\mathcal{S},\mathcal{A},\mathcal{O},\gamma)$, the goal is to learn a model
\begin{equation}
  \label{eq:candidate-model}
  m = (\rho_0^m,T^m,O^m,R^m),
\end{equation}
that explains the realized trajectories and supports downstream tasks. We denote by $P^m=(T^m,O^m,\rho_0^m)$ the reward-free probabilistic model, which we optimize with a probabilistic score.

\subsection{Model proposal}

The LLM receives a natural-language task description, a small offline dataset \(\mathcal{D}\), and a code API that fixes a typed encoding of $(\mathcal{S},\mathcal{A},\mathcal{O})$, which we call the \emph{schema}: each space is written as a product of named, typed fields (e.g.\ grid, agent position, direction, carried object) whose values carry labels (object types, action names). It receives no state values, model code, reward or termination logic, action effects, or initial-state distribution, and within the schema it chooses its own representation (``observation-only'' thus refers to the training signal, not to the absence of a symbolic interface). Renaming the schema's classes to generic identifiers has no detectable effect on the induced programs (Table~\ref{tab:scaffold}). The LLM then generates an executable candidate $m$ defined in \eqref{eq:candidate-model}. We will denote by $O^m_{\text{LLM}}$ the observation model generated by the LLM, to distinguish it from the final observation model $O^m$ we will use.

In our experiments, the offline dataset contains \(N=10\) manually collected trajectories chosen to cover informative parts of each task. Details about data collection can be found in Appendix~\ref{app:experimental-details}.

\subsection{Belief-based model evaluation}
\label{sec: model eval}

For each candidate \(m\), \emph{Pinductor} evaluates whether predicted observations from latent rollouts generated
by $P^m$ can explain the realized observation-action trajectory. A set of \(K\) particles indexed by \(i\) is sampled from \(\rho_0^m\), propagated through \(T^m\) under realized actions by sampling, and scored by comparing a sampled observation
\(o_{t+1}^i \sim O^m_{\text{LLM}}(s_{t+1}^i,a_t)\) against the realized observation
\(o_{t+1}\).

For MiniGrid observations $o=(g,\theta,c)$, we compare using a distance over the visible part of the grid \(g\), agent direction \(\theta\), and carried object \(c\):
\begin{equation}
\label{eq:obs-distance}
d_{\mathrm{obs}}(o',o)
=
d_{\mathrm{grid}}(g',g)
+
\lambda_{\mathrm{dir}}\mathbf{1}[\theta'\neq\theta]
+
\lambda_{\mathrm{carry}}\mathbf{1}[c'\neq c].
\end{equation}
In such environments, the agent can pick up and drop objects; \(c\) denotes the object, if any, currently carried by the agent at the corresponding timestep. This distance is used to soften the (usually deterministic) LLM-generated code observation model $O^m_{\text{LLM}}$, to furnish a soft observation model where realized observations have positive probability
\begin{equation}
\label{eq:obs-kernel}
  O^m(o_{t+1}\mid s_{t+1}^i,a_t)\propto\exp\!\left(
-d_{\mathrm{obs}}(o_{t+1}^i,o_{t+1})/\kappa
\right),
\end{equation}
where \(\kappa>0\) is a parameter. We can interpret $O^m_{\text{LLM}}$ as the mode of $O^m$ and $\kappa$ as the variance, and the sampling step \(o_{t+1}^i \sim O^m_{\text{LLM}}(s_{t+1}^i,a_t)\) as sampling the most likely observation under $O^m$. The full grid-distance definition and constants are given in
Appendix~\ref{app:kernel-likelihood}. The resulting particle-filtered
posterior belief is the distribution of propagated samples $s_{t+1}^i$ reweighted by their likelihood
\begin{equation}
\label{eq:particle-belief}
Q_{t+1}^m(s)
\propto
\sum_{i=1}^K O^m(o_{t+1}\mid s_{t+1}^i,a_t)\,
\delta_{s_{t+1}^i}(s),
\end{equation}
which is exactly the particle-filtered analog of the Bayesian update in \eqref{eq:bayes}. Details on particle filtering are available in Appendix~\ref{app:particle-filtering}. This yields the expected log likelihood score
\begin{equation}
\label{eq:kernel-score}
\mathcal{L}(P^m;\mathcal{D})
=\frac{1}{|\mathcal{D}|}\sum_{\tau\in\mathcal{D}}
\sum_{t=0}^{H_\tau-1}
\mathbb{E}_{s_{t+1}\sim Q_{t+1}^m}
\left[\log O^m(o_{t+1}\mid s_{t+1},a_t)
\right] 
\end{equation}
The inner summation runs from \(t=0\) to \(H_\tau-1\) over the trajectory \(\tau\), where observations \(o_{t+1}\), \(a_t\) and \(Q^m_{t+1}\) implicitly depend on $\tau$. The initial observation \(o_0\) is therefore not directly scored, since it has no preceding action and no candidate prediction to compare against; the initial-state distribution \(\rho_0^m\) is evaluated indirectly through the first propagated observation $o_1\sim O^m_{\text{LLM}}(s_1,a_0)$, from propagated states \(s_1 \sim T^m(s_0, a_0)\) and \(s_0 \sim \rho_0^m\). This score evaluates $P^m$'s fit of observation-action sequences only; reward and termination errors influence model selection indirectly through the LLM's local diagnostics (see Appendix~\ref{app:prompts} for examples). In practice, each trajectory's score is clipped below at \(-100\) to avoid outliers from particle collapse, and on MineSweeper a reward-prediction term is added (Appendix~\ref{app:kernel-likelihood}).

\begin{algorithm}[t]
\caption{Belief-based Pinductor refinement}
\label{alg:belief-llm-refinement}
\begin{algorithmic}[1]
\Require Task description \(c\), API \(\mathcal{I}\), trajectories \(\mathcal{D}\), rounds \(J\), candidates \(M\), UCB constant \(c_{\mathrm{ucb}}\)
\State \(\mathcal{P} \gets \emptyset\), \(\mathcal{C} \gets \emptyset\)
\State \(F_0 \gets \textsc{InitialPrompt}(c,\mathcal{I},\mathcal{D})\)
\Comment{Appendix~\ref{app:prompts}}
\For{\(j=1,\ldots,J\)}
    \State \(p \gets \textsc{UCB1Select}(\mathcal{P},c_{\mathrm{ucb}})\)
    \Comment{\eqref{eq:app-ucb1}}
    \For{\(k=1,\ldots,M\)}
        \State \(F \gets F_0\) if \(p=\emptyset\) else \(\textsc{RefinementPrompt}(c,\mathcal{I},p,\mathcal{L}_p,D_p,Q_p)\)
        \Comment{Appendix~\ref{app:prompts}}
        \State \(m_{j,k} \gets \textsc{LLM}(F)\)
        \State \(\mathcal{L}_{j,k}, B_{j,k}, D_{j,k} \gets \textsc{ParticleFilterKernelScore}(m_{j,k},\mathcal{D})\)
        \Comment{\eqref{eq:obs-distance}--\eqref{eq:kernel-score}}
        \State \(\mathcal{C} \gets \mathcal{C}\cup\{T^{m_{j,k}}\}\)
        \Comment{Appendix~\ref{app:refinement-details}}
        \State \(Q_{j,k} \gets \textsc{QBCDisagreement}(\mathcal{C},B_{j,k},\mathcal{D})\)
        \Comment{\eqref{eq:qbc-ve}}
        \State \(\mathcal{P} \gets \mathcal{P}\cup\{m_{j,k}\}\)
    \EndFor
\EndFor
\State \Return \(m^\star \sim \textsc{NearBest}(\mathcal{P})\) \Comment{\eqref{eq:near-best-set}--\eqref{eq:final-selector}}
\end{algorithmic}
\end{algorithm}

\subsection{Feedback and refinement}

After scoring a candidate, Pinductor asks the LLM to revise its code. The refinement prompt contains the candidate's code, its score $\mathcal{L}_j=\mathcal{L}(P^{m_j};\mathcal{D})$, and a summary of concrete failure cases (execution errors, trajectory segments with large observation distance, reward or termination mismatches) that point the LLM to the code regions needing repair. Since the true transition is unavailable, the prompt also flags transition contexts on which the transition models generated so far \(\mathcal{C}_j=\{T^{m_1},\ldots,T^{m_j}\}\) disagree. We measure disagreement using the normalized vote entropy score
\begin{equation}
\label{eq:qbc-ve}
\operatorname{VE}(s,a)
=
-\frac{1}{\log |\mathcal{C}_j|}
\sum_y
\frac{V_{s,a}(y)}{|\mathcal{C}_j|}
\log
\frac{V_{s,a}(y)}{|\mathcal{C}_j|},
\end{equation}
for a belief particle \(s\) and action \(a\), where \(V_{s,a}(y)\) is the number of committee members predicting next state \(y\). This score is computed for all belief particles visited during filtering; the highest-entropy contexts are added to the prompt with their observation context and action.\footnote{Candidates that use the schema differently remain comparable: the score \eqref{eq:kernel-score} compares predicted with realized observations only, and the vote applies every model to the same particles.}

Refinement-by-execution (REx) repeats this revise procedure over $J$ rounds: at each round, a parent candidate is selected by UCB1 \citep{auer2002finite} from a persistent candidate tree, the LLM returns $M$ revised versions of it, and each is scored and added to the tree. After all rounds, the final model is sampled from the near-best set of candidates whose scores lie within one standard deviation of the best, avoiding over-commitment to small score differences. The UCB1 rule, the final selector, and the diagnostics are detailed in Appendix~\ref{app:refinement-details}. Appendix~\ref{app:search-correspondence} relates the loop to bandit-guided and evolutionary program search such as FunSearch \citep{romera2024funsearch}.

\subsection{Planning and online interaction}

The selected model \(m^\star\) is used for belief-space planning. During the episode, the agent maintains a particle belief \begin{equation}
\label{eq:belief-state}
Q_t(s)
\approx
P^{m^\star}(s_t=s\mid o_{0:t},a_{0:t-1}),
\end{equation}
updated using the same distance-kernel observation likelihood as in \eqref{eq:obs-kernel}. Actions are chosen by a POMDP planner over this belief state; in our experiments, we use the proposed planner in \citet{curtis2025llmguided}, which is an A*-style belief-space planner. After execution, newly collected trajectories are appended to the dataset, and a fresh REx round is triggered to continue refining the model online.

\section{Experiments}
\label{sec:experiments}


\subsection{Experimental setup}
\label{subsec:experimental_setup}

\paragraph{Environments}
We evaluate on five partially observable MiniGrid tasks \citep{chevalier2018minimalistic} of increasing difficulty (\textsc{Empty}, \textsc{Corners}, \textsc{Lava}, \textsc{Four Rooms}, \textsc{Unlock}), and on two suites that are not grid navigation: \textsc{Block}, our own partially observable Blocks World \citep{slaney2001blocks}, in which the agent sees only the stack at its current column and the block it holds, and \textsc{MineSweeper} from POPGym \citep{morad2023popgym} at three difficulties, in which the agent observes only local mine counts. The pipeline ran unchanged on all three suites, with only the natural-language description and the few-line observation distance rewritten (Appendix~\ref{app:envs}).

\paragraph{Baselines}
We compare against POMDP Coder \citep{curtis2025llmguided}, which has privileged access to hidden states during learning and otherwise shares our pipeline, LLM, and evaluation seeds; this comparison tests whether hidden-state supervision can be replaced by belief-based feedback. We also compare against a tabular baseline (lookup-table models estimated from the same demonstrations, also granted hidden states), a random policy, and four learned methods that induce no POMDP: \emph{Direct LLM}, which acts directly from the observation--action history, fully (FO) or partially (PO) observed; \emph{PPO-GRU} \citep{schulman2017proximal,Nippogru2021}; and \emph{DreamerV3} \citep{Hafner2023ARXIV_Mastering_Diverse_Domains} with its replay seeded by our demonstrations (the from-scratch variant is in Table~\ref{tab:full_results}). Two references bound the results: \emph{True models, same planner} (the handcrafted models of \citet{curtis2025llmguided} in our own filter and planner) and \emph{Optimal} (true state, perfect planning). All learning methods receive the same demonstrations; the LLM-based and tabular methods interact for 3 online episodes, the RL baselines for 1{,}000; the information available to each method is listed in Appendix~\ref{app:baselines}. We report mean episode reward and win rate (reward $>0.05$) over 10 seeds of 3 evaluation episodes, and track belief entropy and accuracy relative to the true hidden state as diagnostics only (Fig.~\ref{fig:belief_combined}). In the figures, colours group methods by family: blue for \emph{Pinductor}, orange for LLM-based baselines, green for RL, mauve for count-based baselines, grey for the references.

\paragraph{Metrics}
We report average episode reward as the main measure of downstream decision-making performance, and win rate as a complementary success metric that is less sensitive to reward discounting and episode length. To test whether the learned models perform useful inference under partial-observation, we also track belief entropy and belief accuracy relative to the true hidden state during evaluation. These belief metrics are not used as supervision; they are diagnostics for whether the learned model maintains useful latent-state information.

\paragraph{Protocol and implementation details}
To isolate the effect of removing hidden-state supervision, we match the hyperparameters of \emph{Pinductor} and \citet{curtis2025llmguided} wherever possible and use the same LLM, Qwen 3.6 Plus \citep{qwen36plus}, for both (the LLM ablation also uses Qwen 3 14B \citep{qwen3technicalreport} and Claude Opus 4.7 \citep{anthropic2026claudeopus47}). Pipeline, sampling settings, seeds, demonstrations, refinement budget, planner, and tuning protocol are shared. The two remaining differences are the manipulated variable itself (state-supervised versus belief-based scoring) and the planner's exploration coefficient, tuned for both under the same protocol and optimal at 0 for POMDP Coder and 1 for \emph{Pinductor}, since an agent must explore to disambiguate its belief when it cannot access the state, but need not when it can. Hyperparameters, experimental variants, and implementation details are given in Appendices~\ref{app:hyperparameters}, \ref{app:experimental-details}, and \ref{app:implementation}.

\begin{figure}[!t]
    \centering
    \includegraphics[width=\linewidth]{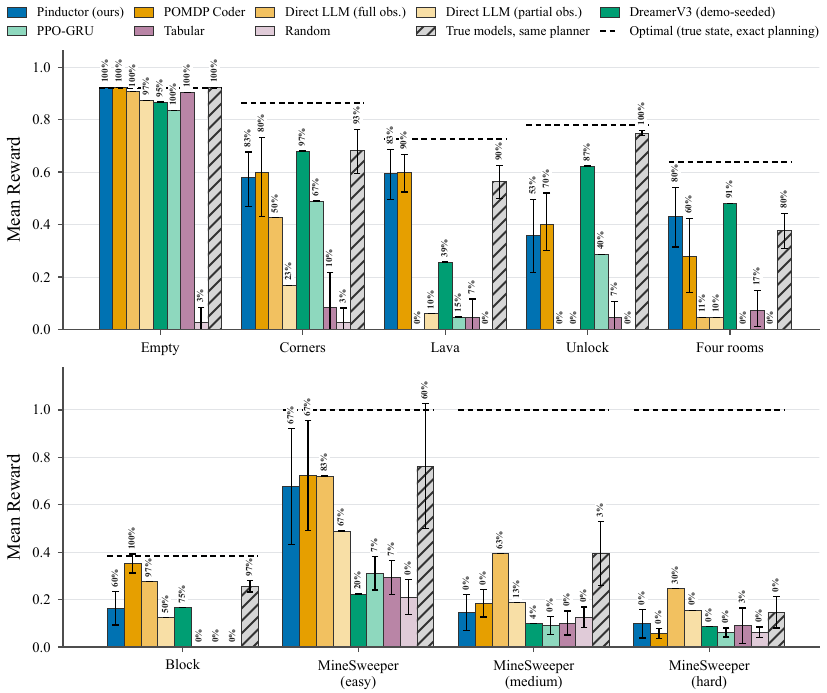}
    \caption{\textbf{Main task performance.}
    Mean episode reward (bars) and win rate (labels) on five MiniGrid environments (top), Block and MineSweeper (bottom; returns normalised to $[0,1]$); dashed lines mark the Optimal reference. \emph{Pinductor} (blue) is comparable to POMDP Coder despite never accessing hidden states, outperforms the non-LLM baselines given the same interaction budget, and is competitive with RL baselines that interact for 330x more online episodes. Error bars are 95\% confidence intervals; arms without error bars are reported as means only (Appendix~\ref{app:randomness-control}).}
    \label{fig:reward_win_rate}
\end{figure}
\subsection{Main Results}

We evaluate whether \emph{Pinductor} can replace hidden-state supervision with belief-based feedback for LLM POMDP induction and downstream task performance. The central question is not only whether the learned models lead to high downstream reward but also quantifying sample efficiency and useful latent-state inference required for downstream planning under partial observability. We therefore evaluate task performance, belief quality, and sample efficiency across 5 MiniGrid environments of varying complexity, and task performance on Block and MineSweeper. In Appendix~\ref{app:add-experiments}, we additionally report performance in stochastic MiniGrid variants, and sensitivity to the choice of LLM and to the prompt supplied by the user.

\emph{Pinductor} achieves performance comparable to the state-of-the-art LLM baseline while using strictly less information for model induction. As shown in Fig.~\ref{fig:reward_win_rate}, \emph{Pinductor} also substantially outperforms the non-LLM baselines across the tested MiniGrid environments, including the tabular baseline, indicating that the learned models support effective downstream planning. The apparent differences between \emph{Pinductor} and the privileged-state baseline are small relative to the variability across environments and seeds, especially given the sparse-reward nature of these tasks. We therefore interpret these results as evidence that \emph{Pinductor} can match the performance of a state-access model-induction baseline in this setting, without relying on ground-truth latent states.

\paragraph{Beyond MiniGrid} The bottom row of Fig.~\ref{fig:reward_win_rate} reports Block and MineSweeper, where the same pipeline ran unchanged except for the natural-language description and the few-line observation distance. On MineSweeper (easy), \emph{Pinductor} matches the privileged baseline (0.68 against 0.72 mean reward, both at a 67\% win rate); on medium and hard, every method collapses, the true-model reference included: winning requires 30 or 54 consecutive safe clicks over a combinatorially large hidden layout, so the limit there is belief filtering and planning over a long horizon rather than model induction (Appendix~\ref{app:new-envs}). Direct LLM (FO) fares better because its state reveals the mine layout, which the model-based methods must infer. On Block, the methods with privileged access to the hidden state clearly dominate those without: POMDP Coder reaches 0.35 (100\% win rate) while \emph{Pinductor} reaches 0.16 (60\%). Part of this gap comes from belief filtering: the true models run through our own filter and planner still win 97\% of episodes but reach only 0.26, as disambiguating the state from observations takes longer. The rest is a limitation of model induction under partial observability: Appendix~\ref{app:model-audit} traces it to a single transition convention (the behaviour of the gripper at the edge of the row) that the demonstrations leave unidentified and that online interaction does not reliably correct.

\paragraph{Comparison with reinforcement learning baselines} With three online episodes, \emph{Pinductor} is comparable to demonstration-seeded DreamerV3 after a thousand (ahead on Lava and MineSweeper easy, behind on Unlock and Corners, within 0.06 elsewhere), and it is above PPO-GRU, DreamerV3 trained from scratch (Table~\ref{tab:full_results}), and the tabular baseline on every task but Corners. Direct LLM planning without a model is competitive only on short-horizon tasks and drops to near zero on Lava, Unlock, and Four rooms under partial observability. The full pipeline, not the prior alone, is responsible for these results: a single induction call reaches a win rate of 8\% on Lava and 5\% on Four rooms, against 83\% and 80\% for the full pipeline (Table~\ref{tab:single_call}). Finally, the gap between Optimal and the true models run through our own filter and planner (Table~\ref{tab:oracle_vs_gt}) shows how much of the remaining headroom is due to partial observability and finite planning rather than to model induction.

\begin{figure}[!t]
    \centering
    
    \begin{subfigure}[t]{0.49\linewidth}
        \centering
        \includegraphics[width=\linewidth]{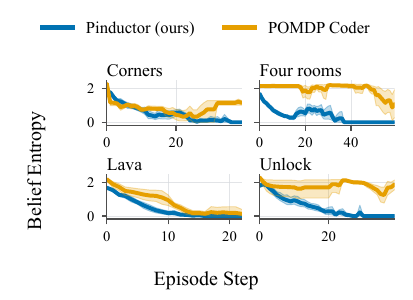}
        \caption{Belief uncertainty.}
        \label{fig:belief_entropy}
    \end{subfigure}
    \hfill
    \begin{subfigure}[t]{0.49\linewidth}
        \centering
        \includegraphics[width=\linewidth]{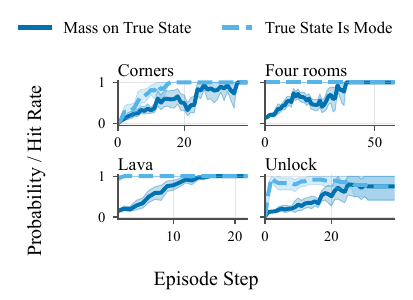}
        \caption{Belief accuracy.}
        \label{fig:belief_true_state}
    \end{subfigure}
    
    \caption{\textbf{Belief dynamics.}
    Left: belief entropy over episode steps. Right: posterior mass on the true hidden state and MAP-belief accuracy. Beliefs become more concentrated and accurate as observations accumulate. \emph{Pinductor} shows smoother entropy reduction due to graded observation-distance reweighting, unlike the privileged state-access baseline's hard exact matching.}
    \label{fig:belief_combined}
\end{figure}

\emph{Pinductor} also maintains meaningful belief states during planning. Its belief becomes more concentrated over the course of an episode and assigns increasing mass to the true hidden state as observations accumulate, which is visible on Fig.~\ref{fig:belief_combined}. This is notable because \emph{Pinductor} receives less supervision during model induction than the state-access baseline, yet still supports the filtering computation needed for planning under partial observability. The smoother entropy reduction comes from our observation-distance kernel (\ref{eq:obs-distance}), which reweights particles by graded observation similarity rather than using the hard exact-match criterion of the state-access baseline. These diagnostics suggest that \emph{Pinductor} is not merely producing successful policies, but learning candidate POMDPs that support informative posterior updates.

\begin{figure}[!t]
    \centering
    \includegraphics[width=1\linewidth]{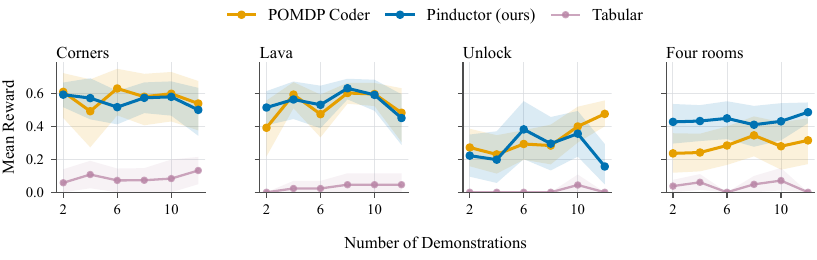}
    \caption{\textbf{Offline sample efficiency.}
    Mean discounted episode reward over the three online episodes, as a function of the number of offline demonstrations used for initial model induction (Appendix~\ref{app:offline-sweep}). \emph{Pinductor} performs comparably to the state-access LLM baseline despite receiving less information per sample.}
    \label{fig:offline_sample_eff}
\end{figure}

The offline sample-efficiency results in Fig.~\ref{fig:offline_sample_eff} suggest that \emph{Pinductor} does not need many demonstrations to become effective. Even though each trajectory provides less information than in the state-access LLM baseline, \emph{Pinductor} achieves comparable performance at the same sample counts. Moreover, both LLM-based methods reach relatively strong performance after only a few trajectories, suggesting that LLM priors help identify plausible environment structure quickly. In particular, the offline data allows LLM-based methods to propose useful candidate programs, as opposed to constituting enough data to fully identify the transition, observation, and reward models from scratch.

Taken together, these results suggest that LLM-guided POMDP induction can succeed without hidden-state supervision in the tested domains. \emph{Pinductor} performs comparably to the privileged-state access LLM baseline, while retaining the advantages of explicit POMDP models: executable components, belief updates, and downstream planning under partial observability. This indicates that LLM priors, grounded by observation-level feedback, can induce useful models from a small number of offline trajectories.





\section{Conclusions and Limitations}
\label{sec:conclusion}
\label{sec:limitations}

Can language-model priors substitute for state supervision when learning
POMDPs? Our results suggest they can. \emph{Pinductor} induces executable POMDP world models from a belief-based likelihood signal, without access to hidden states. Despite this, our method matches privileged-state access LLM baselines, significantly outperforms tabular baselines, learns from a small number of trajectories, and produces beliefs that converge on the true latent state.

Still, the current treatment could be improved in several ways. First, our evaluation covers three structured symbolic suites; it remains to test \emph{Pinductor} on domains with continuous observations, where in preliminary CartPole experiments both LLM-based methods did poorly (although some candidates produced sensible ODE integrators, suggesting that the representation is reachable and parameter fitting is the difficulty), and to assess transfer learning \citep{tang2024worldcoder}. Second, the observation distance is hand-specified per domain: the kernel form transferred unchanged across the three suites, with only a few lines rewritten each time, but the LLM prior currently used to optimize the world model could also be used to propose the distance, the planner, and the demonstration buffer, which are currently fixed. Third, the bootstrap particle filter suffers from weight degeneracy as the state dimension grows and observations become sparse \citep{doucet2009tutorial}; the rejuvenation step (Appendix~\ref{particle-filtering-rejuv}) mitigates this without removing it, and the cost of filtering and scoring is expected to grow accordingly. Fourth, on MineSweeper medium and hard even the true models fail with our filter and planner, so stronger inference and planning should directly benefit \emph{Pinductor}. Fifth, the reliance on LLM API calls induces high variance in \emph{Pinductor} and related methods \citep{curtis2025llmguided}: replaying one byte-identical request fifty times returned an exactly correct Block model 36\% of the time, and three distinct programs in the first three replies (Appendix~\ref{app:api-variance}). Future work should aim to reduce this variance.



\newpage

\bibliographystyle{unsrtnat}
\bibliography{references}

\newpage

\appendix

\section{Methodological Details}
\label{app:method-details}

This appendix provides implementation details omitted from the main methodology
section for space. Appendix~\ref{app:kernel-likelihood} gives the full
distance-kernel likelihood used for model evaluation. Appendix~\ref{particle-filtering-rejuv}
describes the particle filter and rejuvenation procedure. Appendix~\ref{app:refinement-details}
gives additional details on refinement, UCB1 selection, and final model
selection.

\subsection{Distance-kernel likelihood}
\label{app:kernel-likelihood}

The main paper defines the MiniGrid observation distance as
\[
d_{\mathrm{obs}}(o',o)
=
d_{\mathrm{grid}}(g',g)
+
\lambda_{\mathrm{dir}}\mathbf{1}[\theta'\neq\theta]
+
\lambda_{\mathrm{carry}}\mathbf{1}[c'\neq c],
\]
where \(g\) is the partial field-of-view grid, \(\theta\) is the agent direction,
and \(c\) is the carried object. The grid component is
\begin{equation}
\label{eq:app-grid-distance}
d_{\mathrm{grid}}(g',g)
=
\begin{cases}
-\log\!\left(
\varepsilon
+
(1-\varepsilon)
\left(
\frac{
\sum_{(i,j)\in\mathcal{V}}
\mathbf{1}[g'_{ij}=g_{ij}]
}
{|\mathcal{V}|}
\right)^{\alpha_{\mathrm{grid}}}
\right),
& \text{if at least one visible cell matches},\\[0.6em]
\infty,
& \text{otherwise}.
\end{cases}
\end{equation}
Here, \(\mathcal{V}\) denotes the set of visible non-agent cells. We use
\(\varepsilon=10^{-6}\) for numerical smoothing and \(\alpha_{\mathrm{grid}}=3\) to penalize
weak matches super-linearly. Throughout the experiments, we set
\(\lambda_{\mathrm{dir}}=\lambda_{\mathrm{carry}}=1\).

For a candidate model \(m\), let
\(o^m(s,a)\sim O^m_{\text{LLM}}(s,a)\) denote the observation predicted by
the LLM-generated observation program. The induced softened observation model is
\begin{equation}
\label{eq:app-distance-kernel}
O^m(o\mid s,a)
\propto
\exp\!\left(
-d_{\mathrm{obs}}(o^m(s,a),o)/\kappa
\right),
\end{equation}
where \(\kappa>0\) controls the sharpness of the likelihood. Smaller values of
\(\kappa\) penalize observation mismatches more severely, while larger values
make the score more tolerant to near-misses.


The particle-filtered kernel score to evaluate probabilistic models \(P^m\) used in the main paper is a
posterior expected kernel log-likelihood:
\begin{equation}
\label{eq:app-kernel-score-expanded}
\mathcal{L}(P^m;\mathcal{D})
=
\frac{1}{|\mathcal{D}|}\sum_{\tau\in\mathcal{D}}
\sum_{t=0}^{H_\tau-1}
\mathbb{E}_{s_{t+1}\sim Q_{t+1}^{m,\tau}}
\left[
\log O^m(o_{t+1}^{\tau}\mid s_{t+1},a_t^{\tau})
\right],
\end{equation}
using the softened observation model
above, up to fixed normalization constants. This objective is used only as a
differentiability-free model-ranking signal. It does not estimate the log marginal likelihood. In the implementation, each trajectory's inner sum is clipped below at \(-100\), so that a single trajectory on which the particle filter collapses cannot dominate the average. On MineSweeper, where every click is rewarded or penalised and hitting a mine shows up mainly in the reward and termination, we add a reward-prediction term \(-\sum_{t}\mathbb{E}_{Q_{t+1}^{m,\tau}}\big[(R^m(s_t,a_t,s_{t+1})-r_{t+1})^2\big]\) to each trajectory's score, so that the score also penalises models whose reward and termination logic disagrees with the data.

\subsection{Particle filtering and rejuvenation}
\label{particle-filtering-rejuv}

For each trajectory \(\tau\), the particle filter maintains a set of \(K\)
particles representing the belief \(Q_t^{m,\tau}\). Particles are initialized from the
candidate initial-state model:
\[
s_0^i \sim \rho_0^m,
\qquad i=1,\ldots,K.
\]
At each timestep, particles are propagated through the candidate transition
model:
\[
s_{t+1}^i \sim T^m(\cdot\mid s_t^i,a_t^\tau).
\]
The LLM-generated observation program then produces
\[
o_{t+1}^i \sim O^m_{\text{LLM}}(s_{t+1}^i,a_t^\tau),
\]
and the particle receives kernel weight
\begin{equation}
\label{eq:particle-weight}
w_{t+1}^i
\propto
O^m(o_{t+1}^\tau\mid s_{t+1}^i,a_t^\tau)
\propto
\exp\!\left(
-d_{\mathrm{obs}}(o_{t+1}^i,o_{t+1}^\tau)/\kappa
\right).
\end{equation}
Weights are normalized to obtain the particle belief \(Q_{t+1}^{m,\tau}\). We resample the belief after every observation update: weights in \eqref{eq:particle-weight} carry only the current step's kernel likelihood (not a cumulative product), and the belief passed to step \(t+1\) is reconstructed from these weights as integer particle counts proportional to \(w_{t+1}^i\). This makes the per-step expression a valid posterior approximation under bootstrap-style resampling. The
effective sample size is
\begin{equation}
\label{eq:app-ess}
\mathrm{ESS}_{t+1}
=
\frac{1}{\sum_i (\bar{w}_{t+1}^i)^2},
\end{equation}
where \(\bar{w}_{t+1}^i\) are weights normalized to sum to $1$.

We trigger belief rejuvenation when the total post-observation weight drops below
  \(K\) (i.e., when the kernel-weighted particle mass cannot represent a full          
  population). When this happens, we draw fresh particles from \(\rho_0^m\), roll them
forward through the recorded action history \(a_{0:t}^{\tau}\), and keep rollouts with
probability proportional to their cumulative kernel likelihood:
\begin{equation}
\label{eq:app-rejuvenation-score}
\prod_{\ell=0}^{t}
\exp\!\left(
-d_{\mathrm{obs}}(O^m_{\text{LLM}}(s_{\ell+1},a_\ell^\tau),o_{\ell+1}^\tau)/\kappa
\right).
\end{equation}
This procedure reintroduces diversity when the particle population collapses,
while still conditioning on the observed observation-action history. Note that under rejuvenation the particle population can transiently exceed the configured \(K\) particles, so the reported per-step ESS values in refinement diagnostics may temporarily exceed \(K\); the score itself remains well-defined since it depends only on the normalised weights.

In all experiments, we use \(K=10\) particles and \(\kappa=0.2\). The same
filtering procedure is used for both offline model evaluation and online belief
tracking during planning. See Appendix~\ref{app:particle-filtering} for general background on particle filtering.

\subsection{Feedback, refinement, and candidate selection}
\label{app:refinement-details}

Each evaluated candidate model produces three feedback objects used in the next
refinement prompt.

First, the scalar score
\[
\mathcal{L}_j=\mathcal{L}(P^{m_j};\mathcal{D})
\]
summarizes global compatibility with the offline trajectories. This score is
used both for UCB1 parent selection during refinement and for final model
selection.

Second, local diagnostics \(D_j\) summarize failure modes. These include runtime
errors, invalid outputs, low-likelihood trajectory segments, observation
contexts where the predicted observation differs substantially from the recorded
observation, and reward or termination mismatches. These diagnostics are included in the refinement prompt so that the
LLM can make targeted edits to the generated code rather than regenerate the
entire model from scratch. Candidate models that raise execution errors during particle filtering receive \(\mathcal{L}_j 
  = -\infty\) and are excluded from the near-best set \(\mathcal{N}\); they remain in
  the persistent pool \(\mathcal{P}\) so their diagnostics inform the next refinement  
  prompt, but cannot be sampled as the final model.   

Third, the QBC signal \(Q_j\) summarizes high-disagreement transition contexts.
For the committee
\[
\mathcal{C}_j=\{T^{m_1},\ldots,T^{m_j}\},
\]
each model predicts a next state for a belief particle \(s\) and action \(a\), yielding votes
\[
V_{s,a}(y)=\sum_{T\in\mathcal{C}_j}\mathbf{1}[T(s,a)=y],
\]
and we compute the normalized vote entropy of ~\eqref{eq:qbc-ve}, with \(\operatorname{VE}(s,a)=0\) when fewer than two transition models are available. The disagreement
score is computed on particles sampled from the filtering beliefs, using both the realized actions and counterfactual actions, so the LLM sees where the current model family is unsure about the dynamics. We then
include the highest-disagreement contexts in the prompt, together with the
corresponding observation context and action. The QBC signal is diagnostic only:
it is not added to the model-selection objective.

\paragraph{UCB1 parent selection}
The candidate population is maintained as a persistent refinement tree rather than a single chain of edits: in each REx round, \emph{Pinductor} refines one existing candidate, asks the LLM for \(M\) revised candidates, evaluates them, and adds them to the pool. High-scoring candidates are more likely to be refined, but candidates that have been explored less often can also be selected. At each
round, \emph{Pinductor} selects a parent node \(p\) according to UCB1:
\begin{equation}
\label{eq:app-ucb1}
\mathrm{UCB}(p)
=
\bar{\mathcal{L}}(p)
+
c_{\mathrm{ucb}}
\sqrt{
\frac{\log n_{\mathrm{parent}(p)}}{n_p+1}
},
\end{equation}
where \(\bar{\mathcal{L}}(p)\) is the mean kernel score of candidates generated from node
\(p\), \(n_p\) is the number of times \(p\) has been expanded, and
\(n_{\mathrm{parent}(p)}\) is the visit count of its parent. If the candidate
pool is empty, the root prompt is used. This tree-structured expansion allows
refinement to concentrate on promising candidate families while preserving an
exploration bonus for less-expanded branches. We use a smoothed denominator \((n_p + 1)\) inside the square root, so the
  exploration bonus stays finite even for unexpanded nodes (\(n_p=0\)). Fresh nodes
  still receive the largest bonus (no prior visits to amortize) but cannot dominate
  the score outright, allowing a sibling with a good \(\bar{\mathcal{L}}\) to overtake them.
  The exploration constant is \(c_{\mathrm{ucb}}=1.0\) in all experiments.

\paragraph{Final selection}
After all REx rounds, the final model is sampled from the near-best set
\begin{equation}
\label{eq:near-best-set}
\mathcal{N}
=
\{j : \mathcal{L}_j \ge \mathcal{L}^\star - \mathrm{std}(\{\mathcal{L}_{j'}\})\},
\qquad
\mathcal{L}^\star=\max_j \mathcal{L}_j.
\end{equation}
Sampling from this set rather than taking a hard argmax avoids over-committing
to small differences in the noisy particle-filtered score. The final selector is
\begin{equation}
\label{eq:final-selector}
\Pr(m^\star=m_j)
=
\frac{\exp(\mathcal{L}_j/\beta)}
{\sum_{k\in\mathcal{N}}\exp(\mathcal{L}_k/\beta)},
\qquad j\in\mathcal{N}.
\end{equation}
We use \(\beta=0.1\). As \(\beta\to 0\), this recovers hard argmax selection within the
near-best set. Since \(\mathcal{N}\) always contains the current best-scoring
candidate, the procedure preserves access to the best model found while allowing
selection of a structurally different candidate with statistically similar
score.

\subsection{Relation to bandit-guided and evolutionary program search}
\label{app:search-correspondence}

The refinement loop of Alg.~\ref{alg:belief-llm-refinement} resembles genetic algorithms, cross-entropy-style population search, and bandit-based optimisation. The correspondence is exact rather than loose, and we make it explicit here.

\begin{table}[H]
\centering
\small
\caption{Correspondence between \emph{Pinductor}'s refinement loop and bandit-guided stochastic program search. The closest relative is FunSearch \citet{romera2024funsearch}, which also pairs an LLM mutation operator with a scored population; the differences are the fitness (a filtered likelihood computed without hidden states), the parent-selection rule (UCB1 on a persistent tree rather than island-based sampling), and the diagnostic feedback injected into each mutation prompt.}
\label{tab:search_correspondence}
\footnotesize
\begin{tabular}{lp{9.2cm}}
\toprule
\textbf{Search component} & \textbf{In \emph{Pinductor}} \\
\midrule
Population & Persistent candidate pool $\mathcal{P}$, organised as a refinement tree \\
Fitness & Particle-filtered kernel log-likelihood $\mathcal{L}(P^m;\mathcal{D})$, \eqref{eq:kernel-score} \\
Parent selection & UCB1 over tree nodes \citep{auer2002finite}, \eqref{eq:app-ucb1} \\
Mutation operator & LLM refinement call conditioned on score, local diagnostics $D_j$, and disagreement $Q_j$ \\
Offspring per round & $M$ candidates from the selected parent \\
Selection & Softened argmax over the near-best set, \eqref{eq:near-best-set}--\eqref{eq:final-selector} \\
Termination & Fixed number of rounds $J$ (offline), retriggered after each online episode \\
\bottomrule
\end{tabular}
\end{table}

Two consequences follow. First, the exploration bonus of UCB1 plays the role of mutation diversity: less-expanded branches of the tree are revisited even when a sibling currently scores higher. Second, the vote-entropy signal \eqref{eq:qbc-ve} is a query-by-committee heuristic layered on top of the search: it does not change the fitness, but it directs the mutation operator to the transition contexts where the population disagrees most.

\subsection{Hyperparameters}
\label{app:hyperparameters}

Unless otherwise stated, all experiments use \(K=10\) particles, bandwidth
\(\kappa=0.2\) (for \emph{Pinductor}), \(M=5\) candidates per REx round, final-selection temperature
\(T=0.1\), direction and carried-object penalties
\(\lambda_{\mathrm{dir}}=\lambda_{\mathrm{carry}}=1\) (for \emph{Pinductor}), grid-distance exponent
\(\alpha_{\mathrm{grid}}=3\) (for \emph{Pinductor}), smoothing constant \(\varepsilon=10^{-6}\), and a planner budget of \(5{,}000\) belief-state expansions.
\newtext{On Block, both methods use \(K=100\) particles, a planner budget of \(20{,}000\) belief-state expansions, an entropy coefficient of 0 (defined below), a per-step action cost of 0.01, and a QMDP cost-to-go heuristic (scaled by 0.01) computed from the models the planner currently holds, which are the induced ones; \emph{Pinductor} uses \(\kappa=1\). On MineSweeper, both methods use \(K=80\) particles, a planner budget of 30 belief-state expansions, an entropy coefficient of 0.05, and \(M=2\) candidates per offline REx round and 1 per online round; \emph{Pinductor} uses \(\kappa=10^{-3}\), selects the parent candidate by Thompson sampling from the same per-node Beta statistics rather than by UCB1, and keeps the highest-scoring candidate as the final model (\(T=0\)).}

We follow the hyperparameter setting of \citet{curtis2025llmguided} closely, but reduce the number of offline and online refinement attempts from 25 to 5 after observing no substantial performance change. The only notable difference between both setups is the entropy coefficient of the belief-space planner, which trades off reward-seeking and information-gathering behaviour. We tune this coefficient using the same protocol for all methods that use the planner and report the best-performing setting for each method. For \emph{Pinductor}, increasing the value of this coefficient led to better exploration and higher rewards on average, while it did not lead to any significant performance gain for POMDP Coder. \oldtext{In the experiments, \emph{Pinductor} used a coefficient of 1.0, and POMDP Coder used a coefficient of 0 (which is the default value in its own implementation).} \newtext{On MiniGrid, \emph{Pinductor} used a coefficient of 1.0, and POMDP Coder used a coefficient of 0 (which is the default value in its own implementation); the Block and MineSweeper values are given above.}

\subsection{Role of the components}

\emph{Pinductor} separates four roles that are often entangled in partially observable
model learning. The LLM provides a structured prior over executable model
programs. Particle filtering provides the learning signal by asking whether the
candidate model can maintain plausible latent trajectories that explain partial
observations. The distance kernel makes this evaluation robust to mismatches in
high-dimensional partial observations. QBC disagreement identifies transition
contexts where plausible candidate models make incompatible predictions. Finally,
belief-space planning uses the selected model in the same representation used
for evaluation, closing the loop between model proposal, latent-state inference,
diagnostic refinement, and action selection.

\section{Particle Filtering Background}
\label{app:particle-filtering}

Particle filtering is a sequential Monte Carlo method for approximately
maintaining beliefs over latent states in partially observable dynamical systems.
In a POMDP with transition model \(T\) and observation model \(O\), the exact
Bayes filter updates a belief \(Q_t(s_t)\) by first predicting through the
transition model and then conditioning on the new observation. After action
\(a_t\), the predictive belief is
\[
\bar{Q}_{t+1}(s') = \sum_{s\in\mathcal{S}} T(s'\mid s,a_t)\,Q_t(s),   
\]
and the observation update is
\[
Q_{t+1}(s')
\propto
\sum_{s'\in\mathcal{S}} O(o_{t+1}\mid s',a_t)\bar{Q}_{t+1}(s').
\]
Particle filtering approximates this belief with a finite set of weighted
particles \(\{(s_t^i,w_t^i)\}_{i=1}^K\). In the bootstrap particle filter,
particles are propagated through the transition model,
\[
s_{t+1}^i \sim T(\cdot\mid s_t^i,a_t),
\]
and reweighted by the observation likelihood,
\[
w_{t+1}^i
\propto
O(o_{t+1}\mid s_{t+1}^i,a_t).
\]
Weights are then normalized, and resampling may be used to avoid degeneracy when
most particle mass concentrates on a small number of particles.

\emph{Pinductor} follows this basic filtering structure, but adapts it to executable
LLM-generated POMDP programs. The candidate transition and observation models
may be deterministic programs rather than soft probability distributions.
After propagating a particle through the candidate transition model,
\emph{Pinductor} obtains a predicted observation
\[
o_{t+1}^i\sim O^m_{\text{LLM}}(s_{t+1}^i,a_t),
\]
and compares it to the observed \(o_{t+1}\) using the observation distance
\(d_{\mathrm{obs}}\). This distance induces the kernel weight
\[
w_{t+1}^i
\propto
O^m(o_{t+1}\mid s_{t+1}^i,a_t)
\propto
\exp\!\left(
-\frac{
d_{\mathrm{obs}}(o_{t+1}^i,o_{t+1})
}{\kappa}
\right),
\]
which plays the role of the observation likelihood in the filtering update. Thus, the update remains a standard particle-filtering update in form, where the LLM-generated likelihood is softened into a distance-kernel likelihood. This allows \emph{Pinductor} to assign positive likelihood to any realized observation, enabling learning, in spite of the fact that the LLM-generated observation model may be deterministic.

The implementation also uses a rejuvenation step when particle mass becomes too small. Rather than relying only on particles propagated from the previous belief, \emph{Pinductor} samples additional candidate initial states from \(\rho_0^m\), rolls them forward through the observed action history, and keeps them with weight proportional to their cumulative kernel likelihood along the trajectory. This replenishes the belief with latent trajectories that remain compatible with the observations, which is important because LLM-generated models can be imperfect and a strict particle filter can otherwise collapse after successive mismatches.

After reweighting and rejuvenation, the weighted particles define the model's filtered belief \(Q_t^m\). The same belief is used both for scoring candidate models and for downstream planning. This is central to \emph{Pinductor}: the evaluation does not ask whether any particle matches a privileged hidden state, but whether the candidate model can maintain a belief whose latent trajectories explain the observation stream.

\section{Distance-Kernel Likelihood and Energy Interpretation}
\label{app:likelihood_energy}

\emph{Pinductor} evaluates executable candidate models in a setting where the latent
state is never observed. We provide some clarifications for the scoring objective
used in the main text \eqref{eq:kernel-score} and its connection to both likelihood-based filtering and
energy-based inference.

Let \(m\) be a candidate executable POMDP model. Given a latent state
\(s_{t+1}\) and action \(a_t\), the candidate observation program predicts observations
\[
o_{t+1}\sim O^m_{\text{LLM}}(s_{t+1},a_t).
\]
Since in practice the LLM generated program is usually deterministic, it usually defines a Dirac likelihood, which causes problems for our objective since most realized observations have zero mass under this model. To circumvent this problem, \emph{Pinductor} defines an observation compatibility energy
\[
E_m(s_{t+1},o_{t+1},a_t)
=
d_{\mathrm{obs}}(O^m_{\text{LLM}}(s_{t+1},a_t),o_{t+1}),
\]
where \(d_{\mathrm{obs}}\) is the task-specific observation distance. This
energy induces an unnormalized distance-kernel likelihood
\[
K_\kappa^m(o_{t+1}\mid s_{t+1},a_t)
=
\exp\!\left(
-\frac{
E_m(s_{t+1},o_{t+1},a_t)
}{\kappa}
\right),
\]
with bandwidth \(\kappa>0\). Observations that are better explained by the
candidate state and realized action receive higher kernel values, while observations far from the candidate prediction receive exponentially smaller values. If desired, this kernel can be normalized, giving the soft likelihood model
\(O^m(o_{t+1}\mid s_{t+1},a_t)\propto
K_\kappa^m(o_{t+1}\mid s_{t+1},a_t)\).
Let \(Q_t^{m,\tau}\) denote the particle belief at time $t$, induced by model \(m\) after
observing trajectory \(\tau\). After action \(a_t^\tau\), the predictive belief is
obtained by propagating particles through the candidate transition model:
\[
\bar{Q}_{t+1}^{m,\tau}(s')
=
\sum_{s\in\mathcal{S}} T^m(s'\mid s,a_t^\tau) Q_t^{m,\tau}(s).
\]
This is simply the push-forward of the
particle set through \(T^m\) and constitutes a prior about the state at time $t+1$. The Bayesian update \eqref{eq:bayes} proceeds to reweight the prior into the posterior. This proceeds by reweighting by the likelihood of realized observations \(o_{t+1}^\tau\):
\[
Q_{t+1}^{m,\tau}(s')
\propto
\sum_{s'\in\mathcal{S}} \bar{Q}_{t+1}^{m,\tau}(s')
O^m(o_{t+1}^\tau\mid s',a_t^\tau).
\]
Note that in practice we sum over all particles as opposed to all possible states (as states that are not assigned a particle have prior probability zero), which makes the number of summands tractable. Equivalently, each particle in $\bar{Q}_{t+1}^{m,\tau}$ receives weight
\[
w_{t+1}^i
\propto
O^m(o_{t+1}^\tau\mid s_{t+1}^i,a_t^\tau)
\propto
\exp\!\left(
-\frac{
d_{\mathrm{obs}}(O^m_{\text{LLM}}(s_{t+1}^i,a_t^\tau),o_{t+1}^\tau)
}{\kappa}
\right).
\]
This is exactly a Boltzmann reweighting with energy \(E_m(s_{t+1}, o_{t+1}^\tau,a_t^\tau)\) and temperature \(\kappa\): low-energy particles receive high posterior mass, while high-energy particles are suppressed.

The model ranking score used by \emph{Pinductor} is the filtered expected log soft-observation
likelihood:
\[
\mathcal{L}(P^m;\mathcal{D})
=
\frac{1}{|\mathcal{D}|}\sum_{\tau\in\mathcal{D}}
\sum_{t=0}^{H_\tau-1}
\mathbb{E}_{s_{t+1}\sim Q_{t+1}^{m,\tau}}
\left[
\log O^m(o_{t+1}^\tau\mid s_{t+1},a_t^\tau)
\right].
\]
Substituting the unnormalized kernel gives, up to additive constants independent
of \(m\),
\[
\mathcal{L}(P^m;\mathcal{D})
=
-\frac{1}{\kappa}
\frac{1}{|\mathcal{D}|}\sum_{\tau\in\mathcal{D}}
\sum_{t=0}^{H_\tau-1}
\mathbb{E}_{s_{t+1}\sim Q_{t+1}^{m,\tau}}
\left[
d_{\mathrm{obs}}(O^m_{\text{LLM}}(s_{t+1},a_t^\tau),o_{t+1}^\tau)
\right].
\]
The $1/\kappa$ prefactor is a positive constant that does not affect $\arg\max_m$ at fixed $\kappa$, so maximizing the score is equivalent to minimizing the expected observation energy under the filtered belief induced by the candidate model. Note however, that, since we sample models through a softmax \eqref{eq:final-selector}, $\kappa$ does play the role of a temperature parameter there jointly with $T$. Also, and unlike $T$, $\kappa$ controls the variance of the soft likelihood and therefore shapes the belief $Q_{t+1}^{m,\tau}$ under which the expectation is taken: it is a genuine hyperparameter of the method, not a cosmetic scale.

The resulting score has three useful properties for strict POMDP induction.
First, it is computable from observations, actions, rewards, and termination
signals without access to hidden-state labels. Second, it is defined for
executable programs, including deterministic transition and observation
functions, because it only requires simulating latent particles and comparing
predicted observations to observed ones. Third, it is aligned with the belief
state used for planning: a candidate model scores well only if its own filtering
process maintains latent particles that explain the observed trajectory. In this
sense, the score evaluates the operational object needed by the planner, namely
the model-induced belief, rather than an oracle comparison to privileged states.

We therefore view \(\mathcal{L}(P^m;\mathcal{D})\) as a particle-filtered
expected likelihood, or equivalently as a negative expected observation energy
under posterior beliefs. It should not be read as a variational ELBO
or as an exact marginal likelihood.

\section{Experimental details}
\label{app:experimental-details}

\subsection{Stochastic MiniGrid variants}
We use stochastic MiniGrid variants to test whether model induction remains robust when the same task can instantiate into multiple concrete layouts. We do not include Corners in this experiment, since this environment is already stochastic: the goal corner, agent starting position, and agent direction vary across episodes. Lava and Unlock already contain some stochasticity, but we increase the episode-level variation in the quantities that are relevant to model learning. In the standard Lava environment, the lava wall position and gap position vary across episodes; we increase the stochasticity to the agent's starting position and direction, and to the goal position. In the standard Unlock environment, the agent's start position and direction, as well as the key position, vary across episodes; we increase the stochasticity to the wall and goal position. These variants make it harder for a generated model to memorize a single layout and instead require it to represent the latent structure of the task.

\subsection{Offline datasets}
Our offline datasets differ from those used in the original POMDP Coder paper \citep{curtis2025llmguided}. The original datasets were composed of oracle trajectories, i.e. successful demonstrations generated by an expert policy. This provides clean examples of how to solve the task, but it gives little or no evidence about failure modes. For instance, in Lava, a success-only dataset may never show what happens when the agent walks into lava, so the learner must infer this terminal event from the environment description or prior knowledge rather than from data.

We therefore construct the offline datasets by mixing successful and unsuccessful episodes, including failures and timeouts when available. Empirically, this improved performance in some environments. The goal is to increase behavioral diversity in the replay buffer and encourage the LLM to learn transition, observation, and reward structure rather than memorize the oracle trajectory distribution. In particular, mixed datasets expose both positive terminal events, such as reaching the goal, and negative or zero-reward outcomes, such as entering lava or failing to complete the task within the horizon.

Both LLM-based POMDP methods used in the paper use the same datasets for all experiments to ensure fair comparison.

\paragraph{Offline sample-efficiency sweep}
\label{app:offline-sweep}
For Fig.~\ref{fig:offline_sample_eff}, each method induces its initial model from the first \(N_D\in\{2,4,\ldots,12\}\) demonstrations (nested subsets) and then plays three online episodes, refining the model after each. We report the mean discounted return of these three episodes, averaged over 10 seeds; shaded bands are 95\% bootstrap confidence intervals over seed means. The points at \(N_D=10\) for the two LLM-based methods are the runs of the main experiment.

\paragraph{Demonstrations for the additional suites}
Every learning method consumes the same demonstrations, so their composition bounds what any of them can do. On Block, the demonstration set consists of 15 episodes (780 steps) produced by a breadth-first optimal solver; every episode is successful, and the mean episode length is 52 steps. Because the solver never pushes the gripper into the edge of the row, the transition convention at the boundary is not identifiable from the demonstrations, which matters for the analysis of Appendix~\ref{app:model-audit}. On MineSweeper, the demonstration set consists of 50 episodes per difficulty generated by a random policy; a positive return occurs in 10 of the 50 episodes on easy and in none of the 50 on medium and hard, so no method ever observes the task being completed there. Random rollouts do, however, exercise the terminal hazard (clicking a mine), which the mixed MiniGrid datasets were designed to expose as well.

\subsection{True-model reference and Optimal bound}
\label{app:gt-reference}
In the plots, ``True models, same planner'' refers to runs in which the learned model components are replaced by the manually specified models from \citet{curtis2025llmguided}, run through our own particle filter and belief-space planner. These models were manually verified and serve as a reference for good model quality. However, they should not be interpreted as an optimal-policy oracle: the state remains hidden, the planning budget is finite, and the downstream planner remains approximate. Consequently, a generated model can sometimes outperform this reference in terms of realized reward. For this reference, the planning stage is similar to POMDP Coder; the only difference is that the models are handcrafted.

To make the distinction explicit, we additionally report \emph{Optimal}: an oracle that observes the true state and plans exactly (breadth-first search on the true dynamics for MiniGrid and Block; an omniscient safe-cell policy for MineSweeper), using none of our pipeline. Optimal is an upper bound that no partially observable method can reach; the true-model reference is the level a method could in principle match. Table~\ref{tab:oracle_vs_gt} reports both on all nine environments.

\subsection{Environment descriptions and natural-language sweep}
For the main experiments, we provide the LLM with the strongest environment
description level, denoted L3 in Fig.~\ref{fig:ablation_prompt_levels}. The
natural-language sweep varies how much prior information the LLM receives, from
no description at L0 to the full task description at L3. Thus, the sweep
measures sensitivity to the LLM's environment prior: moving from L0 to L3 gives
the model increasingly explicit information about the environment and the preferred POMDP structure. In all
main-result experiments, we use L3.

Below, we list the exact natural-language descriptions used at each level. The
description text provided to the LLM is shown in quotation marks and italics.

\paragraph{\textsc{Corners}}
\begin{description}
    \item[\textbf{L0}:] No natural-language environment description.

    \item[\textbf{L1}:] \textit{``10x10 grid.''}

    \item[\textbf{L2}:] \textit{``10x10 grid with boundary walls. Goal is in a
    corner of the grid. Agent starts at an interior position.''}

    \item[\textbf{L3}:] \textit{``10x10 grid with boundary walls. Goal is placed
    in a random corner of the grid. Agent starts at a random interior position
    with random direction. No obstacles. Both agent start and goal corner vary
    between episodes.''}
\end{description}

\paragraph{\textsc{Lava}}
\begin{description}
    \item[\textbf{L0}:] No natural-language environment description.

    \item[\textbf{L1}:] \textit{``10x10 grid with lava.''}

    \item[\textbf{L2}:] \textit{``10x10 grid with boundary walls. A vertical
    wall of lava with one passable gap. Agent starts at $(1,1)$. Goal is at the
    bottom-right corner.''}

    \item[\textbf{L3}:] \textit{``10x10 grid with boundary walls. A vertical wall
    of lava spans most of one column with exactly one gap to pass through.
    Agent starts at $(1,1)$. Goal is at the bottom-right corner. The lava column
    position and gap position vary between episodes.''}
\end{description}

\paragraph{\textsc{FourRooms}}
\begin{description}
    \item[\textbf{L0}:] No natural-language environment description.

    \item[\textbf{L1}:] \textit{``19x19 grid.''}

    \item[\textbf{L2}:] \textit{``19x19 grid divided into 4 rooms by internal
    walls, each connected by one doorway. Agent and goal are placed in different
    rooms.''}

    \item[\textbf{L3}:] \textit{``19x19 grid divided into 4 rooms by internal
    walls, each room connected by one doorway. Agent and goal are placed at
    random positions in different rooms. Room layout is fixed, but agent start
    position, direction, and goal position vary between episodes.''}
\end{description}

\paragraph{\textsc{Unlock}}
\begin{description}
    \item[\textbf{L0}:] No natural-language environment description.

    \item[\textbf{L1}:] \textit{``Grid world with a locked door.''}

    \item[\textbf{L2}:] \textit{``Grid world with boundary walls and a locked
    door separating the map into two sections. A key is in the accessible
    section; the goal is behind the locked door.''}

    \item[\textbf{L3}:] \textit{``Grid world with boundary walls and a locked
    door separating the map into two sections. A key is placed in the accessible
    section, and the goal is located behind the locked door. Agent starts in the
    initial section with random position and direction. The room layout is
    fixed, while agent start position, key position, and goal position vary
    between episodes.''}
\end{description}

For the two additional suites, only the full description was used (the natural-language sweep was not run on them). The description is byte-identical between \emph{Pinductor} and POMDP Coder.

\paragraph{\textsc{Block}}
\textit{``A Blocks World manipulation task. A gripper moves along a row of columns; each column holds a vertical stack of distinct numbered blocks (bottom to top). The goal is to rearrange the blocks into a fixed target configuration; the task is solved the moment the stacks match the target and the gripper holds nothing. There are 4 actions: 0, 1, 2, 3. Two of them move the gripper between adjacent columns (left / right) and two manipulate blocks (pick up the top block of the current column / put the held block down on the current column). Work out the exact effect of each action - including when an action does nothing - from the demonstrations. The gripper observes ONLY the stack at its current column plus the block it is holding; it never sees its absolute column index or the other columns, and its starting column is unknown. Reward is sparse and terminal: +1.0 on the step that reaches the target configuration, 0.0 otherwise. Concretely, EACH EPISODE the 6 blocks (numbered 1..6) START in a FRESH RANDOM configuration: they are scattered uniformly at random across the 3 columns, in random stacking order (any arrangement of the 6 blocks over the 3 columns is possible), the gripper holds nothing, and the gripper's STARTING COLUMN is uniformly random in \{0,1,2\}. The TARGET configuration (fixed across episodes) is: column 0 = [1,2,3,4,5,6] (block 1 at bottom, then 2,3,4,5, and 6 on top) - i.e. the sorted tower - columns 1 and 2 empty. The initial-state model must reproduce this DISTRIBUTION: random stacks over the 3 columns each episode, random gripper column, fixed target.''}

\paragraph{\textsc{MineSweeper} (easy)}
\textit{``PopGym MineSweeperEasy with wrappers PreviousAction, Antialias, Markovian(FULL\_IN\_INFO\_DICT), Flatten, and DiscreteAction. The action space is Discrete(16): action a clicks one square on the 4x4 board, row = a // 4 and col = a \% 4. The hidden state is the 4x4 grid flattened by PopGym, with 0=CLEAR, 1=MINE, and 2=VIEWED. The reset observation is a flattened 13-element binary vector: entries 0-2 one-hot encode the observed adjacent-mine count, entries 3-6 one-hot encode previous action row, entries 7-10 one-hot encode previous action column, and entries 11-12 are antialias/reset-vs-step bits. A safe unseen click marks the square VIEWED and gives a small positive reward; clicking a mine terminates with a negative reward; clicking an already viewed square gives a smaller negative reward. The objective is to reveal all non-mine cells while avoiding mines.''}
The medium and hard descriptions substitute the board size (6$\times$6, 8$\times$8), the action count (36, 64), and the observation length (21, 28 entries).

\section{General details on environments}
\label{app:envs}

We evaluate on five deterministic MiniGrid environments, described first, and on two further suites, Block and MineSweeper, described at the end of this section. For the interesting MiniGrid environments, we
illustrate what each method observes at the start of an episode and at the
moment the agent reaches the reward, by replaying the same recorded
\emph{Pinductor} trajectory under both observation models. \emph{Pinductor} only
ever sees a 3$\times$3 partial field of view (the current FOV is highlighted
in red, while previously visited cells remain visible as the cumulative
trajectory; cells never visited are masked as \texttt{unseen}). The main baseline \citep{curtis2025llmguided} has post-hoc access to the full underlying
state, shown unmasked.

\begin{figure}[H]
  \centering
  \begin{subfigure}[t]{0.235\linewidth}
    \centering
    \includegraphics[width=\linewidth]{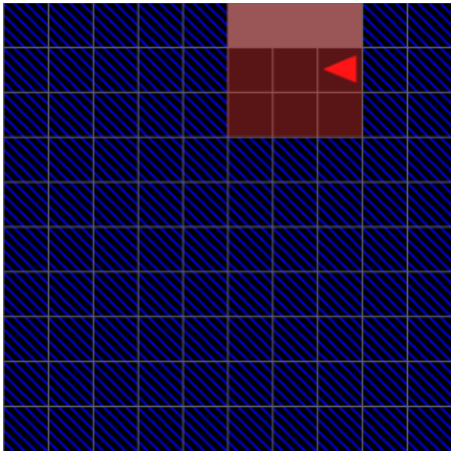}
    \caption{Pinductor, start}
  \end{subfigure}\hfill
  \begin{subfigure}[t]{0.235\linewidth}
    \centering
    \includegraphics[width=\linewidth]{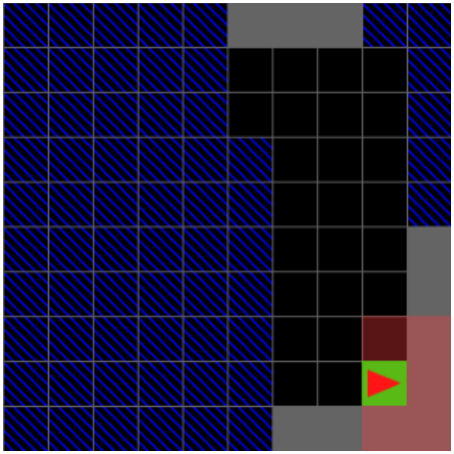}
    \caption{Pinductor, finish}
  \end{subfigure}\hfill
  \begin{subfigure}[t]{0.235\linewidth}
    \centering
    \includegraphics[width=\linewidth]{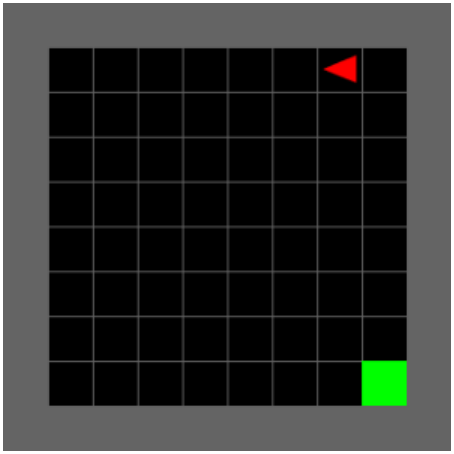}
    \caption{Curtis et al., start}
  \end{subfigure}\hfill
  \begin{subfigure}[t]{0.235\linewidth}
    \centering
    \includegraphics[width=\linewidth]{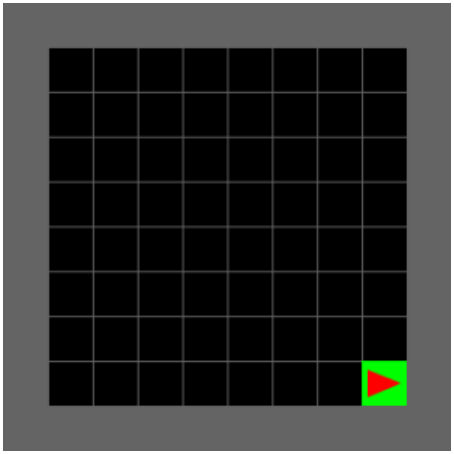}
    \caption{Curtis et al., finish}
  \end{subfigure}
  \caption{\textbf{Corners} -- observability comparison on a winning episode.}
  \label{fig:env:corners}
\end{figure}

\paragraph{Corners (\texttt{CornerGoalRandom-Empty-10x10-v0})} A 10$\times$10
empty room with the goal placed uniformly at random in one of the four
corners. The empty room is rotationally near-symmetric, so a small partial
FOV produces almost identical observations from each corner (Fig.~\ref{fig:env:corners}).
The main difficulty is therefore \emph{position disambiguation}: the agent
must navigate while reasoning about \emph{which} corner contains the goal
under substantial pose ambiguity.

\begin{figure}[H]
  \centering
  \begin{subfigure}[t]{0.235\linewidth}
    \centering
    \includegraphics[width=\linewidth]{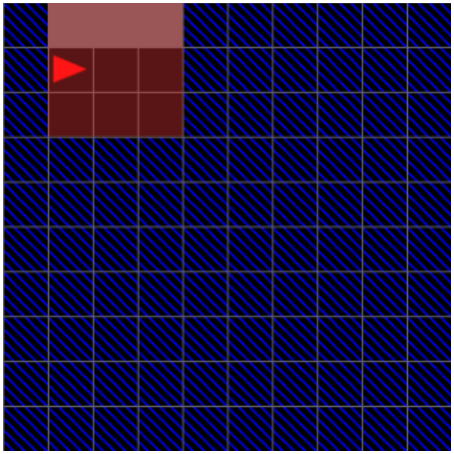}
    \caption{Pinductor, start}
  \end{subfigure}\hfill
  \begin{subfigure}[t]{0.235\linewidth}
    \centering
    \includegraphics[width=\linewidth]{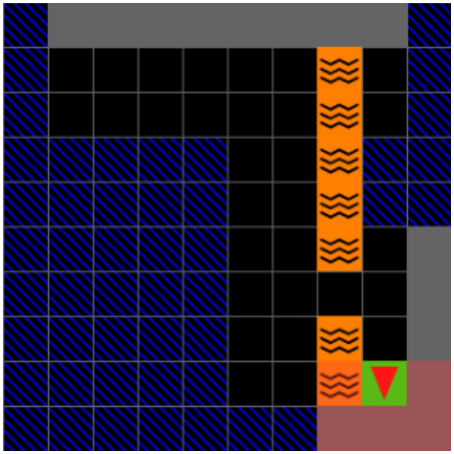}
    \caption{Pinductor, finish}
  \end{subfigure}\hfill
  \begin{subfigure}[t]{0.235\linewidth}
    \centering
    \includegraphics[width=\linewidth]{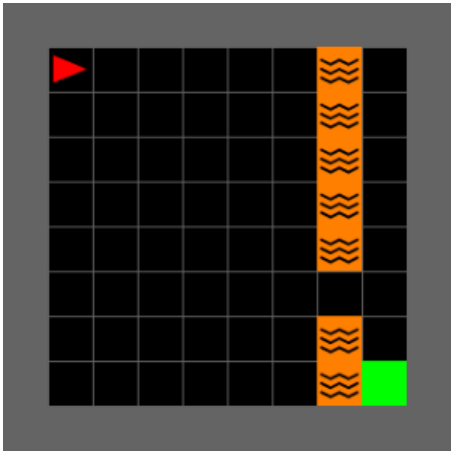}
    \caption{Curtis et al., start}
  \end{subfigure}\hfill
  \begin{subfigure}[t]{0.235\linewidth}
    \centering
    \includegraphics[width=\linewidth]{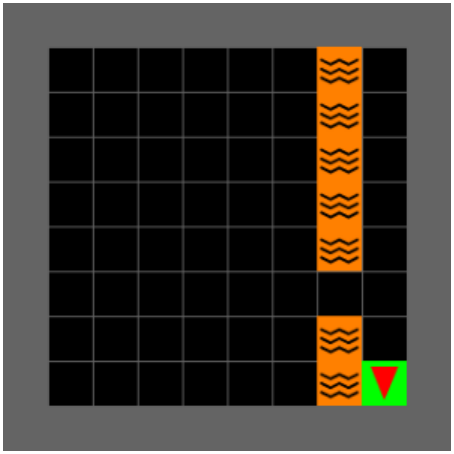}
    \caption{Curtis et al., finish}
  \end{subfigure}
  \caption{\textbf{Lava} -- observability comparison on a winning episode.}
  \label{fig:env:lava}
\end{figure}

\paragraph{Lava (\texttt{MyMiniGrid-LavaWall-v0})} A room split by a wall of
lava with a single safe passage (Fig.~\ref{fig:env:lava}). Stepping on a lava
cell terminates the episode with zero reward. The difficulty is \emph{learning
the terminal hazard}: a model that fails to encode lava as a terminating,
low-reward cell will produce a planner that takes the apparent shortcut and
dies. The reward signal alone is sparse, so the model must combine the
natural-language description and trajectory cues to identify lava semantics.

\begin{figure}[H]
  \centering
  \begin{subfigure}[t]{0.235\linewidth}
    \centering
    \includegraphics[width=\linewidth]{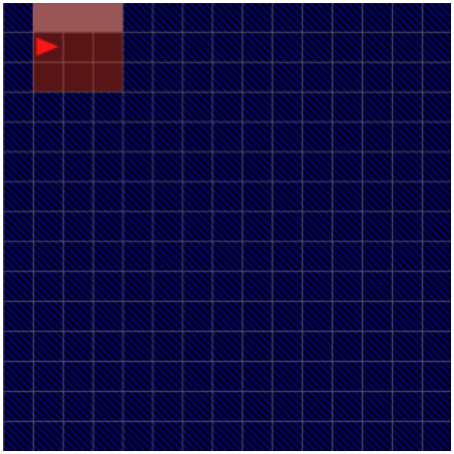}
    \caption{Pinductor, start}
  \end{subfigure}\hfill
  \begin{subfigure}[t]{0.235\linewidth}
    \centering
    \includegraphics[width=\linewidth]{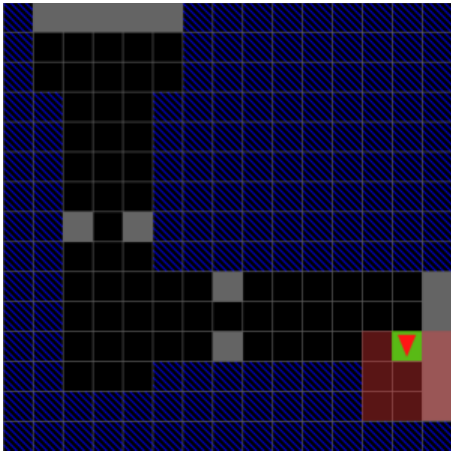}
    \caption{Pinductor, finish}
  \end{subfigure}\hfill
  \begin{subfigure}[t]{0.235\linewidth}
    \centering
    \includegraphics[width=\linewidth]{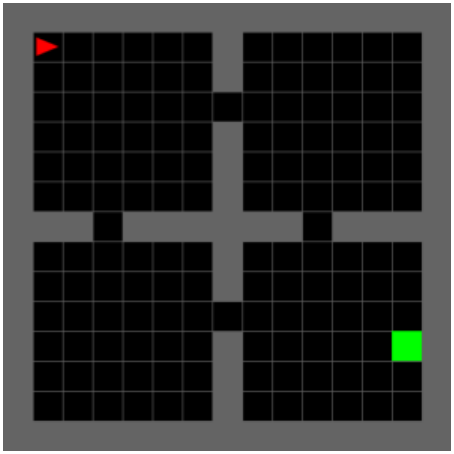}
    \caption{Curtis et al., start}
  \end{subfigure}\hfill
  \begin{subfigure}[t]{0.235\linewidth}
    \centering
    \includegraphics[width=\linewidth]{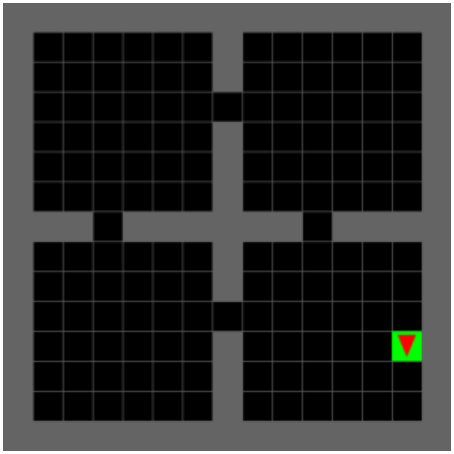}
    \caption{Curtis et al., finish}
  \end{subfigure}
  \caption{\textbf{Four Rooms} -- observability comparison on a winning episode.}
  \label{fig:env:four_rooms}
\end{figure}

\paragraph{Four Rooms (\texttt{MyMiniGrid-FourRooms-v0})} Four rooms
connected by narrow gap doorways, with the goal placed at a random position
that varies across episodes (Fig.~\ref{fig:env:four_rooms}). The difficulty
is \emph{long-horizon navigation}: episodes are long, the agent must commit
to traversing specific doorways without seeing the goal, and the random
goal placement prevents the model from collapsing to a fixed plan. The
3$\times$3 FOV makes each doorway choice a near-blind decision conditioned
on belief over goal location.

\begin{figure}[H]
  \centering
  \begin{subfigure}[t]{0.235\linewidth}
    \centering
    \includegraphics[width=\linewidth]{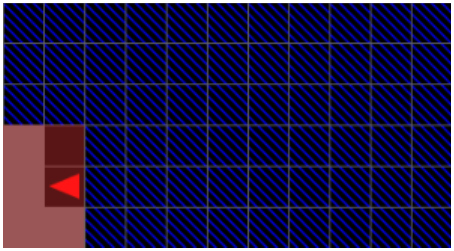}
    \caption{Pinductor, start}
  \end{subfigure}\hfill
  \begin{subfigure}[t]{0.235\linewidth}
    \centering
    \includegraphics[width=\linewidth]{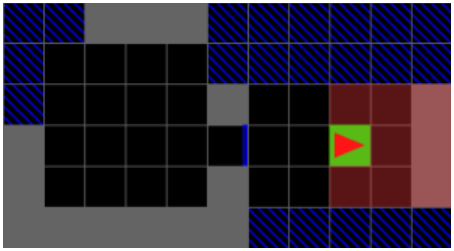}
    \caption{Pinductor, finish}
  \end{subfigure}\hfill
  \begin{subfigure}[t]{0.235\linewidth}
    \centering
    \includegraphics[width=\linewidth]{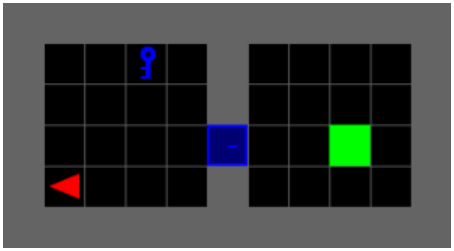}
    \caption{Curtis et al., start}
  \end{subfigure}\hfill
  \begin{subfigure}[t]{0.235\linewidth}
    \centering
    \includegraphics[width=\linewidth]{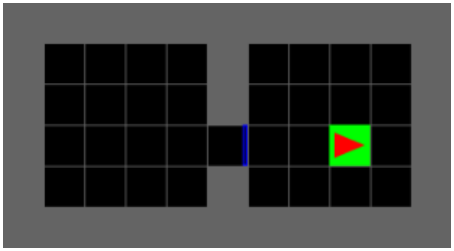}
    \caption{Curtis et al., finish}
  \end{subfigure}
  \caption{\textbf{Unlock} -- observability comparison on a winning episode.}
  \label{fig:env:unlock}
\end{figure}

\paragraph{Unlock (\texttt{MyUnlockEnv-v0})} An 11$\times$6 grid split into
two 6$\times$6 rooms by a wall pierced by a single locked door at column~5
(Fig.~\ref{fig:env:unlock}). A key, whose colour matches the door, lies
somewhere in the accessible left room; the goal sits at a fixed position
behind the locked door. The difficulty is \emph{compositional structure}:
the agent must execute a multi-step plan (locate key $\to$ pick up $\to$
unlock door $\to$ traverse $\to$ reach goal), each phase conditioned on the
previous one. A correct world model must encode the \texttt{carrying}
channel, the door-state transition under the unlock action, and the
colour-matching constraint between key and door, all from observation
traces alone.

\subsection{Block and MineSweeper}
\label{app:new-envs}

\paragraph{Block} Our own partially observable implementation of the classic Blocks World family \citep{slaney2001blocks}. The hidden state consists of three columns, each holding a vertical stack of distinct blocks, the gripper's column, and the block it holds (if any). The observation is the pair (stack at the current column, held block); the agent never observes its absolute column or the other columns. Four actions move the gripper left or right or pick up the top block of the current column or put the held block down on it; invalid actions (moving past the edge of the row, picking from an empty column, dropping with an empty gripper) leave the state unchanged. Each episode, the six blocks are scattered uniformly at random over the three columns in random stacking order, and the gripper starts in a uniformly random column; the target configuration is fixed (the sorted tower in column 0). The reward is sparse and terminal ($+1$ on the step that reaches the target), episodes are capped at 120 steps, and returns are discounted with $\gamma=0.98$. The reachable state space has 105{,}840 states and 423{,}360 state--action pairs, which allows every induced program to be compared with the true environment exhaustively (Appendix~\ref{app:model-audit}); optimal solutions of the 30 evaluation instances range from 19 to 62 steps. All methods use the same 15 demonstration episodes and are evaluated on the same 30 held-out instances (10 seeds $\times$ 3 episodes). The difficulty is \emph{partial observability of position}: the agent must infer where it stands in the row from what it has seen and moved, and the demonstrations never show what happens at the edge of the row. The observation distance is the MiniGrid kernel of Appendix~\ref{app:kernel-likelihood} applied to a fixed-length encoding of (stack, held): $d_{\mathrm{obs}} = -\log(\varepsilon + (1-\varepsilon)\,r^{3})$ with $r$ the fraction of matching entries, and $\infty$ when no entry matches or the shapes differ. Rewriting the distance for this suite took five lines of code; no MiniGrid-specific code is involved.

\paragraph{MineSweeper} The MineSweeper task of the POPGym benchmark \citep{morad2023popgym} at its three difficulties: a $4\times4$ board with 16 actions (easy), $6\times6$ with 36 (medium), and $8\times8$ with 64 (hard). Each action clicks one cell of a hidden mine grid. The observation is a short binary vector that one-hot encodes the number of mines adjacent to the clicked cell together with the row and column of the previous action (13, 21, and 28 entries, respectively). A safe click gives a small positive reward and marks the cell as viewed, clicking an already viewed cell gives a smaller negative reward, and clicking a mine terminates the episode with a negative reward; the objective is to reveal all safe cells (14, 30, and 54, respectively). Medium and hard are hard for any model-based agent: the hidden layout has about \(1.9\times10^{6}\) and \(1.5\times10^{11}\) possible mine configurations (against 120 on easy), tracked with \(K=80\) particles, and the planning budget kept from easy (30 node expansions) affords only one to two steps of lookahead, so even the true models win at most 3\% of episodes. The fully observed Direct LLM baseline is not comparable here: the MineSweeper state reveals the full mine layout, so the LLM only needs to click cells without mines, and its residual per-click errors compound over 30 or 54 clicks (a per-click error of about 1.5\% and 2\% already explains its 63\% and 30\% win rates). The model-based methods must instead infer the layout from adjacency counts, which is where they fail. Medium and hard therefore expose an inference and planning limitation shared by all model-based methods rather than a limitation of model induction, \oldtext{and we expect a larger particle budget or a deeper planner to improve all of them.} \newtext{and larger particle and planning budgets alone are not enough to close it: improving all of them requires better inference and planning, not only more of the current ones.} Because the raw discounted return is negative for most policies, we report returns normalised to $[0,1]$ between the worst achievable return ($-0.5-1/n_{\mathrm{safe}}$) and the best ($1$), so that Optimal is $1$ by construction. The demonstrations are 50 random-policy episodes per difficulty (Appendix~\ref{app:experimental-details}). The difficulty is \emph{inference about a hidden layout} from sparse local counts with no spatial structure in the observation itself. The observation distance is the mean squared difference between the two binary vectors ($-\log\varepsilon$ on a shape mismatch), again a few lines of code.

\section{General details on baselines}
\label{app:baselines}

\paragraph{Information regimes}
The methods operate in different information regimes, which we state explicitly. All learning methods receive the same demonstrations (10 episodes on MiniGrid, 15 on Block, 50 on MineSweeper). The LLM-based methods and the tabular baseline interact for 3 online episodes, whereas PPO-GRU and DreamerV3 interact for 1{,}000 online episodes. The LLM-based methods receive the natural-language task description, byte-identical between \emph{Pinductor} and POMDP Coder, whereas the RL and tabular baselines receive none. POMDP Coder, the tabular baseline, and Direct LLM (FO) additionally receive ground-truth states. The true-model reference runs perfect models with a hidden state and a finite planning budget, so it is the level a method could in principle match; Optimal is an upper bound that no partially observable method can reach (Appendix~\ref{app:gt-reference}). DreamerV3 is trained in two regimes, from scratch or with its replay buffer seeded by the same demonstrations the other methods receive (Appendix~\ref{app:dreamer}); the seeded regime is the stronger and the one shown in the main figure.

\subsection{Tabular Baseline}

The tabular baseline learns each of the four components of the POMDP separately as frequency tables from the offline dataset. 
For the \textit{initial model}, it stores the initial states of each episode and then samples proportionally from the frequency at inference-time:
\[\hat{P}_0(s) = \frac{\#\{e : s^e_0 = s\}}{|\mathcal{D}|}\]
For the \textit{transition model}, it counts the number of times each transition $(s,a,s')$ occurs in the dataset, and similarly samples proportionally at inference-time:
\[\hat{P}(s' \mid s,a) = \frac{\#(s,a,s') \in \mathcal{D}}{\#(s,a) \in \mathcal{D}}\]
For the \textit{reward model}, it accumulates the sum of rewards and terminations for each $(s,a,s')$ triple, and returns the mean reward and Bernoulli-sampled termination at inference-time:
\[\hat{R}(s,a,s') = \frac{\sum r_t}{\#\mathcal{D}[s,a,s']} \qquad \hat{P}_\text{term}(s,a,s') = \frac{\#\text{done}}{\#\mathcal{D}[s,a,s']}\]
For the \textit{observation model}, it counts the number of times each observation $o$ is observed after $(s',a)$ pairs, and samples proportionally at inference-time:
\[\hat{P}(o \mid s', a) = \frac{\#(s',a,o) \in \mathcal{D}}{\#(s',a) \in \mathcal{D}}\]
In all cases, unseen conditions fall back to a silent default.
Note that the tabular baseline requires access to privileged information about the hidden state.

\paragraph{Re-run with 1{,}000 demonstrations}
\label{app:tabular-1000}
The tabular estimator is weak in the low-sample regime that is the target of this paper. To show what it becomes with more data, we also re-ran the same estimator on the four harder MiniGrid environments with 1{,}000 demonstration episodes instead of 10, keeping its privileged hidden-state counts and the same 10 evaluation episodes. These demonstrations are generated by the same full-observability A$^*$ oracle used for the paper's datasets, with $\varepsilon$-greedy exploration ($\varepsilon=0.1$) so that the tables cover states off the oracle's path. Table~\ref{tab:l0_baselines} reports this re-run next to the 10-demonstration result of the main experiments.

\subsection{Direct LLM}
\label{app:direct-llm}
The same LLM used by the model-induction methods (Qwen 3.6 Plus) receives the task description and the observation--action history of the current episode and returns the next action directly. There is no induced model, no belief filter, and no planner. In the fully observed variant (FO) the prompt shows the true state at every step; in the partially observed variant (PO) it shows only the observation. Both variants are evaluated on the same seeds and evaluation instances as the model-induction methods. This baseline measures how much of the performance can be obtained from the LLM prior alone, without an explicit world model.

\subsection{PPO-GRU}
\label{app:ppo-gru}
A recurrent model-free policy trained from reward alone: proximal policy optimisation \citep{schulman2017proximal} with a GRU memory, following the recurrent model-free baseline of \citet{Nippogru2021}. The agent receives no natural-language description and no hidden state; it observes the same partial observations as \emph{Pinductor}. It is trained for 1{,}000 online episodes per environment and evaluated on the same instances as the other methods (on MineSweeper we additionally ran LSTM and MLP variants, which performed similarly or worse; we report the GRU variant). On Block, no training episode was ever rewarded, so the policy never leaves the zero-return regime.

\subsection{DreamerV3}
\label{app:dreamer}
DreamerV3 \citep{Hafner2023ARXIV_Mastering_Diverse_Domains}, a latent world-model agent, is run in its official implementation with the smallest model size and a budget of 1{,}000 online episodes per environment; we report the mean return over the last 100 training episodes at that cutoff. It is trained in two regimes that differ only in what the replay buffer contains at the start of training. In the first, the buffer is empty and the agent learns from scratch through interaction alone. In the second, the buffer is pre-loaded with the exact demonstration set the LLM-based methods receive before the 1{,}000 online episodes start; this makes it, together with the tabular baseline, the only non-LLM method that consumes the demonstrations, and the strongest interaction-based hybrid we could build. The seeded regime is the one shown in Fig.~\ref{fig:reward_win_rate} and Table~\ref{tab:l0_baselines}; Table~\ref{tab:full_results} lists both. \oldtext{Each cell is a single training run, so no error bar is reported.} \newtext{The from-scratch arm is a single training run and the demonstration-seeded arm is the mean over three training seeds; no error bar is reported.} On MineSweeper the demonstrations are random rollouts (Appendix~\ref{app:experimental-details}), so seeding the replay changes little and the two regimes give close results.

\subsection{Single-call induction}
\label{app:single-call}
To separate what the LLM prior supplies from what the refinement loop and online interaction add, we run \emph{Pinductor} with a single induction call: one LLM proposal per run, no refinement round, no online learning, deployed as is (a candidate that fails to execute scores zero). This condition is run with 50 seeds of 3 episodes on every environment; results are in Table~\ref{tab:single_call}.

\section{Ablations and Additional Results}
\label{app:add-experiments}

\paragraph{Performance analysis as a function of LLM choice}
Table~\ref{tab:e4_llm_ablation} evaluates how the choice of LLM affects our pipeline when the offline buffer, refinement budget, particle filter, planner, and prompts are held fixed. The results suggest a threshold-like dependence on model quality rather than a smooth scaling trend. With sparse rewards, weaker models often fail to infer a sufficiently correct executable model and therefore obtain near-zero downstream reward. Once the LLM is capable enough to recover the key environment dynamics and reward structure, performance jumps sharply and then largely saturates.

This pattern is clearest in \textsc{Lava}: Qwen3 14B obtains low reward, while Qwen3.6 Plus and Claude Opus 4.7 both reach similar high performance. In \textsc{Unlock}, the same qualitative trend appears, although the task remains harder overall: the weaker model fails, while the two stronger models obtain comparable nonzero performance. These results indicate that, under sparse rewards, LLM capability primarily matters in crossing the threshold needed to synthesize a usable world model; beyond that point, additional model strength yields smaller gains in downstream planning performance.

\begin{table}[H]
\centering
\small
\caption{\textbf{LLM ablation.}
Mean episode reward $\pm$ 95\% CI ($n=10$ seeds) on \textsc{Lava} and \textsc{Unlock} when varying only the LLM used to propose model code. Win rate (reward $> 0.05$) is shown in parentheses. \textbf{Avg.} is the mean of the two task rewards.}
\label{tab:e4_llm_ablation}
\begin{tabular}{lcccccc}
\toprule
\textbf{LLM}
& \multicolumn{2}{c}{\textbf{Lava}}
& \multicolumn{2}{c}{\textbf{Unlock}}
& \multicolumn{2}{c}{\textbf{Avg.}} \\
\cmidrule(lr){2-3}
\cmidrule(lr){4-5}
\cmidrule(lr){6-7}
& \textbf{Reward} & \textbf{Win}
& \textbf{Reward} & \textbf{Win}
& \textbf{Reward} & \textbf{Win} \\
\midrule
Qwen3 14B
& $0.07 \pm 0.14$ & 10\%
& $0.00 \pm 0.00$ & 0\%
& $0.04 \downarrow$ & 5\% \\

Qwen3.6 Plus
& $0.59 \pm 0.10$ & 83\%
& $0.36 \pm 0.14$ & 53\%
& $0.48 \uparrow$ & 68\% \\

Claude Opus 4.7
& $\mathbf{0.61 \pm 0.07}$ & \textbf{87\%}
& $\mathbf{0.37 \pm 0.22}$ & 50\%
& $\mathbf{0.49 \uparrow}$ & \textbf{68\%} \\
\midrule
\textit{Average}
& $0.43$ & 60\%
& $0.24$ & 34\%
& $0.33$ & 47\% \\
\bottomrule
\end{tabular}
\end{table}

\begin{figure}[H]
    \centering
    \includegraphics[width=\linewidth]{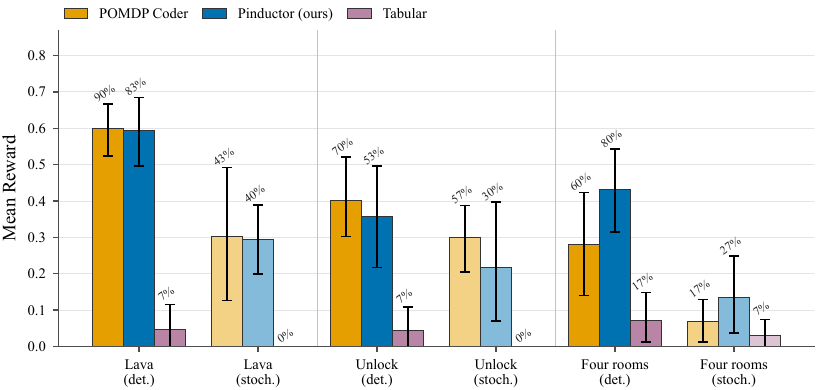}
    \caption{\textbf{Robustness to stochastic environments.}
    Average episode reward and win rate in stochastic MiniGrid variants. \emph{Pinductor} remains comparable to the state-access LLM baseline, suggesting that observation-only model induction remains effective under stochastic transitions and observations.}
    \label{fig:stochastic_robustness}
\end{figure}

\paragraph{Robustness to stochastic environments} Fig.~\ref{fig:stochastic_robustness} tests whether the method remains useful when the environment departs from deterministic dynamics. Performance drops relative to the deterministic versions, as expected, but the comparison with the state-access LLM baseline suggests that removing hidden-state supervision does not by itself cause a disproportionate loss under stochasticity. This indicates that the belief-based evaluation signal can still guide model induction when observations and transitions are less predictable.

\begin{figure}[H]
    \centering
    \includegraphics[width=0.75\linewidth]{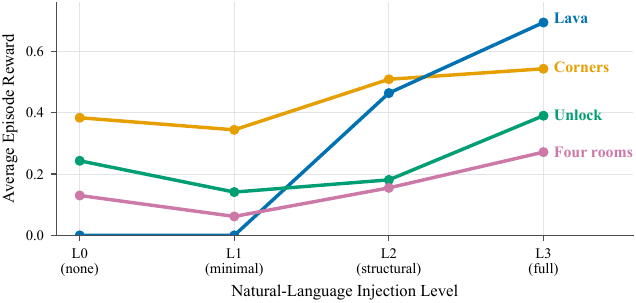}
    \caption{\textbf{Prompt-information ablation.}
    Average episode reward as the natural-language prompt is varied from no task description (L0) to the full description used in main experiments (L3). Performance is task dependent, indicating that \emph{Pinductor} combines LLM priors with trajectory feedback and benefits from structural task cues when the prior alone is underspecified.}
    \label{fig:ablation_prompt_levels}
\end{figure}

\paragraph{Prompt-information ablation} Fig.~\ref{fig:ablation_prompt_levels} tests how much natural-language information the LLM needs in order to propose useful model hypotheses. We sweep four prompt levels: L0 (none), L1 (minimal grid-size information), L2 (structural layout and goal information), and L3 (the full description used in the main experiments). The pattern is task dependent: some environments can be partially recovered from the LLM prior and trajectories alone, while others require structural cues before performance improves. This supports the view that \emph{Pinductor} relies on a combination of semantic priors and empirical feedback; when the prior is underspecified, additional task structure in the prompt can make the model search much more effective.

The sweep conflates two questions that are better kept apart. At L0 it asks whether the LLM can recover a model-learning procedure from trajectories alone; at L3 it asks whether it can exploit an explicitly described environment. Table~\ref{tab:l0_baselines} addresses the first question by comparing \emph{Pinductor} at L0 with baselines that never receive a description. Stripped of every semantic cue, \emph{Pinductor} still beats the tabular baseline trained on the same 10 demonstrations on every environment, holds against the same estimator trained on 1{,}000 demonstrations (ahead on Corners and Four rooms, behind on Lava and Unlock), and is comparable to the online RL baselines trained for a thousand episodes (behind on Corners, ahead on Four rooms, at the floor with DreamerV3 on Lava, in between on Unlock). It remains behind demonstration-seeded DreamerV3, as expected when the same offline data is given to both and the algorithm and a thousand-episode interaction budget are the only differences. The gap between L0 and L3 therefore reflects the method's ability to exploit semantics when they are given, not a shortfall of the L0 setting relative to standard baselines.

\begin{figure}[H]
    \centering
    \includegraphics[width=\linewidth]{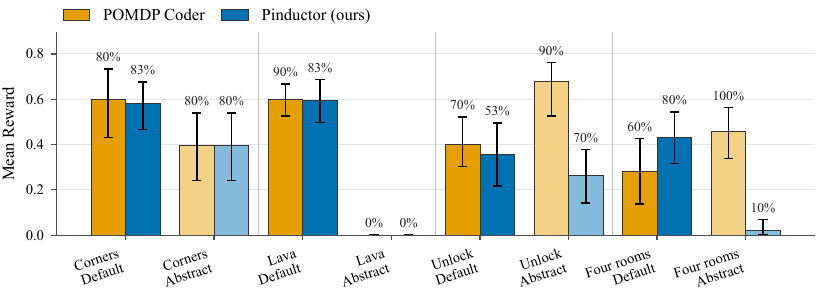}
    \caption{\textbf{Semantic-information ablation.}
    Average episode reward when meaningful environment and object names are replaced with less informative labels. Performance drops across environments, suggesting that \emph{Pinductor} uses semantic cues as structured priors rather than only fitting trajectories.}
    \label{fig:semantic_ablation}
\end{figure}

\paragraph{Semantic-information ablation} Fig.~\ref{fig:semantic_ablation} tests whether performance comes merely from fitting trajectories or also from the semantic content available to the LLM. Replacing meaningful environment and object names with less informative labels reduces performance across environments, indicating that the LLM uses semantic cues to form better initial hypotheses and refinements. This is consistent with the central mechanism of the paper: language models act as structured priors, while the filtering objective uses data to evaluate and correct those priors.

POMDP Coder degrades less under this ablation, and on some environments the abstract labels even help it. A plausible explanation is that its hidden-state trajectories are a semantics-free supervision channel that survives the removal of names and descriptions, and that removing suggestive names can also strip a misleading prior; with 10 seeds and wide intervals, part of the pattern may also be noise. In the outputs of this ablation, explicit references to MiniGrid appear in 48 of the 50 \emph{Pinductor} runs and in 35 of the 50 POMDP Coder runs; this counts mentions rather than causal reliance, and we report it only as a correlate. Appendix~\ref{app:structure-recovery} examines directly what the LLM recovers from trajectories alone, and Appendix~\ref{app:scaffold} ablates the remaining scaffold on Block.

\subsection{Full numeric results behind Fig.~\ref{fig:reward_win_rate}}
\label{app:full-results}
Table~\ref{tab:full_results} lists every arm on every environment, including DreamerV3 trained from scratch (no demonstrations), which Fig.~\ref{fig:reward_win_rate} omits for legibility in favour of the demonstration-seeded variant.

\begin{table}[H]
\centering
\footnotesize
\setlength{\tabcolsep}{4pt}
\caption{\textbf{Full results.} Mean episode reward with win rate in parentheses, all methods and environments. MineSweeper (MS) returns are normalised to $[0,1]$. Both DreamerV3 variants and PPO-GRU interact for 1{,}000 online episodes; the LLM-based methods and the tabular baseline for 3.}
\label{tab:full_results}
\begin{tabular}{lccccc}
\toprule
& \textbf{Empty} & \textbf{Corners} & \textbf{Lava} & \textbf{Unlock} & \textbf{Four rooms} \\
\midrule
Optimal & 0.922 (100\%) & 0.865 (100\%) & 0.727 (100\%) & 0.780 (100\%) & 0.639 (100\%) \\
True models, same planner & 0.922 (100\%) & 0.684 (93\%) & 0.564 (90\%) & 0.748 (100\%) & 0.379 (80\%) \\
POMDP Coder & 0.922 (100\%) & 0.601 (80\%) & 0.599 (90\%) & 0.401 (70\%) & 0.280 (60\%) \\
Direct LLM (full obs.) & 0.906 (100\%) & 0.426 (50\%) & 0.000 (0\%) & 0.000 (0\%) & 0.047 (11\%) \\
Direct LLM (partial obs.) & 0.874 (97\%) & 0.168 (23\%) & 0.060 (10\%) & 0.000 (0\%) & 0.045 (10\%) \\
\emph{Pinductor} & 0.922 (100\%) & 0.582 (83\%) & 0.594 (83\%) & 0.357 (53\%) & 0.433 (80\%) \\
DreamerV3 (demo-seeded) & 0.867 (95\%) & 0.680 (97\%) & 0.257 (39\%) & 0.623 (87\%) & 0.480 (91\%) \\
DreamerV3 (from scratch) & 0.899 (100\%) & 0.701 (100\%) & 0.000 (0\%) & 0.000 (0\%) & 0.000 (0\%) \\
PPO-GRU & 0.834 (100\%) & 0.490 (67\%) & 0.048 (15\%) & 0.285 (40\%) & 0.000 (0\%) \\
Tabular & 0.904 (100\%) & 0.084 (10\%) & 0.046 (7\%) & 0.045 (7\%) & 0.071 (17\%) \\
Random & 0.028 (3\%) & 0.027 (3\%) & 0.000 (0\%) & 0.000 (0\%) & 0.000 (0\%) \\
\bottomrule
\end{tabular}
\vspace{6pt}

\begin{tabular}{lcccc}
\toprule
& \textbf{Block} & \textbf{MS easy} & \textbf{MS medium} & \textbf{MS hard} \\
\midrule
Optimal & 0.383 (100\%) & 1.000 (100\%) & 1.000 (100\%) & 1.000 (100\%) \\
True models, same planner & 0.256 (97\%) & 0.762 (60\%) & 0.394 (3\%) & 0.147 (0\%) \\
POMDP Coder & 0.352 (100\%) & 0.723 (67\%) & 0.184 (0\%) & 0.057 (0\%) \\
Direct LLM (full obs.) & 0.277 (97\%) & 0.721 (83\%) & 0.393 (63\%) & 0.246 (30\%) \\
Direct LLM (partial obs.) & 0.123 (50\%) & 0.489 (67\%) & 0.187 (13\%) & 0.155 (0\%) \\
\emph{Pinductor} & 0.163 (60\%) & 0.676 (67\%) & 0.146 (0\%) & 0.098 (0\%) \\
DreamerV3 (demo-seeded) & 0.165 (75\%) & 0.223 (20\%) & 0.098 (4\%) & 0.087 (0\%) \\
DreamerV3 (from scratch) & 0.000 (0\%) & 0.235 (29\%) & 0.090 (1\%) & 0.054 (0\%) \\
PPO-GRU & 0.000 (0\%) & 0.311 (7\%) & 0.090 (0\%) & 0.061 (0\%) \\
Tabular & 0.000 (0\%) & 0.292 (7\%) & 0.101 (0\%) & 0.090 (3\%) \\
Random & 0.000 (0\%) & 0.211 (0\%) & 0.126 (0\%) & 0.062 (0\%) \\
\bottomrule
\end{tabular}
\end{table}

\subsection{Optimal bound and true-model reference}
\label{app:oracle}
Table~\ref{tab:oracle_vs_gt} reports the two references of Appendix~\ref{app:gt-reference} next to \emph{Pinductor} on all nine environments. The gap between Optimal and the true models run through our own filter and planner is the price of partial observability and finite planning with a perfect model; it is largest on MineSweeper medium and hard and on Four rooms. The gap between the true-model reference and \emph{Pinductor} is the price of induction; it is small on Empty, Lava, Corners, Four rooms, and MineSweeper easy, and large on Unlock and Block.

\begin{table}[H]
\centering
\small
\caption{\textbf{References.} Mean episode reward of the Optimal bound (true state, exact planning), of the true models run through our own filter and planner, and of \emph{Pinductor}, on all nine environments. MineSweeper (MS) returns are normalised to $[0,1]$, so Optimal is $1$ by construction.}
\label{tab:oracle_vs_gt}
\setlength{\tabcolsep}{2.5pt}
\scriptsize
\begin{tabular}{lccccccccc}
\toprule
& \textbf{Empty} & \textbf{Corners} & \textbf{Lava} & \textbf{Unlock} & \textbf{Four rooms} & \textbf{Block} & \textbf{MS easy} & \textbf{MS med.} & \textbf{MS hard} \\
\midrule
Optimal & 0.922 & 0.865 & 0.727 & 0.780 & 0.639 & 0.383 & 1.000 & 1.000 & 1.000 \\
True models, same planner & 0.922 & 0.684 & 0.564 & 0.748 & 0.379 & 0.256 & 0.762 & 0.394 & 0.147 \\
\emph{Pinductor} & 0.922 & 0.582 & 0.594 & 0.357 & 0.433 & 0.163 & 0.676 & 0.146 & 0.098 \\
\bottomrule
\end{tabular}
\end{table}

\subsection{Prior recall versus learning from data}
\label{app:prior-vs-data}
A natural concern is that the LLM recalls the environments from pretraining rather than learning from the trajectories. Two pieces of evidence bear on this. First, Table~\ref{tab:single_call} compares a single induction call, with no refinement and no online learning (Appendix~\ref{app:single-call}), against the full pipeline. Refinement raises the win rate from 8\% to 83\% on Lava and from 5\% to 80\% on Four rooms, the two environments where a single call is worst, and barely changes it on Block, where the single call is already good; the prior proposes, the data decides.

\begin{table}[H]
\centering
\small
\caption{\textbf{Single induction call versus full pipeline.} Mean episode reward with win rate in parentheses. The single call uses one LLM proposal per run (50 seeds $\times$ 3 episodes), with no refinement round and no online learning; the full pipeline is the method of the paper (10 seeds $\times$ 3 episodes).}
\label{tab:single_call}
\begin{tabular}{lcc}
\toprule
& \textbf{Single call} & \textbf{Full pipeline} \\
\midrule
Block & 0.137 (49\%) & 0.163 (60\%) \\
Empty & 0.277 (30\%) & 0.922 (100\%) \\
Corners & 0.118 (14\%) & 0.582 (83\%) \\
Lava & 0.058 (8\%) & 0.594 (83\%) \\
Unlock & 0.078 (11\%) & 0.357 (53\%) \\
Four rooms & 0.024 (5\%) & 0.433 (80\%) \\
MineSweeper (easy) & 0.393 (14\%) & 0.676 (67\%) \\
\bottomrule
\end{tabular}
\end{table}

Second, Block provides a case in which the prior and the belief-based signal contradict each other. The partially observable implementation of Block, together with its transition conventions (behaviour at the edge of the row, guards on invalid pick and drop actions), is ours; these conventions are absent from the description and, because the demonstrations never push the gripper into a wall (0 of 780 steps), they are not identifiable from the offline data: both a blocked move and a wrap-around fit the demonstrations equally well. \oldtext{One of ten seeds ended offline refinement with the wrong (wrap-around) convention, mismatching 32{,}400 of the 423{,}360 transitions, and its plan broke at the first wall. After that single failed episode, every executable candidate the generator produced was exact on the full transition space (8 of 8 across the next two episodes), and the run finished with an exact model. Online interaction is the only stage that collects data outside the demonstration support, and it is what removes the wrong convention. Note that 9 of 10 runs recover the conventions exactly by the end of offline refinement, which the description does not state; this is not evidence of learning them from data, since the prior may simply favour the blocked convention, but the corrected seed is.} \newtext{Two of ten seeds (I and J in Table~\ref{tab:block_seed_audit}) ended offline refinement with the wrong (wrap-around) convention, mismatching 70{,}560 of the 423{,}360 transitions, and kept it in all three evaluation episodes. Online interaction is the only stage that collects data outside the demonstration support, but it does not reliably remove the wrong convention: online refinement also introduced it in three seeds whose first episode ran an exact model (F, G, and H), and only one of them (F) returned to an exact model by its last episode. Note that 8 of 10 runs recover the conventions exactly by the end of offline refinement, which the description does not state; this is not evidence of learning them from data, since the prior may simply favour the blocked convention.}

\subsection{Pinductor without a description versus non-LLM baselines}
\label{app:l0-baselines}
Table~\ref{tab:l0_baselines} places \emph{Pinductor} at prompt level L0 (no natural-language description at all, Fig.~\ref{fig:ablation_prompt_levels}) next to the non-LLM baselines, which never receive a description. Empty is omitted because all methods are near the ceiling there. The tabular baseline is shown both as reported in the main experiments (10 demonstrations) and re-run with 1{,}000 demonstrations (Appendix~\ref{app:tabular-1000}); the RL baselines interact for 1{,}000 online episodes. Stripped of every semantic cue and of the hidden state, \emph{Pinductor} performs on average similarly to the tabular baseline with 100$\times$ more data and privileged state counts, and it uses the offline data as a prior well enough to match baselines trained on roughly 330$\times$ more interaction.

\begin{table}[H]
\centering
\small
\caption{\textbf{No description versus non-LLM baselines.} Mean episode reward on the four harder MiniGrid environments. \emph{Pinductor} at L0 receives no description and 10 demonstrations (3 seeds $\times$ 3 episodes, as in Fig.~\ref{fig:ablation_prompt_levels}). \oldtext{The DreamerV3 rows are single training runs.} \newtext{The from-scratch DreamerV3 row is a single training run; the demonstration-seeded row is the mean over three training seeds, as in Table~\ref{tab:full_results}.}}
\label{tab:l0_baselines}
\begin{tabular}{lcccc}
\toprule
\textbf{Method} & \textbf{Corners} & \textbf{Four rooms} & \textbf{Lava} & \textbf{Unlock} \\
\midrule
\emph{Pinductor} (L0, 10 demos, 3 online episodes) & 0.384 & 0.130 & 0.000 & 0.243 \\
Tabular (10 demos) & 0.084 & 0.071 & 0.046 & 0.045 \\
Tabular (1{,}000 demos, $\varepsilon$-greedy) & 0.322 & 0.098 & 0.070 & 0.299 \\
DreamerV3 (1{,}000 online episodes) & 0.701 & 0.000 & 0.000 & 0.000 \\
PPO-GRU (1{,}000 online episodes) & 0.490 & 0.000 & 0.048 & 0.285 \\
DreamerV3, demo-seeded (1{,}000 online episodes) & \oldtext{0.669} \newtext{0.680} & \oldtext{0.426} \newtext{0.480} & \oldtext{0.486} \newtext{0.257} & \oldtext{0.721} \newtext{0.623} \\
\bottomrule
\end{tabular}
\end{table}

\subsection{Scaffold ablation on Block}
\label{app:scaffold}
Beyond the natural-language description, the LLM receives a code scaffold: a state schema with field names and types, a class API with comments and docstrings, and model templates. Table~\ref{tab:scaffold} ablates this scaffold directly on Block, using the single-call condition so that the effect on the proposed program can be audited exhaustively (a program is \emph{exact} when its transition, reward, and observation functions agree with the true environment on all 423{,}360 state--action pairs). Removing comments and docstrings breaks only the reward function's return convention: no draw is exact overall, but 70\% remain transition-exact. Replacing the description by a neutral sentence is the most damaging change: only 30\% of the draws are executable and 10\% exact. Renaming the classes to generic identifiers has no detectable effect. The scaffold therefore carries little information about the dynamics; the description carries the most.

\begin{table}[H]
\centering
\small
\caption{\textbf{Scaffold ablation on Block (single induction call).} Share of proposed programs that are exact on the full state--action space, with 95\% intervals where the sample allows.}
\label{tab:scaffold}
\begin{tabular}{lcl}
\toprule
\textbf{Change to the prompt} & $n$ & \textbf{Exact (T, R, and O)} \\
\midrule
No ablation & 50 & 36\% ($\pm$ 13\%) \\
Comments and docstrings removed & 10 & 0\%, but only the reward fails: 70\% are transition-exact \\
Description replaced by a neutral sentence & 10 & 10\% ($\pm$ 19\%); only 30\% executable \\
Class names replaced by generic identifiers & 20 & 55\% ($\pm$ 22\%); overlaps the baseline, no effect detected \\
\bottomrule
\end{tabular}
\end{table}

\subsection{Do the induced programs match the true environment?}
\label{app:model-audit}
Because every candidate is a program, we can compare it with the true environment exhaustively rather than by inspection: on Block, all 105{,}840 reachable states $\times$ 4 actions $=423{,}360$ pairs, comparing the induced successor, reward, and observation with the true ones pair by pair; each MiniGrid environment is enumerated the same way. Two questions follow. Where performance almost matches the true-model reference, is the induced model actually the same one, or an alternative that scores well? Where a gap remains, is it attributable to a wrong model?

\paragraph{Block: the gap is the model} Table~\ref{tab:block_seed_audit} audits, for each of the 10 seeds, the model that actually acted in each of its three evaluation episodes (the deployed model can change between episodes through online refinement). Reward and observation models are exact in every episode, so only the transition model is broken out. Two seeds are digit-for-digit identical to the true-model reference on every episode: where performance matches, the model is the same one. Grouped by whether the deployed model is exact (Table~\ref{tab:exactness_return}), the 63\% of episodes that ran an exact model return 0.257 against 0.259 for the true models on the same instances and solve 95\% of them, whereas the 37\% that ran a non-exact model return exactly 0.000. The comparison is controlled rather than correlational: on the same split of instances, the true-model reference shows no difference (Welch $t=0.33$, against $t=12.66$ for \emph{Pinductor}), so instance difficulty is eliminated.

\begin{table}[H]
\centering
\small
\caption{\textbf{Deployed-model audit on Block, per seed.} Share of the 423{,}360 transitions matched by the model that acted in each of the seed's three evaluation episodes, and the seed's mean return against the true models on the same instances.}
\label{tab:block_seed_audit}
\begin{tabular}{lcccc}
\toprule
\textbf{Seed} & \textbf{Transition match, ep.\ 1 / 2 / 3} & \textbf{\emph{Pinductor}} & \textbf{True models} & \textbf{Gap} \\
\midrule
A & 100\% / 100\% / 100\% & 0.307 & 0.307 & $+0.000$ \\
B & 100\% / 100\% / 100\% & 0.213 & 0.213 & $+0.000$ \\
C & 100\% / 100\% / 100\% & 0.301 & 0.326 & $-0.026$ \\
D & 100\% / 100\% / 100\% & 0.246 & 0.256 & $-0.010$ \\
E & 100\% / 100\% / 100\% & 0.222 & 0.214 & $+0.008$ \\
F & 100\% / 83.3\% / 100\% & 0.161 & 0.251 & $-0.090$ \\
G & 100\% / 83.3\% / 83.3\% & 0.095 & 0.232 & $-0.136$ \\
H & 100\% / 83.3\% / 83.3\% & 0.081 & 0.289 & $-0.208$ \\
I & 83.3\% / 83.3\% / 83.3\% & 0.000 & 0.234 & $-0.234$ \\
J & 83.3\% / 83.3\% / 83.3\% & 0.000 & 0.234 & $-0.234$ \\
\bottomrule
\end{tabular}
\end{table}

\begin{table}[H]
\centering
\small
\caption{\textbf{Return by exactness of the deployed model.} Block groups evaluation episodes by the model that acted in them; Lava groups seeds by their final model. No induced MineSweeper model is exact (0 of 149 audited candidates at any difficulty), so the score there does come from a non-equivalent model; for reference, the true models return 0.762 on easy against 0.676 for \emph{Pinductor}.}
\label{tab:exactness_return}
\begin{tabular}{llccc}
\toprule
\textbf{Environment} & \textbf{Share exact} & \textbf{Return, exact} & \textbf{Return, non-exact} & \textbf{True models} \\
\midrule
Block & 63\% of episodes & 0.257 & 0.000 & 0.259 / 0.250 \\
Lava & 50\% of seeds & 0.514 & 0.457 & --- \\
MineSweeper (easy) & 0\% of 149 models & --- & 0.676 & 0.762 \\
\bottomrule
\end{tabular}
\end{table}

When the code differs, it differs in one named place. All eleven non-exact episodes share the same divergence and no other: the two edge cases of the gripper, 70{,}560 of the 423{,}360 pairs, where the true environment clamps the position and the induced program wraps around.

\begin{lstlisting}[style=auditcode]
# ground truth                            # induced program
if action == A_LEFT:                      if action == A_LEFT:
    if pos > 0:                               # Circular left
        pos -= 1                              pos = (pos - 1) % 3
elif action == A_RIGHT:                   elif action == A_RIGHT:
    if pos < n - 1:                           # Circular right
        pos += 1                              pos = (pos + 1) % 3
\end{lstlisting}

Five of the nine action classes (move away from the edge, move into the edge, pick from a non-empty column, and so on) have zero examples in the 780 demonstration steps, covering 24.3\% of the transition space, so the filtered likelihood is flat there by construction. \oldtext{The online phase is the only stage that collects data outside the demonstration support, and it is what removes the wrong convention (Appendix~\ref{app:prior-vs-data}).} \newtext{The online phase is the only stage that collects data outside the demonstration support, but it does not reliably remove the wrong convention: the two seeds that ended offline refinement with it (I and J) keep it in all three episodes, three seeds that started exact (F, G, and H) acquire it during online refinement, and only F returns to an exact model (Table~\ref{tab:block_seed_audit}; Appendix~\ref{app:prior-vs-data}).} The failure is also visible in behaviour: episodes that ran a non-exact model spend 38\% of their actions on moves that change nothing, against 6\% for exact models.

\paragraph{Lava: a wrong model that is right for control} Lava is the counterexample. There, 5 of 10 seeds deploy a non-exact model and all five score: 0.457 on average against 0.514 for the exact ones, one of them returning 0.579, above three of the five exact seeds, while being wrong on 1{,}260 transition cells and 8{,}064 reward cells. The error is a single convention: the induced program moves the lava rule out of the reward function and into the transition function. The true environment lets the agent enter lava and ends the episode there; the induced program forbids entry instead. Both make lava contribute zero future return, so the two greedy policies coincide in every state: with discount $\gamma$, a move into lava is worth $0$ under the true model and $\gamma\,Q(s,\cdot)\le\gamma^{d+1}$ under the induced one, both below the $\gamma^{d}$ of the correct move, so the disagreement never reaches the argmax. On Block the same kind of error lands on the other side of the argmax: believing that a left move at column 0 reaches column 2 makes that action worth more than the two real steps it replaces, so it becomes the argmax. A model error is free when it over-values a dominated action and fatal when it promotes one to the argmax; equivalently, a domain constraint may be placed in the transition function rather than in the reward without loss whenever the constrained action is dominated under both.

\paragraph{Single-call exactness across environments} Auditing the single-call pools of Appendix~\ref{app:single-call} on the full state--action space gives the following counts of exact programs: Block 44/50, Empty 43/50, Corners 42/49, Four rooms 40/49, Lava 5/50, Unlock 5/46, and MineSweeper 0/149 over the three difficulties. Where a single call already reaches exactness (Block, Empty, Corners, Four rooms), the deployed model is the true one; where it does not (Lava, Unlock), the refinement loop is what produces usable models, in line with Table~\ref{tab:single_call}. On Four rooms, 40 of 49 exact models deploy at a return of 0.049 in the single-call condition against an optimum of 0.639, which shows that exactness is necessary but not sufficient: belief filtering and finite-budget planning, not the model, bound the return there.

\subsection{Structure recovery from trajectories alone}
\label{app:structure-recovery}
To separate the ability to recover structure from data from the ability to exploit a described environment, we show the LLM only \texttt{(action, resulting observation)} sequences taken from the very same demonstrations \emph{Pinductor} trains on, with no environment name, no description, no reward, and no termination signal, and ask it to describe the state space, action space, and entities structurally. Observation class names are anonymised (\texttt{Obs(...)}) so that nothing hints at the library; as a control, the same prompts were also run with the original class names. Seven environments (the five MiniGrid tasks, Block, MineSweeper easy), three demonstration subsets each. Table~\ref{tab:structure_recovery} summarises the outcome; anonymisation made no measurable difference anywhere.

\begin{table}[H]
\centering
\small
\caption{\textbf{Structure recovery from anonymised trajectories.} What the LLM infers from (action, observation) sequences alone, three demonstration subsets per environment, judged against the true environment.}
\label{tab:structure_recovery}
\begin{tabular}{lp{4.2cm}p{5.8cm}}
\toprule
\textbf{Environment} & \textbf{Observation structure and action roles} & \textbf{Entities and mechanics} \\
\midrule
Empty & correct in all runs & walls and floor correct; the agent's own marker read as a static object \\
Corners & correct in all runs & agent and goal roles confused or swapped in most runs \\
Lava & correct in all runs & lava hedged as an ``unclear entity'', never identified as a hazard (no reward is shown) \\
Four rooms & correct in all runs & agent marker read as a fixed object, door, or goal \\
Unlock & correct in all runs & key, locked and open door states, and the toggle action correctly reverse-engineered in most runs \\
MineSweeper (easy) & action count correct; roles wrong & never identified; guesses range from a card-matching game to an abstract token puzzle, and the runs disagree with each other \\
Block & correct in all runs & six blocks, LIFO stacks, and the pick and place actions recovered in all runs, including the partial observability itself \\
\bottomrule
\end{tabular}
\end{table}

Three excerpts illustrate the outcome. On Block, with no hint that the values 1--6 are blocks or that the fields are a stack and a hand, the model wrote: \textit{``\texttt{current\_stack}, an ordered tuple representing the vertical arrangement of items at the agent's current location, and \texttt{held}, an integer tracking the grasped item\ldots\ partial observability limited to one stack per step\ldots\ Action 2 acts as a pick-up operation, transferring the last element of \texttt{current\_stack} into \texttt{held}, and Action 3 acts as a place/drop operation\ldots\ stack locations themselves, which serve as LIFO containers.''} On Unlock, semantics were recovered too: \textit{``\texttt{7} represents a key or carriable object, \texttt{6} and \texttt{4} indicate closed and open door states respectively\ldots\ action 5 interacts/toggles environmental elements (changes \texttt{6} to \texttt{4} without displacement).''} On MineSweeper, given only 13-dimensional binary vectors and 16 actions, two of the three runs find a conserved quantity (\textit{``constrained to always contain exactly four \texttt{1}s and nine \texttt{0}s\ldots\ each action triggering a specific permutation pattern''}) but never the task; run with the real class names, one run still concludes it is \textit{``a card-matching task, conceptually analogous to the game of Concentration or Memory''}, so the labels do not carry the task either.

Structure is thus recovered from trajectories alone in every environment, including the partial observability itself. Semantics is harder but not out of reach, and is recovered where the dynamics carry it, as with the key and door. What resists is task identity when the disambiguating signal is exactly what is withheld (reward and termination), as in MineSweeper. The one systematic error is egocentric: the agent's own position marker looks like scenery. Two caveats apply: the judgement of correctness is qualitative, and these results rest on Block, whose partially observable implementation is ours, for the claim that the recovered structure is not recalled from pretraining.

\section{Implementation and Reproducibility Details}
\label{app:implementation}

\subsection{Computational Resources}
\label{subsec:compute}

Table~\ref{tab:wall_clock_nd10} reports the wall-clock time required to run one seed at $N=10$ demonstrations. Since runtime can have heavy-tailed outliers, especially in environments where incorrect model hypotheses lead to longer inference traces, we report the median and interquartile range across seeds rather than the mean and standard deviation.

\begin{table}[h]
\centering
\caption{Wall-clock time per run at $N=10$ demonstrations. Values are median [IQR] minutes over seeds.}
\label{tab:wall_clock_nd10}
\small
\begin{tabular}{lccc}
\toprule
Environment & POMDP Coder & Pinductor (ours) & Tabular \\
\midrule
Corners & 12.0 [9.9, 15.6] & 22.0 [14.0, 28.8] & 0.3 [0.2, 0.3] \\
Lava & 17.2 [15.1, 21.7] & 33.5 [22.1, 44.2] & 1.5 [1.2, 1.7] \\
Unlock & 88.2 [74.7, 100.4] & 90.7 [72.3, 98.1] & 0.6 [0.6, 0.7] \\
Four rooms & 51.1 [18.3, 77.0] & 51.2 [41.8, 58.5] & 2.0 [1.9, 2.1] \\
\bottomrule
\end{tabular}
\end{table}

Table~\ref{tab:compute_hardware} summarizes the compute environment used for the experiments. All runs were executed on a single local CPU workstation without GPU acceleration; LLM calls were made through OpenRouter. Wall-clock measurements therefore include both local computation and remote LLM API latency.

\begin{table}[h]
\centering
\caption{Compute environment used for the experiments.}
\label{tab:compute_hardware}
\small
\begin{tabular}{ll}
\toprule
Resource & Specification \\
\midrule
CPU & AMD Ryzen 7 7800X3D, 8 cores / 16 threads \\
Memory & 30 GiB RAM \\
GPU & None \\
Operating system & Ubuntu 24.04 \\
Execution environment & Single local workstation \\
LLM API provider & OpenRouter \\
\bottomrule
\end{tabular}
\end{table}

\subsection{Cost per run and scaling}
\label{app:cost}

Table~\ref{tab:cost_per_run} reports the LLM calls, tokens, API cost, and wall-clock time of one run (offline induction plus three online episodes) for the LLM-based methods, and the wall-clock time of the RL baselines, on Block, on MiniGrid (mean over environments), and on MineSweeper hard. \emph{Pinductor} issues more calls and consumes more tokens than POMDP Coder because its prompts carry particle-filter diagnostics and disagreement contexts rather than state tuples, and because every candidate is scored by filtering. The RL baselines pay compute rather than API: at a cloud rate of \$0.2 per core-hour, one Block run costs about \$1.36 for PPO-GRU and \$0.70 for DreamerV3, of the same order as the API bill of the LLM methods. Block wall-clock time is dominated by deployment (filtering and planning over 120-step episodes), not by induction. A sweep over the number of MiniGrid distractors ($n_d\in\{2,\ldots,12\}$) shows no monotonic growth of wall-clock time or token usage with task size; it varies difficulty rather than state dimension, however, so it does not test the scaling of the particle filter directly, which remains the limitation discussed in Section~\ref{sec:limitations}.

\begin{table}[H]
\centering
\small
\caption{\textbf{Cost per run.} LLM calls, prompt and completion tokens, API cost, and wall-clock time of one run. RL baselines make no LLM call; their cost is compute ($^*$about \$1.36 per Block run for PPO-GRU and \$0.70 for DreamerV3 at \$0.2 per core-hour).}
\label{tab:cost_per_run}
\begin{tabular}{llrrrr}
\toprule
\textbf{Method} & \textbf{Suite} & \textbf{LLM calls} & \textbf{Tokens in / out} & \textbf{API cost} & \textbf{Wall-clock} \\
\midrule
\emph{Pinductor} & Block & 33 & 375k / 173k & \$0.45 & 3.1 h \\
POMDP Coder & Block & 21 & 63k / 43k & \$0.10 & 0.9 h \\
Single induction call & Block & 1.6 & 17k / 4k & \$0.01 & 1.1 h \\
PPO-GRU & Block & 0 & 0 & \$0$^*$ & 6.8 to 9.5 h \\
DreamerV3 & Block & 0 & 0 & \$0$^*$ & 3.5 h \\
\emph{Pinductor} & MiniGrid (mean) & 14 & 165k / 89k & \$0.22 & 0.7 h \\
POMDP Coder & MiniGrid (mean) & 6 & 58k / 47k & \$0.10 & 0.6 h \\
\emph{Pinductor} & MineSweeper hard & 7 & 170k / 52k & \$0.16 & 2.4 h \\
POMDP Coder & MineSweeper hard & 23 & 98k / 36k & \$0.10 & 0.9 h \\
\bottomrule
\end{tabular}
\end{table}

\subsection{Variance induced by the LLM API}
\label{app:api-variance}

The provider is not deterministic at a fixed seed. Replaying one byte-identical Block induction request 50 times through the API returned three distinct programs in the first three replies, and 18 of the 50 replicates (36\%, $\pm$13 percentage points) were exactly equal to the true environment on the full state--action space; with a varying seed, 14 of 25 (56\%) were. The same replay on MiniGrid gives 34\% to 40\% exact programs on Empty, Corners, and Four rooms and almost none on Lava and Unlock, and 0 of 50 at every MineSweeper difficulty, in line with the single-call exactness counts of Appendix~\ref{app:model-audit}. This variance is inherited by every method that relies on LLM API calls, including POMDP Coder, and is the reason we report seed-level confidence intervals throughout.

\subsection{Randomness Control}
\label{app:randomness-control}

Randomness is controlled at the level of both the run and the evaluation episode. Each experimental condition is evaluated over a fixed set of 10 run seeds. At the start of a run, we seed the Python, NumPy, and Torch random number generators with the run seed. Before each evaluation episode, we reset the Python and NumPy RNGs using a deterministic episode seed, and reset the environment with the same seed. This ensures that, for a given run seed and episode index, all methods are evaluated on the same initial environment instance. For LLM-based methods, we also pass the run seed as the LLM seed when querying the model provider, making LLM sampling as reproducible as the provider allows. We log the served backend and model identity for post-hoc diagnosis.

The error bars in our reward and belief figures capture variability across random seeds. Because the 3 evaluation episodes associated with a single run seed are not independent, we first average the metric within each seed and then compute uncertainty over the resulting 10 seed-level estimates. Unless otherwise stated, error bars denote 95\% percentile bootstrap confidence intervals over these seed-level estimates. Specifically, we resample the 10 seed-level estimates with replacement, compute the mean for each bootstrap sample, and report the 2.5th and 97.5th percentiles of the bootstrap distribution.

We use bootstrap intervals rather than normal-theory intervals because many of our metrics are bounded and can lie near the boundary, for example rewards or win rates close to 0 or 1. In such cases, the usual Gaussian approximation $1.96 \cdot s / \sqrt{n}$ can produce misleading intervals, especially with only $n=10$ independent seeds. The bootstrap procedure avoids assuming an approximately normal sampling distribution and better reflects the empirical variability induced by the random seed. For wall-clock time, which can be heavy-tailed, we instead report the median and interquartile range across seeds.

For the Block and MineSweeper suites and for the baselines added in this revision, we report the intervals produced by their own evaluation pipelines: 95\% intervals over the 30 evaluation episodes on MineSweeper, and normal-theory 95\% intervals over the 10 seed-level means on Block. Arms for which only mean returns are available (Direct LLM and PPO-GRU on MiniGrid, and the DreamerV3 variants\oldtext{, which are single training runs}) are drawn without error bars in Fig.~\ref{fig:reward_win_rate}.

\subsection{Licenses and Third-Party Assets}
\label{subsec:licenses}

MiniGrid~\citep{chevalier2018minimalistic} is used under the Apache~2.0 license. LLM APIs (Claude, Qwen) are used under their respective commercial terms of service. Code from \citet{curtis2025llmguided} is reused with citation; their repository (\url{https://github.com/aidan-curtis/pomdp_coder}) does not specify an explicit license.

\section{Details on prompting methods}
\label{app:prompts}

To make the methodological difference between
\textcolor{PinductorBlue}{\textbf{Pinductor}} and
\textcolor{CurtisOrange}{\textbf{Curtis et al.}}~\citeyearpar{curtis2025llmguided}
concrete, we reproduce verbatim inputs and outputs of (i)~the offline
initial proposal and (ii)~one refinement iteration. Listings are kept in
flow (no figure floats) so each exchange appears exactly where it is
referenced; long demonstration blocks are truncated with explicit
\texttt{[\,...\,]} placeholders to keep the appendix readable. We keep the prompt verbatim from Curtis et al.\, including a templating quirk in their original release where the per-component instructions read \emph{``implement the reward function''} regardless of   
  the actual \texttt{\{model\_name\}}; the LLM correctly infers the intended target from the surrounding context.

Both methods follow a generate--evaluate--refine loop with two regimes:
an \emph{offline} REx loop on the demonstration buffer, and an
\emph{online} regime triggered after each evaluation episode. The two
methods differ on three axes that are visible below:
(i)~\emph{state access} -- \emph{Pinductor}'s prompts only contain
observation--action--reward sequences, while Curtis et al.'s prompts embed
\texttt{Input MinigridState}/\texttt{Output MinigridState} tuples drawn
from privileged hidden states; (ii)~\emph{granularity} -- \emph{Pinductor} issues
\emph{one joint call} returning $(\rho_0, T, O, R)$ at each REx step,
whereas Curtis et al.\ issues \emph{four separate calls} per step (one per
component, instantiating the same template with a different
\texttt{model\_name}); we show their \texttt{transition\_func} call as
representative; (iii)~\emph{feedback signal} -- at refinement time,
\emph{Pinductor} injects the particle-filter kernel score and per-step
disagreement diagnostics, while Curtis et al.\ injects state-level
disagreements between the candidate model's predictions and the recorded
hidden states. Our 'no hidden state' claim refers to state values during trajectories: the LLM is given the state schema (field names + types) as part of the code API but never their realised values during training or evaluation.

Curtis et al.\ ships
an \texttt{online\_update\_models} hook that calls the same REx procedure
as the offline loop, using the same prompt template
(\texttt{po\_model\_refining.txt}). On the deterministic MiniGrid bench,
their online hook is gated by a coverage check that skips the update when
new-episode coverage does not strictly degrade, so the online refinement
calls are not exercised on these runs. The example we show below for
Curtis et al.'s online refinement therefore reuses an iteration-1 REx
refinement call, which is bit-identical to what the online hook would emit
when triggered. \emph{Pinductor}'s online example is a true post-episode~0
refinement call from a recorded run.

\subsection{Initial model proposal (offline, iter 0, \textsc{Lava})}
\label{app:prompts:offline}

\promptheader{PinductorBlue}{Pinductor --- INPUT (joint, partial-obs)}
\begin{lstlisting}[style=oursprompt]
#define system
You are a robot exploring its environment through partial observations.

ENVIRONMENT
-----------
10x10 grid with boundary walls. A vertical wall of lava spans most of one column with exactly one gap to pass through. Agent starts at (1,1). Goal is at the bottom-right corner. The lava column position and gap position VARY between episodes.

GOAL
----


Your task is to model the distribution of initial states, observations, transitions, and rewards in Python.
You need to implement code that captures the dynamics observed in the data below.

IMPORTANT CONSTRAINTS
---------------------
- You have NO access to the true state — it is hidden.
- You only observe the agent's partial field of view after each action.
- Your models must explain how observations change when actions are taken.
- Observations are partial views of a persistent world. Successive observations are snapshots of the same underlying state evolving under your transition model, not independent frames. Keep in mind that the agent itself occupies space in its view, which can momentarily mask part of what is otherwise stable.

You have no access to the true states as they stay hidden. 
Below are samples of Actions and Observations from the environment for different episodes. Each Input (action) is followed by the Output (observation) received after taking that action. You should model the different models that explain the observed dynamics. 
Note: The samples do NOT show the underlying State because it is hidden.

Observation format: Each observation is a grid that represents the agent's field of view (point of view). The agent is always at the same fixed position within this grid. The rest of the grid cells represent the environment as perceived by the agent at that particular step. The grid thus encodes what the agent sees in front of it at each step.

Use the samples and the environment specifications below to make an informed guess about the different models.

You do not observe true initial states directly. Infer the initial-state distribution from the environment description and from the first observation of each episode. 'initial_func' should sample a plausible hidden state consistent with those clues.

OBSERVED DATA
-------------


--- Episode 1 ---

MinigridObservation(image=
[[ 2  2  2]
 [ 1  1 12]
 [ 1  1  1]], dir=0, carrying=None)


Action: forward

MinigridObservation(image=
[[ 2  2  2]
 [11  1 12]
 [11  1  1]], dir=0, carrying=None)
Reward: 0, Terminated: False


Action: forward

MinigridObservation(image=
[[ 2  2  2]
 [ 1 11 12]
 [ 1 11  1]], dir=0, carrying=None)
Reward: 0, Terminated: False


[ ... 68 steps omitted ... ]
--- Episode 5 ---

MinigridObservation(image=
[[ 2  2  2]
 [ 1  1 12]
 [ 1  1  1]], dir=0, carrying=None)


Action: forward

MinigridObservation(image=
[[ 2  2  2]
 [11  1 12]
 [11  1  1]], dir=0, carrying=None)
Reward: 0, Terminated: False


Action: forward

MinigridObservation(image=
[[ 2  2  2]
 [ 1 11 12]
 [ 1 11  1]], dir=0, carrying=None)
Reward: 0, Terminated: False


[ ... 172 steps omitted ... ]

CODE TEMPLATE
-------------
Implement the initial_func, observation_func, transition_func, reward_func, and initial_func functions following the template below.

'''
# type: ignore
# ─────────────────────────────────────────────────────────────────────
# LLM context template — NOT an importable module.
#
# This file is read as plain text and injected into the LLM prompt so
# the model understands the Minigrid state/observation API.  It is
# never imported by Python code.  The authoritative implementations
# live in minigrid_env.py.
# ─────────────────────────────────────────────────────────────────────
from __future__ import annotations

from dataclasses import dataclass
from enum import IntEnum
from typing import Any, List, Optional, Tuple

import numpy as np
from numpy.typing import NDArray

AGENT_DIR_TO_STR = {0: ">", 1: "V", 2: "<", 3: "^"}
DIR_TO_VEC = [
    # Pointing right (positive X)
    np.array((1, 0)),
    # Down (positive Y)
    np.array((0, 1)),
    # Pointing left (negative X)
    np.array((-1, 0)),
    # Up (negative Y)
    np.array((0, -1)),
]

SEE_THROUGH_WALLS = True


class ObjectTypes(IntEnum):
    unseen = 0
    empty = 1
    wall = 2
    open_door = 4
    closed_door = 5
    locked_door = 6
    key = 7
    ball = 8
    box = 9
    goal = 10
    lava = 11
    agent = 12


class Direction(IntEnum):
    facing_right = 0
    facing_down = 1
    facing_left = 2
    facing_up = 3


class Actions(IntEnum):
    left = 0  # Turn left
    right = 1  # Turn right
    forward = 2  # Move forward
    pickup = 3  # Pick up an object
    drop = 4  # Drop an object
    toggle = 5  # Toggle/activate an object
    done = 6  # Done completing the task


@dataclass
class MinigridObservation(Observation):
    """POMDP-clean observation — three perceptual fields the agent reads
    each step.

    Args:
        'image': 3x3 field of view in front of the agent.
        'agent_dir': compass — facing direction
                     (0=right, 1=down, 2=left, 3=up). Proprioception.
        'carrying': object the agent holds ('None' if empty).
                    Proprioception.

    ''reward'' and ''terminated'' are NOT fields of the observation:
    they are *outcomes of the transition* computed by ''reward_func'',
    not perceptual signals that ''obs_func'' should be asked to
    reproduce. The agent still sees them per step (they appear in the
    OBSERVED DATA block alongside each obs), but ''obs_func'' should
    fill only ''image'' (and may set ''agent_dir'' / ''carrying'' if
    the underlying state carries them).

    Absolute position and the full grid are NOT part of the observation
    — the agent never sees them. Any code that reads or writes
    ''obs.agent_pos'' will crash.
    """

    image: NDArray[np.int8]
    agent_dir: int = 0
    carrying: Optional[int] = None


@dataclass
class MinigridState(State):
    """An agent exists in an indoor multi-room environment represented by a
    grid."""

    grid: NDArray[np.int8]  # Indexed as grid[x, y] where x=column, y=row
    agent_pos: Tuple[int, int]  # (x, y) position in the grid
    agent_dir: int  # 0=right, 1=down, 2=left, 3=up
    carrying: Optional[int]  # ObjectType being carried, or None

    @property
    def front_pos(self) -> Tuple[int, int]:
        """Get the (x, y) position of the cell directly in front of the agent."""
        return (
            np.array(self.agent_pos) + np.array(DIR_TO_VEC[self.agent_dir])
        ).tolist()

    @property
    def width(self) -> int:
        return self.grid.shape[0]  # grid[x, y]: shape[0] = width (x-axis)

    @property
    def height(self) -> int:
        return self.grid.shape[1]  # grid[x, y]: shape[1] = height (y-axis)

    def get_type_indices(self, obj_type: int) -> List[Tuple[int, int]]:
        """Return list of (x, y) positions where grid contains obj_type."""
        idxs = np.where(self.grid == obj_type)
        return list(zip(idxs[0], idxs[1]))  # List of (x, y) tuples


    def get_field_of_view(self, view_size: int) -> NDArray[np.int8]:
        """Returns the field of view in front of the agent.

        DO NOT modify this function.
        """

        # Get the extents of the square set of tiles visible to the agent
        # Facing right
        if self.agent_dir == 0:
            topX = self.agent_pos[0]
            topY = self.agent_pos[1] - view_size // 2
        # Facing down
        elif self.agent_dir == 1:
            topX = self.agent_pos[0] - view_size // 2
            topY = self.agent_pos[1]
        # Facing left
        elif self.agent_dir == 2:
            topX = self.agent_pos[0] - view_size + 1
            topY = self.agent_pos[1] - view_size // 2
        # Facing up
        elif self.agent_dir == 3:
            topX = self.agent_pos[0] - view_size // 2
            topY = self.agent_pos[1] - view_size + 1
        else:
            assert False, "invalid agent direction"

        fov = np.full((view_size, view_size), ObjectTypes.wall, dtype=self.grid.dtype)

        # Compute the overlapping region in the grid.
        gx0 = max(topX, 0)
        gy0 = max(topY, 0)
        gx1 = min(topX + view_size, self.grid.shape[0])
        gy1 = min(topY + view_size, self.grid.shape[1])

        # Determine where the overlapping region goes in the padded array.
        px0 = max(0, -topX)
        py0 = max(0, -topY)

        # Copy the overlapping slice.
        fov[px0 : px0 + (gx1 - gx0), py0 : py0 + (gy1 - gy0)] = self.grid[
            gx0:gx1, gy0:gy1
        ]

        for _ in range(self.agent_dir + 1):
            # Rotate left
            fov = np.rot90(fov.T, k=1).T

        agent_pos = (fov.shape[0] // 2, fov.shape[1] - 1)
        fov[agent_pos] = ObjectTypes.agent

        return fov

def initial_func(empty_state:MinigridState):
    """
    Input:
        empty_state (MinigridState): An empty state with only the walls filled into the grid
    Returns:
        state (MinigridState): the initial state of the environment
    """
    raise NotImplementedError

def observation_func(state, action, empty_obs):
    """
    Args:
        state (MinigridState): the state of the environment
        action (int): the previous action that was taken
        empty_obs (MinigridObservation): an empty observation that needs to be filled and returned
    Returns:
        obs (MinigridObservation): observation of the agent
    """
    raise NotImplementedError

def transition_func(state, action):
    """
    Args:
        state (MinigridState): the state of the environment 
        action (int): action to be taken in state 'state'
    Returns:
        new_state (MinigridState): the new state of the environment
    """
    raise NotImplementedError

def reward_func(state, action, next_state):
    """
    Args:
        state (MinigridState): the state of the environment
        action (int): the action to be executed
        next_state (MinigridState): the next state of the environment
    Returns:
        reward (float): the reward of that state
        done (bool): whether the episode is done
    """
    raise NotImplementedError
'''

INSTRUCTIONS
------------
1. First, explain in plain English what you believe the dynamics are:
   - Which hidden variables are randomized at the start of each episode?
   - How does the agent's view change when it moves or turns?
   - What objects exist and how does the agent interact with them?
   - What actions succeed or fail, and why?
2. Then, implement the initial_func, observation_func, transition_func, reward_func, and initial_func functions.

Rules:
- You must implement the initial_func, observation_func, transition_func, reward_func, and initial_func functions.
- Create helper functions only INSIDE the scope of the initial_func, observation_func, transition_func, reward_func, and initial_func functions.
- Do not create new classes or import new modules.
- Do not overfit to specific sample indices — use general rules.
- Put the initial_func, observation_func, transition_func, reward_func, and initial_func functions in a python code block.

Generalization principles
-------------------------
Your goal is to build a **world model**: a compact, general description of
how the environment behaves. The samples below are a few draws from an
infinite distribution of layouts; the model you write must work on draws
you have never seen.

1. **Rules over memorization.** Express conditions on the semantic content
   of states, not on the literal arrays present in the samples. A rule that
   hard-codes a specific layout fragment will break on the next draw.
2. **Cover the space, not the sample.** Enumerate every plausible case the
   environment can produce and decide what should happen in each, even
   when the sample does not exercise it. Missing a rare case is usually
   worse than being slightly inexact on a common one.
3. **Parsimony.** Prefer the shortest rule consistent with the data. Long
   nested branches enumerating specific coordinates almost always overfit.
4. **No episode or step references.** "Episode N ended at step T" is not a
   rule you can use at deployment. Conditions must depend only on the
   arguments your function receives.
5. **One coherent hypothesis.** When several samples disagree, choose the
   single most general rule consistent with all of them — do not stack
   special cases for each one.

\end{lstlisting}

\promptheader{PinductorBlue}{Pinductor --- OUTPUT}
\begin{lstlisting}[style=oursprompt]
Based on the environment description and analyzed observations, here is the inferred model of the dynamics:

**1. Dynamics & Hidden Variables**
- **Grid & Layout:** A fixed $10 \times 10$ grid. Boundary walls are at $x=0, x=9, y=0, y=9$. The interior $(1..8, 1..8)$ is mostly empty.
- **Lava Column:** A single column $x = l$ ($1 < l < 9$) is filled with **lava**, except for exactly one **gap** at $y = g$. This gap allows the agent to pass.
- **Randomization:** The lava column index $l$ and the gap row index $g$ are randomized at the start of each episode.
- **Goal:** Located at $(8, 8)$. Stepping onto it yields a reward of +1 and ends the episode.
- **Transitions:** Moving forward moves the agent 1 step in the current direction. Turning changes direction. Hitting walls, boundaries, or lava blocks forward movement.
- **Observations:** A $3 \times 3$ field of view centered on the agent, rotated to face forward. The agent is always visible at 'image[1, 2]'.

**2. Implementation**

'''python
# type: ignore
# ─────────────────────────────────────────────────────────────────────
# LLM context template — NOT an importable module.
#
# This file is read as plain text and injected into the LLM prompt so
# the model understands the Minigrid state/observation API.  It is
# never imported by Python code.  The authoritative implementations
# live in minigrid_env.py.
# ─────────────────────────────────────────────────────────────────────
from __future__ import annotations

from dataclasses import dataclass
from enum import IntEnum
from typing import Any, List, Optional, Tuple

import numpy as np
from numpy.typing import NDArray

AGENT_DIR_TO_STR = {0: ">", 1: "V", 2: "<", 3: "^"}
DIR_TO_VEC = [
    # Pointing right (positive X)
    np.array((1, 0)),
    # Down (positive Y)
    np.array((0, 1)),
    # Pointing left (negative X)
    np.array((-1, 0)),
    # Up (negative Y)
    np.array((0, -1)),
]

SEE_THROUGH_WALLS = True


class ObjectTypes(IntEnum):
    unseen = 0
    empty = 1
    wall = 2
    open_door = 4
    closed_door = 5
    locked_door = 6
    key = 7
    ball = 8
    box = 9
    goal = 10
    lava = 11
    agent = 12


class Direction(IntEnum):
    facing_right = 0
    facing_down = 1
    facing_left = 2
    facing_up = 3


class Actions(IntEnum):
    left = 0  # Turn left
    right = 1  # Turn right
    forward = 2  # Move forward
    pickup = 3  # Pick up an object
    drop = 4  # Drop an object
    toggle = 5  # Toggle/activate an object
    done = 6  # Done completing the task


@dataclass
class MinigridObservation(Observation):
    """POMDP-clean observation — three perceptual fields the agent reads
    each step.
    """
    image: NDArray[np.int8]
    agent_dir: int = 0
    carrying: Optional[int] = None


@dataclass
class MinigridState(State):
    """An agent exists in an indoor multi-room environment represented by a grid."""
    grid: NDArray[np.int8]  # Indexed as grid[x, y] where x=column, y=row
    agent_pos: Tuple[int, int]  # (x, y) position in the grid
    agent_dir: int  # 0=right, 1=down, 2=left, 3=up
    carrying: Optional[int]  # ObjectType being carried, or None

    @property
    def front_pos(self) -> Tuple[int, int]:
        """Get the (x, y) position of the cell directly in front of the agent."""
        return (
            np.array(self.agent_pos) + np.array(DIR_TO_VEC[self.agent_dir])
        ).tolist()

    @property
    def width(self) -> int:
        return self.grid.shape[0]  # grid[x, y]: shape[0] = width (x-axis)

    @property
    def height(self) -> int:
        return self.grid.shape[1]  # grid[x, y]: shape[1] = height (y-axis)

    def get_type_indices(self, obj_type: int) -> List[Tuple[int, int]]:
        """Return list of (x, y) positions where grid contains obj_type."""
        idxs = np.where(self.grid == obj_type)
        return list(zip(idxs[0], idxs[1]))  # List of (x, y) tuples

    def get_field_of_view(self, view_size: int) -> NDArray[np.int8]:
        """Returns the field of view in front of the agent.
        DO NOT modify this function.
        """
        if self.agent_dir == 0:
            topX = self.agent_pos[0]
            topY = self.agent_pos[1] - view_size // 2
        elif self.agent_dir == 1:
            topX = self.agent_pos[0] - view_size // 2
            topY = self.agent_pos[1]
        elif self.agent_dir == 2:
            topX = self.agent_pos[0] - view_size + 1
            topY = self.agent_pos[1] - view_size // 2
        elif self.agent_dir == 3:
            topX = self.agent_pos[0] - view_size // 2
            topY = self.agent_pos[1] - view_size + 1
        else:
            assert False, "invalid agent direction"

        fov = np.full((view_size, view_size), ObjectTypes.wall, dtype=self.grid.dtype)

        gx0 = max(topX, 0)
        gy0 = max(topY, 0)
        gx1 = min(topX + view_size, self.grid.shape[0])
        gy1 = min(topY + view_size, self.grid.shape[1])

        px0 = max(0, -topX)
        py0 = max(0, -topY)

        fov[px0 : px0 + (gx1 - gx0), py0 : py0 + (gy1 - gy0)] = self.grid[
            gx0:gx1, gy0:gy1
        ]

        for _ in range(self.agent_dir + 1):
            fov = np.rot90(fov, k=1)

        agent_pos = (fov.shape[0] // 2, fov.shape[1] - 1)
        fov[agent_pos] = ObjectTypes.agent

        return fov


def initial_func(empty_state: MinigridState):
    """
    Input:
        empty_state (MinigridState): An empty state with only the walls filled into the grid
    Returns:
        state (MinigridState): the initial state of the environment
    """
    state = MinigridState(
        grid=empty_state.grid.copy(),
        agent_pos=(1, 1),
        agent_dir=0,
        carrying=None
    )

    # Randomize lava column and gap
    # Lava spans most of one column with exactly one gap.
    # We sample a column index between 2 and 7, and a gap row between 1 and 8.
    rng = np.random.default_rng()
    lava_col = rng.integers(2, 8)
    gap_row = rng.integers(1, 9)

    for y in range(1, 9):
        if y == gap_row:
            state.grid[lava_col, y] = ObjectTypes.empty
        else:
            state.grid[lava_col, y] = ObjectTypes.lava

    # Place goal at bottom-right corner (8, 8)
    state.grid[8, 8] = ObjectTypes.goal

    return state


def observation_func(state, action, empty_obs):
    """
    Args:
        state (MinigridState): the state of the environment
        action (int): the previous action that was taken
        empty_obs (MinigridObservation): an empty observation that needs to be filled and returned
    Returns:
        obs (MinigridObservation): observation of the agent
    """
    image = state.get_field_of_view(view_size=3)
    return MinigridObservation(
        image=image.astype(np.int8),
        agent_dir=state.agent_dir,
        carrying=state.carrying
    )


def transition_func(state, action):
    """
    Args:
        state (MinigridState): the state of the environment 
        action (int): action to be taken in state 'state'
    Returns:
        new_state (MinigridState): the new state of the environment
    """
    new_state = MinigridState(
        grid=state.grid.copy(),
        agent_pos=state.agent_pos,
        agent_dir=state.agent_dir,
        carrying=state.carrying
    )

    if action == Actions.left:
        new_state.agent_dir = (state.agent_dir - 1) % 4
    elif action == Actions.right:
        new_state.agent_dir = (state.agent_dir + 1) % 4
    elif action == Actions.forward:
        fx, fy = new_state.front_pos
        # Check boundaries
        if 0 <= fx < new_state.width and 0 <= fy < new_state.height:
            obj = new_state.grid[fx, fy]
            # Can move if empty, goal, or agent (shouldn't happen)
            # Blocked by wall, lava
            if obj != ObjectTypes.wall and obj != ObjectTypes.lava:
                new_state.agent_pos = (fx, fy)
    elif action == Actions.drop:
        new_state.carrying = None
    elif action in (Actions.pickup, Actions.toggle, Actions.done):
        pass  # No effect in this environment

    return new_state


def reward_func(state, action, next_state):
    """
    Args:
        state (MinigridState): the state of the environment
        action (int): the action to be executed
        next_state (MinigridState): the next state of the environment
    Returns:
        reward (float): the reward of that state
        done (bool): whether the episode is done
    """
    # Check if agent moved to the goal
    if next_state.agent_pos != state.agent_pos:
        x, y = next_state.agent_pos
        if next_state.grid[x, y] == ObjectTypes.goal:
            return 1.0, True
    return 0.0, False
'''
\end{lstlisting}

\promptheader{CurtisOrange}{Curtis et al.\ --- INPUT (\texttt{transition\_func} call, full-state)}
\begin{lstlisting}[style=curtisprompt]
#define system
You are a robot exploring its environment. 

Environment Description: 10x10 grid with boundary walls. A vertical wall of lava spans most of one column with exactly one gap to pass through. Agent starts at (1,1). Goal is at the bottom-right corner. The lava column position and gap position VARY between episodes.
Goal Description: 

Your goal is to model the distribution of next states conditioned on actions and previous states. 
You need to implement the python code to model the world, as seen in the provided experiences. 
Please follow the template to implement the code. 
The code needs to be directly runnable (state, action) and return a sample (next_state). 


Below are a few samples from the environment distribution. These are only samples from a larger distribution that your should model.

Input MinigridState: agent_pos=(5, 4)
agent_dir=1
carrying=None
grid=[
[ 2,  2,  2,  2,  2,  2,  2,  2,  2,  2, ],
[ 2,  1,  1,  1,  1,  1,  1,  1,  1,  2, ],
[ 2,  1,  1,  1,  1,  1,  1,  1,  1,  2, ],
[ 2, 11, 11, 11,  1, 11, 11, 11, 11,  2, ],
[ 2,  1,  1,  1,  1,  1,  1,  1,  1,  2, ],
[ 2,  1,  1,  1,  1,  1,  1,  1,  1,  2, ],
[ 2,  1,  1,  1,  1,  1,  1,  1,  1,  2, ],
[ 2,  1,  1,  1,  1,  1,  1,  1,  1,  2, ],
[ 2,  1,  1,  1,  1,  1,  1,  1, 10,  2, ],
[ 2,  2,  2,  2,  2,  2,  2,  2,  2,  2, ],
]

Input Actions: 2
Output MinigridState: agent_pos=(5, 5)
agent_dir=1
carrying=None
grid=[
[ 2,  2,  2,  2,  2,  2,  2,  2,  2,  2, ],
[ 2,  1,  1,  1,  1,  1,  1,  1,  1,  2, ],
[ 2,  1,  1,  1,  1,  1,  1,  1,  1,  2, ],
[ 2, 11, 11, 11,  1, 11, 11, 11, 11,  2, ],
[ 2,  1,  1,  1,  1,  1,  1,  1,  1,  2, ],
[ 2,  1,  1,  1,  1,  1,  1,  1,  1,  2, ],
[ 2,  1,  1,  1,  1,  1,  1,  1,  1,  2, ],
[ 2,  1,  1,  1,  1,  1,  1,  1,  1,  2, ],
[ 2,  1,  1,  1,  1,  1,  1,  1, 10,  2, ],
[ 2,  2,  2,  2,  2,  2,  2,  2,  2,  2, ],
]



Input MinigridState: agent_pos=(2, 1)
agent_dir=0
carrying=None
grid=[
[ 2,  2,  2,  2,  2,  2,  2,  2,  2,  2, ],
[ 2,  1,  1,  1,  1,  1,  1,  1,  1,  2, ],
[ 2,  1,  1,  1,  1,  1,  1,  1,  1,  2, ],
[ 2,  1,  1,  1,  1,  1,  1,  1,  1,  2, ],
[ 2, 11, 11, 11, 11, 11,  1, 11, 11,  2, ],
[ 2,  1,  1,  1,  1,  1,  1,  1,  1,  2, ],
[ 2,  1,  1,  1,  1,  1,  1,  1,  1,  2, ],
[ 2,  1,  1,  1,  1,  1,  1,  1,  1,  2, ],
[ 2,  1,  1,  1,  1,  1,  1,  1, 10,  2, ],
[ 2,  2,  2,  2,  2,  2,  2,  2,  2,  2, ],
]

Input Actions: 2
Output MinigridState: agent_pos=(3, 1)
agent_dir=0
carrying=None
grid=[
[ 2,  2,  2,  2,  2,  2,  2,  2,  2,  2, ],
[ 2,  1,  1,  1,  1,  1,  1,  1,  1,  2, ],
[ 2,  1,  1,  1,  1,  1,  1,  1,  1,  2, ],
[ 2,  1,  1,  1,  1,  1,  1,  1,  1,  2, ],
[ 2, 11, 11, 11, 11, 11,  1, 11, 11,  2, ],
[ 2,  1,  1,  1,  1,  1,  1,  1,  1,  2, ],
[ 2,  1,  1,  1,  1,  1,  1,  1,  1,  2, ],
[ 2,  1,  1,  1,  1,  1,  1,  1,  1,  2, ],
[ 2,  1,  1,  1,  1,  1,  1,  1, 10,  2, ],
[ 2,  2,  2,  2,  2,  2,  2,  2,  2,  2, ],
]




[ ... 2 more (state, action) -> next-state demonstrations omitted for brevity ... ]

Input MinigridState: agent_pos=(3, 1)
agent_dir=1
carrying=None
grid=[
[ 2,  2,  2,  2,  2,  2,  2,  2,  2,  2, ],
[ 2,  1,  1,  1,  1,  1,  1,  1,  1,  2, ],
[ 2,  1,  1,  1,  1,  1,  1,  1,  1,  2, ],
[ 2,  1,  1,  1,  1,  1,  1,  1,  1,  2, ],
[ 2, 11, 11, 11, 11, 11,  1, 11, 11,  2, ],
[ 2,  1,  1,  1,  1,  1,  1,  1,  1,  2, ],
[ 2,  1,  1,  1,  1,  1,  1,  1,  1,  2, ],
[ 2,  1,  1,  1,  1,  1,  1,  1,  1,  2, ],
[ 2,  1,  1,  1,  1,  1,  1,  1, 10,  2, ],
[ 2,  2,  2,  2,  2,  2,  2,  2,  2,  2, ],
]

Input Actions: 2
Output MinigridState: agent_pos=(3, 2)
agent_dir=1
carrying=None
grid=[
[ 2,  2,  2,  2,  2,  2,  2,  2,  2,  2, ],
[ 2,  1,  1,  1,  1,  1,  1,  1,  1,  2, ],
[ 2,  1,  1,  1,  1,  1,  1,  1,  1,  2, ],
[ 2,  1,  1,  1,  1,  1,  1,  1,  1,  2, ],
[ 2, 11, 11, 11, 11, 11,  1, 11, 11,  2, ],
[ 2,  1,  1,  1,  1,  1,  1,  1,  1,  2, ],
[ 2,  1,  1,  1,  1,  1,  1,  1,  1,  2, ],
[ 2,  1,  1,  1,  1,  1,  1,  1,  1,  2, ],
[ 2,  1,  1,  1,  1,  1,  1,  1, 10,  2, ],
[ 2,  2,  2,  2,  2,  2,  2,  2,  2,  2, ],
]





Here is the template for the transition_func function. Please implement
the reward function following the template. The code needs to be directly
runnable.

'''
# type: ignore
from __future__ import annotations

from dataclasses import dataclass
from enum import IntEnum
from typing import Any, List, Optional, Tuple

import numpy as np
from numpy.typing import NDArray

AGENT_DIR_TO_STR = {0: ">", 1: "V", 2: "<", 3: "^"}
DIR_TO_VEC = [
    # Pointing right (positive X)
    np.array((1, 0)),
    # Down (positive Y)
    np.array((0, 1)),
    # Pointing left (negative X)
    np.array((-1, 0)),
    # Up (negative Y)
    np.array((0, -1)),
]

SEE_THROUGH_WALLS = True


class ObjectTypes(IntEnum):
    unseen = 0
    empty = 1
    wall = 2
    open_door = 4
    closed_door = 5
    locked_door = 6
    key = 7
    ball = 8
    box = 9
    goal = 10
    lava = 11
    agent = 12


class Direction(IntEnum):
    facing_right = 0
    facing_down = 1
    facing_left = 2
    facing_up = 3


class Actions(IntEnum):
    left = 0  # Turn left
    right = 1  # Turn right
    forward = 2  # Move forward
    pickup = 3  # Pick up an object
    drop = 4  # Drop an object
    toggle = 5  # Toggle/activate an object
    done = 6  # Done completing the task


@dataclass
class MinigridObservation(Observation):
    """
    Args:
        'image': field of view in front of the agent.

        'agent_pos': agent's position in the real world. It differs from the position
                     in the observation grid.
        'agent_dir': agent's direction in the real world. It differs from the direction
                     of the agent in the observation grid.
        'carrying': what the agent is carrying at the moment.
    """

    image: NDArray[np.int8]
    agent_pos: Tuple[int, int]
    agent_dir: int
    carrying: Optional[int] = None


@dataclass
class MinigridState(State):
    """An agent exists in an indoor multi-room environment represented by a
    grid."""

    grid: NDArray[np.int8]
    agent_pos: Tuple[int, int]
    agent_dir: int
    carrying: Optional[int]

    @property
    def front_pos(self) -> Tuple[int, int]:
        """Get the position of the cell that is right in front of the agent."""

        return (
            np.array(self.agent_pos) + np.array(DIR_TO_VEC[self.agent_dir])
        ).tolist()

    @property
    def width(self) -> int:
        return self.grid.shape[0]

    @property
    def height(self) -> int:
        return self.grid.shape[1]

    def get_type_indices(self, type: int) -> List[Tuple[int, int]]:
        idxs = np.where(self.grid == type)  # Returns (row_indices, col_indices)
        return list(zip(idxs[0], idxs[1]))  # Combine row and column indices


    def get_field_of_view(self, view_size: int) -> NDArray[np.int8]:
        """Returns the field of view in front of the agent.

        DO NOT modify this function.
        """

        # Get the extents of the square set of tiles visible to the agent
        # Facing right
        if self.agent_dir == 0:
            topX = self.agent_pos[0]
            topY = self.agent_pos[1] - view_size // 2
        # Facing down
        elif self.agent_dir == 1:
            topX = self.agent_pos[0] - view_size // 2
            topY = self.agent_pos[1]
        # Facing left
        elif self.agent_dir == 2:
            topX = self.agent_pos[0] - view_size + 1
            topY = self.agent_pos[1] - view_size // 2
        # Facing up
        elif self.agent_dir == 3:
            topX = self.agent_pos[0] - view_size // 2
            topY = self.agent_pos[1] - view_size + 1
        else:
            assert False, "invalid agent direction"

        fov = np.full((view_size, view_size), ObjectTypes.wall, dtype=self.grid.dtype)

        # Compute the overlapping region in the grid.
        gx0 = max(topX, 0)
        gy0 = max(topY, 0)
        gx1 = min(topX + view_size, self.grid.shape[0])
        gy1 = min(topY + view_size, self.grid.shape[1])

        # Determine where the overlapping region goes in the padded array.
        px0 = max(0, -topX)
        py0 = max(0, -topY)

        # Copy the overlapping slice.
        fov[px0 : px0 + (gx1 - gx0), py0 : py0 + (gy1 - gy0)] = self.grid[
            gx0:gx1, gy0:gy1
        ]

        for _ in range(self.agent_dir + 1):
            # Rotate left
            fov = np.rot90(fov.T, k=1).T

        agent_pos = (self.grid.shape[0] // 2, self.grid.shape[1] - 1)
        self.grid[agent_pos] = ObjectTypes.agent

        return fov

def transition_func(state, action):
    """
    Args:
        state (MinigridState): the state of the environment 
        action (int): action to be taken in state 'state'
    Returns:
        new_state (MinigridState): the new state of the environment
    """
    raise NotImplementedError
'''

Explain what you believe is the distribution of next states conditioned on actions and previous states in english.
Additionally, please implement code to model the logic of the world. Please implement the 
code following the template. Only output the definition for ' transition_func . 
You must implement the ' transition_func '  function.
Create any helper function inside the scope of ' transition_func '. 
Do not create any helper function outside the scope of ' transition_func '.
Do not output examples usage. 
Do not create any new classes.
Do not rewrite existing classes. 
Do not import any new modules from anywhere.
Do not overfit to the specific samples.
Put the ' transition_func ' function in a python code block.
\end{lstlisting}

\promptheader{CurtisOrange}{Curtis et al.\ --- OUTPUT}
\begin{lstlisting}[style=curtisprompt]
The environment follows a deterministic grid-world transition model. Given a state and an action, the next state is uniquely determined:
- **Turning (Actions 0 & 1)**: Changes the agent's direction cyclically ('-1' or '+1' modulo 4) without altering position or the environment grid.
- **Forward Move (Action 2)**: The agent attempts to move one cell in its current facing direction. The move succeeds only if the target cell is within grid boundaries and does not contain an impassable obstacle (walls or lava). If successful, 'agent_pos' updates; otherwise, the agent stays in place.
- **Other Actions (3-6)**: Have no effect on the agent's pose or the grid in this specific setup, leaving the state unchanged.
- **Grid & Inventory**: The 'grid' and 'carrying' fields remain static across these transitions. Lava acts as a hard barrier, matching the "wall of lava" description where only the gap allows passage.

'''python
def transition_func(state, action):
    """
    Args:
        state (MinigridState): the state of the environment 
        action (int): action to be taken in state 'state'
    Returns:
        new_state (MinigridState): the new state of the environment
    """
    def _is_cell_walkable(grid, x, y):
        # Check boundaries
        if y < 0 or y >= grid.shape[0] or x < 0 or x >= grid.shape[1]:
            return False
        # Check against impassable terrain (wall and lava)
        cell = grid[y][x]
        return cell != ObjectTypes.wall and cell != ObjectTypes.lava

    new_pos = state.agent_pos
    new_dir = state.agent_dir
    new_carrying = state.carrying

    if action == Actions.left:
        new_dir = (state.agent_dir - 1) % 4
    elif action == Actions.right:
        new_dir = (state.agent_dir + 1) % 4
    elif action == Actions.forward:
        dx, dy = DIR_TO_VEC[state.agent_dir]
        next_x = new_pos[0] + dx
        next_y = new_pos[1] + dy
        
        if _is_cell_walkable(state.grid, next_x, next_y):
            new_pos = (next_x, next_y)
            
    # Actions 3 through 6 do not modify pose or inventory in this environment configuration
    
    return MinigridState(
        grid=state.grid,
        agent_pos=new_pos,
        agent_dir=new_dir,
        carrying=new_carrying
    )
'''
\end{lstlisting}

\subsection{Refinement iteration (\textsc{Unlock} for Pinductor online; \textsc{Lava} iter-1 for Curtis et al.)}
\label{app:prompts:online}

\promptheader{PinductorBlue}{Pinductor --- INPUT (post-episode-0 online refit, joint, particle-filter feedback)}
\lstinputlisting[style=oursprompt]{examples/ours_online_input.txt}

\promptheader{PinductorBlue}{Pinductor --- OUTPUT}
\begin{lstlisting}[style=oursprompt]
Looking at the evaluation results, I can identify several critical issues:

1. **Initial state not properly cleaning old objects**: The 'initial_func' copies 'empty_state.grid' but never clears previous episode's key/goal/door positions before placing new ones, causing contamination.
2. **Wall structure at column 5 not enforced**: The locked door is placed at a random row in column 5, but the rest of column 5 must be filled with walls to form the partition.
3. **Agent position type inconsistency**: 'initial_func' uses tuples while 'transition_func' uses lists, which can cause subtle numpy indexing issues.
4. **Forward movement blocking logic incomplete**: The agent should not move into walls, closed doors, or locked doors, but the current allowed set doesn't explicitly account for all cases.
5. **The 'empty'/wall boundary assumption**: The code assumes boundary walls exist, but they need to be explicitly set in every initialization.

Here's the corrected implementation:

'''python
def initial_func(empty_state: MinigridState) -> MinigridState:
    """Samples a plausible initial state."""
    h = empty_state.grid.shape[1]
    w = empty_state.grid.shape[0]
    
    state = MinigridState(
        grid=np.full((w, h), ObjectTypes.wall, dtype=np.int8),
        agent_pos=(0, 0),
        agent_dir=0,
        carrying=None
    )
    
    # Clear interior (leave boundary walls)
    state.grid[1:w-1, 1:h-1] = ObjectTypes.empty
    
    # Place partition wall at column 5, rows 1 to h-2
    state.grid[5, 1:h-1] = ObjectTypes.wall
    
    # Random agent position in left room (columns 1-4)
    agent_x = random.randint(1, 4)
    agent_y = random.randint(1, h - 2)
    state.agent_pos = (agent_x, agent_y)
    state.agent_dir = random.randint(0, 3)
    
    # Place key in left room (not on agent)
    while True:
        key_x = random.randint(1, 4)
        key_y = random.randint(1, h - 2)
        if (key_x, key_y) != state.agent_pos:
            break
    state.grid[key_x, key_y] = ObjectTypes.key
    
    # Place locked door in partition at column 5
    door_y = random.randint(1, h - 2)
    state.grid[5, door_y] = ObjectTypes.locked_door
    
    # Place goal in right room (columns 6-9)
    while True:
        goal_x = random.randint(6, 9)
        goal_y = random.randint(1, h - 2)
        if (goal_x, goal_y) != state.agent_pos:
            break
    state.grid[goal_x, goal_y] = ObjectTypes.goal
    
    return state

def observation_func(state, action, empty_obs):
    """Returns the agent's current field of view."""
    empty_obs.image = state.get_field_of_view(3)
    empty_obs.agent_dir = state.agent_dir
    empty_obs.carrying = state.carrying
    return empty_obs

def transition_func(state, action):
    """Computes the next state based on the action."""
    new_state = MinigridState(
        grid=state.grid.copy(),
        agent_pos=list(state.agent_pos),
        agent_dir=state.agent_dir,
        carrying=state.carrying
    )
    
    front_x, front_y = new_state.front_pos
    
    if action == Actions.left:
        new_state.agent_dir = (new_state.agent_dir - 1) % 4
    elif action == Actions.right:
        new_state.agent_dir = (new_state.agent_dir + 1) % 4
    elif action == Actions.forward:
        # Check bounds
        if 0 <= front_x < new_state.width and 0 <= front_y < new_state.height:
            target = new_state.grid[front_x, front_y]
            # Can move through: empty, goal, open_door, key
            # Blocked by: wall, locked_door, closed_door
            if target in (ObjectTypes.empty, ObjectTypes.goal, ObjectTypes.open_door, ObjectTypes.key):
                new_state.agent_pos = [front_x, front_y]
    elif action == Actions.pickup:
        if new_state.carrying is None:
            if 0 <= front_x < new_state.width and 0 <= front_y < new_state.height:
                if new_state.grid[front_x, front_y] == ObjectTypes.key:
                    new_state.carrying = ObjectTypes.key
                    new_state.grid[front_x, front_y] = ObjectTypes.empty
    elif action == Actions.drop:
        if new_state.carrying is not None:
            if 0 <= front_x < new_state.width and 0 <= front_y < new_state.height:
                if new_state.grid[front_x, front_y] == ObjectTypes.empty:
                    new_state.grid[front_x, front_y] = new_state.carrying
                    new_state.carrying = None
    elif action == Actions.toggle:
        if 0 <= front_x < new_state.width and 0 <= front_y < new_state.height:
            if new_state.grid[front_x, front_y] == ObjectTypes.locked_door and new_state.carrying == ObjectTypes.key:
                new_state.grid[front_x, front_y] = ObjectTypes.open_door
                
    return new_state

def reward_func(state, action, next_state):
    """Computes reward and termination."""
    reward = 0.0
    done = False
    
    # Terminate and reward when agent steps onto goal via forward action
    if action == Actions.forward:
        ax, ay = next_state.agent_pos
        if 0 <= ax < next_state.width and 0 <= ay < next_state.height:
            if next_state.grid[ax, ay] == ObjectTypes.goal:
                reward = 1.0
                done = True
            
    return reward, done
'''

**Key fixes:**
1. **Proper grid initialization**: Uses 'np.full' with walls, then explicitly clears the interior and places the partition wall at column 5. This eliminates contamination from previous episodes.
2. **Explicit wall placement**: Column 5 is filled with walls at all interior rows, then one cell is replaced with the locked door.
3. **Consistent position types**: Uses 'list(state.agent_pos)' in transition to match tuple-to-list conversion, preventing numpy broadcasting issues.
4. **Bounds checking in reward**: Added explicit bounds validation before accessing the goal position in 'reward_func'.
5. **Clean object placement**: Used 'while True' loops with position checks to avoid placing objects on the agent.
\end{lstlisting}

\promptheader{CurtisOrange}{Curtis et al.\ --- INPUT (\texttt{transition\_func} REx iter 1, identical template to online hook)}
\begin{lstlisting}[style=curtisprompt]
#define system
You are a robot exploring its environment. 

10x10 grid with boundary walls. A vertical wall of lava spans most of one column with exactly one gap to pass through. Agent starts at (1,1). Goal is at the bottom-right corner. The lava column position and gap position VARY between episodes.

Your goal is to model the distribution of next states conditioned on actions and previous states of the world in python. 

You have tried it before and came up with one partially correct solution, but it is not perfect. 

The observed distribution disagrees with the generated model in several cases.
You need to improve your code to come closer to the true distribution.

Environment Description: 10x10 grid with boundary walls. A vertical wall of lava spans most of one column with exactly one gap to pass through. Agent starts at (1,1). Goal is at the bottom-right corner. The lava column position and gap position VARY between episodes.
Goal Description: 

Here is a solution you came up with before. 

'''
# type: ignore
from __future__ import annotations

from dataclasses import dataclass
from enum import IntEnum
from typing import Any, List, Optional, Tuple

import numpy as np
from numpy.typing import NDArray

AGENT_DIR_TO_STR = {0: ">", 1: "V", 2: "<", 3: "^"}
DIR_TO_VEC = [
    # Pointing right (positive X)
    np.array((1, 0)),
    # Down (positive Y)
    np.array((0, 1)),
    # Pointing left (negative X)
    np.array((-1, 0)),
    # Up (negative Y)
    np.array((0, -1)),
]

SEE_THROUGH_WALLS = True


class ObjectTypes(IntEnum):
    unseen = 0
    empty = 1
    wall = 2
    open_door = 4
    closed_door = 5
    locked_door = 6
    key = 7
    ball = 8
    box = 9
    goal = 10
    lava = 11
    agent = 12


class Direction(IntEnum):
    facing_right = 0
    facing_down = 1
    facing_left = 2
    facing_up = 3


class Actions(IntEnum):
    left = 0  # Turn left
    right = 1  # Turn right
    forward = 2  # Move forward
    pickup = 3  # Pick up an object
    drop = 4  # Drop an object
    toggle = 5  # Toggle/activate an object
    done = 6  # Done completing the task


@dataclass
class MinigridObservation(Observation):
    """
    Args:
        'image': field of view in front of the agent.

        'agent_pos': agent's position in the real world. It differs from the position
                     in the observation grid.
        'agent_dir': agent's direction in the real world. It differs from the direction
                     of the agent in the observation grid.
        'carrying': what the agent is carrying at the moment.
    """

    image: NDArray[np.int8]
    agent_pos: Tuple[int, int]
    agent_dir: int
    carrying: Optional[int] = None


@dataclass
class MinigridState(State):
    """An agent exists in an indoor multi-room environment represented by a
    grid."""

    grid: NDArray[np.int8]
    agent_pos: Tuple[int, int]
    agent_dir: int
    carrying: Optional[int]

    @property
    def front_pos(self) -> Tuple[int, int]:
        """Get the position of the cell that is right in front of the agent."""

        return (
            np.array(self.agent_pos) + np.array(DIR_TO_VEC[self.agent_dir])
        ).tolist()

    @property
    def width(self) -> int:
        return self.grid.shape[0]

    @property
    def height(self) -> int:
        return self.grid.shape[1]

    def get_type_indices(self, type: int) -> List[Tuple[int, int]]:
        idxs = np.where(self.grid == type)  # Returns (row_indices, col_indices)
        return list(zip(idxs[0], idxs[1]))  # Combine row and column indices


    def get_field_of_view(self, view_size: int) -> NDArray[np.int8]:
        """Returns the field of view in front of the agent.

        DO NOT modify this function.
        """

        # Get the extents of the square set of tiles visible to the agent
        # Facing right
        if self.agent_dir == 0:
            topX = self.agent_pos[0]
            topY = self.agent_pos[1] - view_size // 2
        # Facing down
        elif self.agent_dir == 1:
            topX = self.agent_pos[0] - view_size // 2
            topY = self.agent_pos[1]
        # Facing left
        elif self.agent_dir == 2:
            topX = self.agent_pos[0] - view_size + 1
            topY = self.agent_pos[1] - view_size // 2
        # Facing up
        elif self.agent_dir == 3:
            topX = self.agent_pos[0] - view_size // 2
            topY = self.agent_pos[1] - view_size + 1
        else:
            assert False, "invalid agent direction"

        fov = np.full((view_size, view_size), ObjectTypes.wall, dtype=self.grid.dtype)

        # Compute the overlapping region in the grid.
        gx0 = max(topX, 0)
        gy0 = max(topY, 0)
        gx1 = min(topX + view_size, self.grid.shape[0])
        gy1 = min(topY + view_size, self.grid.shape[1])

        # Determine where the overlapping region goes in the padded array.
        px0 = max(0, -topX)
        py0 = max(0, -topY)

        # Copy the overlapping slice.
        fov[px0 : px0 + (gx1 - gx0), py0 : py0 + (gy1 - gy0)] = self.grid[
            gx0:gx1, gy0:gy1
        ]

        for _ in range(self.agent_dir + 1):
            # Rotate left
            fov = np.rot90(fov.T, k=1).T

        agent_pos = (self.grid.shape[0] // 2, self.grid.shape[1] - 1)
        self.grid[agent_pos] = ObjectTypes.agent

        return fov



def transition_func(state, action):
    """
    Args:
        state (MinigridState): the state of the environment 
        action (int): action to be taken in state 'state'
    Returns:
        new_state (MinigridState): the new state of the environment
    """
    def _is_cell_walkable(grid, x, y):
        # Check boundaries
        if y < 0 or y >= grid.shape[0] or x < 0 or x >= grid.shape[1]:
            return False
        # Check against impassable terrain (wall and lava)
        cell = grid[y][x]
        return cell != ObjectTypes.wall and cell != ObjectTypes.lava

    new_pos = state.agent_pos
    new_dir = state.agent_dir
    new_carrying = state.carrying

    if action == Actions.left:
        new_dir = (state.agent_dir - 1) % 4
    elif action == Actions.right:
        new_dir = (state.agent_dir + 1) % 4
    elif action == Actions.forward:
        dx, dy = DIR_TO_VEC[state.agent_dir]
        next_x = new_pos[0] + dx
        next_y = new_pos[1] + dy
        
        if _is_cell_walkable(state.grid, next_x, next_y):
            new_pos = (next_x, next_y)
            
    # Actions 3 through 6 do not modify pose or inventory in this environment configuration
    
    return MinigridState(
        grid=state.grid,
        agent_pos=new_pos,
        agent_dir=new_dir,
        carrying=new_carrying
    )

'''


Here are some samples from the real world that were impossible under your model
Input MinigridState: agent_pos=(2, 2)
agent_dir=1
carrying=None
grid=[
[ 2,  2,  2,  2,  2,  2,  2,  2,  2,  2, ],
[ 2,  1,  1,  1,  1,  1,  1,  1,  1,  2, ],
[ 2,  1,  1,  1,  1,  1,  1,  1,  1,  2, ],
[ 2, 11, 11, 11,  1, 11, 11, 11, 11,  2, ],
[ 2,  1,  1,  1,  1,  1,  1,  1,  1,  2, ],
[ 2,  1,  1,  1,  1,  1,  1,  1,  1,  2, ],
[ 2,  1,  1,  1,  1,  1,  1,  1,  1,  2, ],
[ 2,  1,  1,  1,  1,  1,  1,  1,  1,  2, ],
[ 2,  1,  1,  1,  1,  1,  1,  1, 10,  2, ],
[ 2,  2,  2,  2,  2,  2,  2,  2,  2,  2, ],
]

Input Actions: 2
Output MinigridState: agent_pos=(2, 3)
agent_dir=1
carrying=None
grid=[
[ 2,  2,  2,  2,  2,  2,  2,  2,  2,  2, ],
[ 2,  1,  1,  1,  1,  1,  1,  1,  1,  2, ],
[ 2,  1,  1,  1,  1,  1,  1,  1,  1,  2, ],
[ 2, 11, 11, 11,  1, 11, 11, 11, 11,  2, ],
[ 2,  1,  1,  1,  1,  1,  1,  1,  1,  2, ],
[ 2,  1,  1,  1,  1,  1,  1,  1,  1,  2, ],
[ 2,  1,  1,  1,  1,  1,  1,  1,  1,  2, ],
[ 2,  1,  1,  1,  1,  1,  1,  1,  1,  2, ],
[ 2,  1,  1,  1,  1,  1,  1,  1, 10,  2, ],
[ 2,  2,  2,  2,  2,  2,  2,  2,  2,  2, ],
]




And here are some samples from your code under the same conditions
Input MinigridState: agent_pos=(2, 2)
agent_dir=1
carrying=None
grid=[
[ 2,  2,  2,  2,  2,  2,  2,  2,  2,  2, ],
[ 2,  1,  1,  1,  1,  1,  1,  1,  1,  2, ],
[ 2,  1,  1,  1,  1,  1,  1,  1,  1,  2, ],
[ 2, 11, 11, 11,  1, 11, 11, 11, 11,  2, ],
[ 2,  1,  1,  1,  1,  1,  1,  1,  1,  2, ],
[ 2,  1,  1,  1,  1,  1,  1,  1,  1,  2, ],
[ 2,  1,  1,  1,  1,  1,  1,  1,  1,  2, ],
[ 2,  1,  1,  1,  1,  1,  1,  1,  1,  2, ],
[ 2,  1,  1,  1,  1,  1,  1,  1, 10,  2, ],
[ 2,  2,  2,  2,  2,  2,  2,  2,  2,  2, ],
]

Input Actions: 2
Output MinigridState: agent_pos=(2, 2)
agent_dir=1
carrying=None
grid=[
[ 2,  2,  2,  2,  2,  2,  2,  2,  2,  2, ],
[ 2,  1,  1,  1,  1,  1,  1,  1,  1,  2, ],
[ 2,  1,  1,  1,  1,  1,  1,  1,  1,  2, ],
[ 2, 11, 11, 11,  1, 11, 11, 11, 11,  2, ],
[ 2,  1,  1,  1,  1,  1,  1,  1,  1,  2, ],
[ 2,  1,  1,  1,  1,  1,  1,  1,  1,  2, ],
[ 2,  1,  1,  1,  1,  1,  1,  1,  1,  2, ],
[ 2,  1,  1,  1,  1,  1,  1,  1,  1,  2, ],
[ 2,  1,  1,  1,  1,  1,  1,  1, 10,  2, ],
[ 2,  2,  2,  2,  2,  2,  2,  2,  2,  2, ],
]




Here are some samples from the real world that were impossible under your model

[ ... 5 more (state, action) -> next-state demonstrations omitted for brevity ... ]

Input MinigridState: agent_pos=(3, 3)
agent_dir=1
carrying=None
grid=[
[ 2,  2,  2,  2,  2,  2,  2,  2,  2,  2, ],
[ 2,  1,  1,  1,  1,  1,  1,  1,  1,  2, ],
[ 2,  1,  1,  1,  1,  1,  1,  1,  1,  2, ],
[ 2,  1,  1,  1,  1,  1,  1,  1,  1,  2, ],
[ 2, 11, 11, 11,  1, 11, 11, 11, 11,  2, ],
[ 2,  1,  1,  1,  1,  1,  1,  1,  1,  2, ],
[ 2,  1,  1,  1,  1,  1,  1,  1,  1,  2, ],
[ 2,  1,  1,  1,  1,  1,  1,  1,  1,  2, ],
[ 2,  1,  1,  1,  1,  1,  1,  1, 10,  2, ],
[ 2,  2,  2,  2,  2,  2,  2,  2,  2,  2, ],
]

Input Actions: 2
Output MinigridState: agent_pos=(3, 3)
agent_dir=1
carrying=None
grid=[
[ 2,  2,  2,  2,  2,  2,  2,  2,  2,  2, ],
[ 2,  1,  1,  1,  1,  1,  1,  1,  1,  2, ],
[ 2,  1,  1,  1,  1,  1,  1,  1,  1,  2, ],
[ 2,  1,  1,  1,  1,  1,  1,  1,  1,  2, ],
[ 2, 11, 11, 11,  1, 11, 11, 11, 11,  2, ],
[ 2,  1,  1,  1,  1,  1,  1,  1,  1,  2, ],
[ 2,  1,  1,  1,  1,  1,  1,  1,  1,  2, ],
[ 2,  1,  1,  1,  1,  1,  1,  1,  1,  2, ],
[ 2,  1,  1,  1,  1,  1,  1,  1, 10,  2, ],
[ 2,  2,  2,  2,  2,  2,  2,  2,  2,  2, ],
]






Explain what you believe is the distribution of next states conditioned on actions and previous states in english, then improve your code to better model the true distribution.

Please implement the code for the following the template. 
You must implement the ' transition_func ' function. 

The code needs to be directly runnable (state, action) and return a sample (next_state). 

Do not output examples. 
Do not create any new classes. 
Do not rewrite existing classes.
Do not import any new modules from anywhere.
Do not list out specific indices that overfit to the examples, but include ranges.
Put the ' transition_func ' function in a python code block.

\end{lstlisting}

\promptheader{CurtisOrange}{Curtis et al.\ --- OUTPUT}
\lstinputlisting[style=curtisprompt]{examples/curtis_online_output.txt}








\newpage
\section*{NeurIPS Paper Checklist}

\begin{enumerate}

\item {\bf Claims}
    \item[] Question: Do the main claims made in the abstract and introduction accurately reflect the paper's contributions and scope?
    \item[] Answer: \answerYes{}
    \item[] Justification: The abstract and introduction state three contributions — inducing POMDP models without hidden-state access, a belief-based kernel likelihood objective, and empirical characterization of the method's dependencies — each of which is directly evaluated in Section~\ref{sec:experiments} (main results, sample efficiency, LLM ablation, prompt and semantic ablations).

\item {\bf Limitations}
    \item[] Question: Does the paper discuss the limitations of the work performed by the authors?
    \item[] Answer: \answerYes{}
    \item[] Justification: Limitations are discussed in the Discussion section, covering the hand-designed observation distance function, manually collected offline trajectories, the absence of cross-environment generalization evaluation, and the absence of evaluations outside MiniGrid.

\item {\bf Theory assumptions and proofs}
    \item[] Question: For each theoretical result, does the paper provide the full set of assumptions and a complete (and correct) proof?
    \item[] Answer: \answerNA{}
    \item[] Justification: The paper presents no theoretical results.

    \item {\bf Experimental result reproducibility}
    \item[] Question: Does the paper fully disclose all the information needed to reproduce the main experimental results of the paper to the extent that it affects the main claims and/or conclusions of the paper (regardless of whether the code and data are provided or not)?
    \item[] Answer: \answerYes{}
    \item[] Justification: The full algorithm is given in Algorithm~\ref{alg:belief-llm-refinement}; the objective and observation distance function are defined in Section~\ref{sec:methods}; environments, baselines, hyperparameters, seeding protocol, and evaluation procedure are described in Section~\ref{subsec:experimental_setup}; remaining hyperparameter values are provided in Appendix~\ref{app:hyperparameters}. Code and offline trajectory data are released as supplementary material.

\item {\bf Open access to data and code}
    \item[] Question: Does the paper provide open access to the data and code, with sufficient instructions to faithfully reproduce the main experimental results, as described in supplemental material?
    \item[] Answer: \answerYes{}
    \item[] Justification: We release the code and offline trajectory datasets at submission time in a zip file. A GitHub repository containing all elements for reproducing the main experiments will be made available upon acceptance.

\item {\bf Experimental setting/details}
    \item[] Question: Does the paper specify all the training and test details (e.g., data splits, hyperparameters, how they were chosen, type of optimizer) necessary to understand the results?
    \item[] Answer: \answerYes{}
    \item[] Justification: Hyperparameters, seeding protocol, and evaluation procedure are described in Section~\ref{subsec:experimental_setup}; remaining hyperparameter values are provided in Appendix~\ref{app:hyperparameters}. Pinductor is the optimizer itself and is explained in detail.

\item {\bf Experiment statistical significance}
    \item[] Question: Does the paper report error bars suitably and correctly defined or other appropriate information about the statistical significance of the experiments?
    \item[] Answer: \answerYes{}
    \item[] Justification: Error bars / shaded regions are 95\% CI computed over the 10 independent run seeds, each summarised by its mean over the 3 evaluation episodes. The source of variability (run seed) and the evaluation protocol (10 seeds $\times$ 3 episodes) are described in Appendix~\ref{app:randomness-control}.

\item {\bf Experiments compute resources}
    \item[] Question: For each experiment, does the paper provide sufficient information on the computer resources (type of compute workers, memory, time of execution) needed to reproduce the experiments?
    \item[] Answer: \answerYes{}
    \item[] Justification: Compute resource details (hardware, wall-clock time per run, total compute, and LLM API call counts) are described in Appendix~\ref{subsec:compute}.

\item {\bf Code of ethics}
    \item[] Question: Does the research conducted in the paper conform, in every respect, with the NeurIPS Code of Ethics \url{https://neurips.cc/public/EthicsGuidelines}?
    \item[] Answer: \answerYes{}
    \item[] Justification: The research conforms with the NeurIPS Code of Ethics.

\item {\bf Broader impacts}
    \item[] Question: Does the paper discuss both potential positive societal impacts and negative societal impacts of the work performed?
    \item[] Answer: \answerNA{}
    \item[] Justification: This paper presents foundational research on world-model learning from partial observations. There is no direct path to specific negative societal applications.

\item {\bf Safeguards}
    \item[] Question: Does the paper describe safeguards that have been put in place for responsible release of data or models that have a high risk for misuse (e.g., pre-trained language models, image generators, or scraped datasets)?
    \item[] Answer: \answerNA{}
    \item[] Justification: The paper does not release pretrained language models, image generators, or scraped datasets. The released code is a research implementation with no identified misuse risk.

\item {\bf Licenses for existing assets}
    \item[] Question: Are the creators or original owners of assets (e.g., code, data, models), used in the paper, properly credited and are the license and terms of use explicitly mentioned and properly respected?
    \item[] Answer: \answerYes{}
    \item[] Justification: All existing assets are properly credited and licenses mentioned in Appendix~\ref{subsec:licenses}

\item {\bf New assets}
    \item[] Question: Are new assets introduced in the paper well documented and is the documentation provided alongside the assets?
    \item[] Answer: \answerNA{}

    \item[] Justification: The paper does not introduce a new dataset, benchmark, or standalone asset as a primary contribution. The environments are based on MiniGrid, and all third-party assets are credited in Appendix~\ref{subsec:licenses}. Code and trajectory data used for the experiments are made available.

\item {\bf Crowdsourcing and research with human subjects}
    \item[] Question: For crowdsourcing experiments and research with human subjects, does the paper include the full text of instructions given to participants and screenshots, if applicable, as well as details about compensation (if any)?
    \item[] Answer: \answerNA{}
    \item[] Justification: The paper does not involve crowdsourcing or research with human subjects.

\item {\bf Institutional review board (IRB) approvals or equivalent for research with human subjects}
    \item[] Question: Does the paper describe potential risks incurred by study participants, whether such risks were disclosed to the subjects, and whether Institutional Review Board (IRB) approvals (or an equivalent approval/review based on the requirements of your country or institution) were obtained?
    \item[] Answer: \answerNA{}
    \item[] Justification: The paper does not involve crowdsourcing or research with human subjects.

\item {\bf Declaration of LLM usage}
    \item[] Question: Does the paper describe the usage of LLMs if it is an important, original, or non-standard component of the core methods in this research? Note that if the LLM is used only for writing, editing, or formatting purposes and does \emph{not} impact the core methodology, scientific rigor, or originality of the research, declaration is not required.
    \item[] Answer: \answerYes{}
    \item[] Justification: LLMs are a core and original component of Pinductor: they propose and iteratively repair executable POMDP programs based on observation-action trajectories and belief-based feedback. This is described throughout Section~\ref{sec:methods} and evaluated in Section~\ref{sec:experiments}.

\end{enumerate}

\end{document}